\documentclass[preview,10pt]{elsarticle}

\usepackage{lineno,hyperref}

\usepackage{xcolor}

\usepackage{multirow}
\usepackage{multicol}
\usepackage{float}
\usepackage{verbatim}

\usepackage{subfig}
\usepackage{rotating}
\usepackage{caption}

\begin{document}

\begin{frontmatter}

\title{Dynamic Price of Parking Service based on Deep Learning}

\author{Alejandro Luque-Cerpa}
\address{Department of Computer Science and Artificial Intelligence \\ University of Seville, Seville, Spain}
\ead{aluquecerp@gmail.com}

\author{Miguel A. Guti\'errez-Naranjo}
\address{Department of Computer Science and Artificial Intelligence \\ University of Seville, Seville, Spain}
\ead{magutier@us.es}

\author{Miguel C\'ardenas-Montes\corref{mycorrespondingauthor}}
\address{Department of Fundamental Research, Centro de Investigaciones Energ\'eticas Medioambientales y Tecnol\'ogicas, Madrid, Spain}
\cortext[mycorrespondingauthor]{Corresponding author}
\ead{miguel.cardenas@ciemat.es}

\begin{abstract}
The improvement of air-quality in urban areas is one of the main concerns of public government bodies.
This concern emerges from the evidence between the air quality and the public health.
Major efforts from government bodies in this area include monitoring and forecasting systems, banning more pollutant motor vehicles, and traffic limitations during the periods of low-quality air.
In this work, a proposal for dynamic prices in regulated parking services is presented.
The dynamic prices in parking service must discourage motor vehicles parking when low-quality episodes are predicted.
For this purpose, diverse deep learning strategies are evaluated.
They have in common the use of collective air-quality measurements for forecasting labels about air quality in the city.
The proposal is evaluated by using economic parameters and deep learning quality criteria at Madrid (Spain).
\end{abstract}

\begin{keyword}
Air Quality\sep Deep Learning \sep Convolutional Neural Networks \sep LSTM \sep U-Net 
\end{keyword}

\end{frontmatter}


\section{Introduction}
Air pollution is one of the most critical health issues in urban areas with a tough concern for governments and citizens. The scientific literature shows its relation with the population health \cite{Linares2006,Diaz1999,AlberdiOdriozola1998,Nel804,doi:10.1080/10473289.2006.10464485,doi:10.3109/08958378.2011.593587,BoogaardErp,India,ERivas,JMSantamaria}. Furthermore, the population growth in urban areas in the forthcoming years will aggravate this issue. 

At this point, a major contributor to the air pollution is the motor vehicles. German Environment Agency and other studies estimate that road transportation is responsible for about 60\% of emissions of $NO_2$ in cities \cite{BORGE20181561,EBannon}. For monitoring and mitigating the adverse effect of air pollution, governments have proposed a wide variety of actions, including: the creation of networks of monitoring stations in the cities and outskirts, the establishing of grading protocol scenarios, the speed limit reduction in the accesses to the city centers, promotion of public transport, the creation of Low Emission Zones, and prohibition of use of public parking when the air quality falls below critical levels. In the past, some of these actions have demonstrated their efficiency \cite{BORGE20181561,MadridCentral}. However, to the best of our knowledge, a dynamic price for public parking service as a function of the pollution level has not been proposed yet. 

For cities where it is illegal to forbid the parking during the low-quality air episodes, this application aims at discouraging the use of public parking, and therefore the private vehicles, during these episodes. By dynamically adjusting the price of public parking during the high air pollution days, restrictive actions ---over the whole city or by district--- can be implemented. 

In the current work, diverse deep learning algorithms aimed at providing the necessary support for dynamic pricing of public parking are proposed and evaluated, namely, different models based on Convolutional Neural Networks, LSTM layers and U-Time architecture are compared. Our proposals endorse important features such as a collective behavior ---the prediction is based on measurements of multiple monitoring stations---, and measurability ---both from accuracy of prediction and economic side---. In order to fix a realistic prediction adapted to the use of motor vehicles, days are divided in four blocks of 6 hours each one, while the prediction focuses on the block II (from 06:00 to 12:00) and III (from 12:00 to 18:00). The blocks I (from 00:00 to 06:00) and IV (from 18:00 to 24:00) are discarded for this study due to the low outdoor activities during those periods.

This proposal aligns with the Sustainable Development Goals (SDGs) of the United Nations and concretely it may act as enabler for the goals Goal 3 "Ensure healthy lives and promote well-being for all at all ages" and particularly for the Target 3.9. By 2030, substantially reduce the number of deaths and illnesses from hazardous chemicals and air, water and soil pollution and contamination" \cite{SDG}. 

Concerning the data, it has been extracted from the Air Quality Monitoring Network of Madrid \cite{opendatamadrid}. This public repository offers hourly and daily data from more than 24 monitoring stations, including three categories: suburban (stations in parks in urban areas), traffic (term for stations affected by traffic and close to a principal street or road), and background (urban background station affected by both traffic and background pollution). For this work, 12 monitoring stations have been selected.

The paper is organized as follows: In Section \ref{section:MM} the main background of the work, including a description of the dataset, the deep architectures used are presented. The Results and the Analysis are shown in Section \ref{section:results}. Finally, the Conclusions are presented in Section \ref{section:conclusions}.

\section{Methods and Materials\label{section:MM}}
\subsection{Data Acquisition and Preprocessing}

The datasets used in this work are obtained from Madrid's City Council Open Data website\footnote{https://datos.madrid.es/portal/site/egob/}. They include daily and hourly measurements of the concentration of  diverse pollutants, such as $CO$, $NO$ or $CH_4$. We will study the hourly data related to $NO_2$, measured in $\mu$g/m$^3$, between 2010/01 and 2019/12. The dataset contains some missing values due to instrument failure and other reasons. In order to fill the missing data, they are replaced with the corresponding monthly mean. A min-max normalization method for scaling the data to the [0,1] range is also used. 

\begin{figure}
 \centering
    \includegraphics[width=1.0\textwidth]{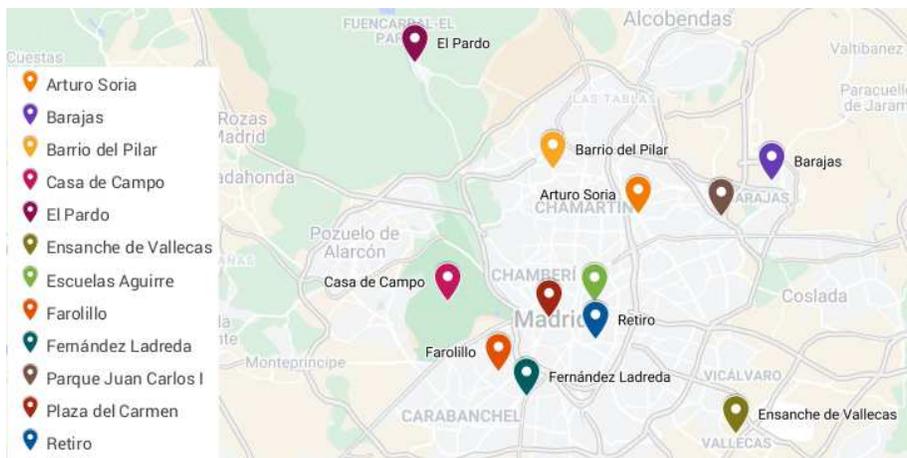}
 \caption{Partial map of Madrid. The monitoring stations used in this study are marked.}
 \label{fig:stations}
\end{figure}

The data are obtained from 24 air quality stations distributed in 5 different zones. Our data is collected from 12 of these stations: Escuelas Aguirre (EA) Barrio del Pilar (BP), Plaza del Carmen (PlC), Retiro (Re), Ensanche de Vallecas (EV), Arturo Soria (AS), Barajas (Ba), Juan Carlos I (PJCI), El Pardo (ElP), Casa de Campo (CC), Fern\'andez Ladreda (FL) y Farolillo (Faro). The list of stations includes suburban, traffic, and background (Fig.  \ref{fig:stations}).

Since the goal of this work is to compute a dynamic price for the regulated parking service during the periods where the main outdoors activities are undertaken, as pointed above, the day is divided in four blocks: one block for each quarter of the day. The proposal focuses on the air quality prediction for the second (6:00-12:00) and third (12:00-18:00) blocks. 

Regarding the air-quality levels, Madrid City Council through the \textit{Action protocol for nitrogen dioxide pollution episodes}\footnote{https://www.madrid.es/UnidadesDescentralizadas/AreasUrbanas\_EducacionAmbiental/\newline Catalogo/AirQualityPlan2011-15.pdf} defines three alert levels:

\begin{itemize}
    \item Pre-warning: when in any two stations of the same area, the 180 $\frac{\mu g}{m^3}$ of $NO_2$ level is exceeded for two consecutive hours simultaneously or in any three stations of the surveillance network the 180 $\frac{\mu g}{m^3}$ level is exceeded for three consecutive hours simultaneously.
    \item Warning: when in any two stations of the same zone the 200 $\frac{\mu g}{m^3}$ of $NO_2$ level is exceeded during two consecutive hours simultaneously or in any three stations of the surveillance network the 200 $\frac{\mu g}{m^3}$ are exceeded during three consecutive hours simultaneously.
    \item Alert: when any three stations in the same zone (or two if it is zone 4) exceed 400 $\frac{\mu g}{m^3}$ for three consecutive hours simultaneously.
\end{itemize}

Let us remark that these pollution levels are not coherent with the air quality of a city like Madrid: the 200 $\frac{\mu g}{m^3}$ of $NO_2$ level is hardly reached, and the 400 $\frac{\mu g}{m^3}$ of $NO_2$ level is never reached. Because of this, two different rules with four alert levels each of them are proposed in this work. Instead of levels based on absolute values, our proposal is based on {\it annual percentiles}. The first rule is the following:

{\bf Rule I}
\begin{itemize}
    \item Pre-warning: when the 75th annual percentile is exceeded for three consecutive hours simultaneously in any three stations of the surveillance network.
    \item Warning: when the 90th annual percentile is exceeded for three consecutive hours simultaneously in any three stations of the surveillance network.
    \item Alert: when the 95th annual percentile is exceeded for three consecutive hours simultaneously in any three stations of the surveillance network.
    \item No alert: if none of the previous conditions is satisfied.
\end{itemize}

Every 6-hours group is classified in one of these four alert levels. Since the \textit{warning} and \textit{alert} level are so similar, it is expected that deep learning algorithms will hardly differentiate them. Because of this, a second rule ---less stringent--- with another four pollution levels are proposed:

{\bf Rule II}
\begin{itemize}
    \item Pre-warning: when the 50th annual percentile is exceeded for three consecutive hours simultaneously in any three stations of the surveillance network.
    \item Warning: when the 75th annual percentile is exceeded for three consecutive hours simultaneously in any three stations of the surveillance network.
    \item Alert: when the 95th annual percentile is exceeded for three consecutive hours simultaneously in any three stations of the surveillance network.
    \item No alert: if none of the previous conditions is satisfied.
\end{itemize}

\subsection{Convolutional Neural Networks}
Convolutional Neural Networks (CNN) is one of the most popular Deep Learning architectures \cite{Goodfellow-et-al-2016}. They are designed to deal with pieces of information placed on a $n$-dimensional grid and, therefore, they are profusely used to deal with digital images, although nowadays they have applications in a wide range of research areas including time series analysis \cite{DBLP:conf/hais/Montes19} or navigation in indoor environments \cite{muratov2020convolutional}. Many different architectures have been defined in the class of CNN, but all of them consist of a sequence of convolutional layers, which deal with local environments of the grid by using the so-called {\it kernels} together with other kind ways of processing information, which can involve pooling-based layers, dropout and other regularization techniques.

Although CNN are usually associated with image or audio classification ---2D grid examples---- or video sequence ---3D grid examples---, it can also be applied to univariate time series analysis ---1D grid examples--- or multivariate time series ---2D grid examples---. The key idea in all these applications is the use of several convolutional layers in order to abstract local semantic features by the use of kernels which can be optimized in the training process.

In this paper, our target is to predict the alert level of pollution in a city from a dataset which includes spatio-temporal information. The data has a time-line dimension but also a spatial dimension due the geographical distribution of the weather stations. Such distribution produces some kind of correlation in the data, since close stations report close pollution levels.

In order to handle such complexity, several Deep Learning architectures have been considered. Data have been presented as a two-dimensional matrix with one column for each weather station and a row for each time unit. Since CNN is adapted to data placed on a grid, one of the considered options has been to apply this architecture to our 2D matrix.

\subsection{LSTM\label{section:LSTM}}
If CNN has become the standard tool for studying images with Deep Learning techniques, Long short-term memory (LSTM) \cite{HochreiterS1997} architecture has become a standard for the study of sequential information. This architecture belongs to the so-called gated recurrent neural networks and the main motivation for using them to create paths along time to avoid the vanishing gradient problem. For this purpose, instead of considering simple neurons which apply an activation function to the affine transformation of inputs, LSTM architectures consider the so-called {\it LSTM cells} which have internal recurrence in addition to the outer recurrence of the recurrent neural network.

The key idea on these LSTM cells is to keep information in a {\it cell state} for later, preventing older signals from vanishing during processing. LSTM systems can add or remove information of such cell states by structures called {\it gates}. In LSTM, there are three types of gates: input gate, forget gate and output gate. In 2014, Cho et al. \cite{Choetal2014} presented a new type of gated recurrent neural network with fewer parameters than LSTM, as it lacks an output gate.

As pointed above, collected pollution data are sequentially grouped by time units. Such a temporal dimension has been exploited in our analysis of the data by using a LSTM architecture.

\subsection{U-Time}
In 2015, Ronneberger {\it al} \cite{Ronnenberg2015} presented a new Deep Learning architecture for biomedical image segmentation called U-net. The networks with this architecture are basically CNN endowed with new features. Technically, a contracting path and an expansive path are considered, which give to the model the U-shaped architecture. The contracting path is a standard CNN which includes convolutions layers, ReLU as activation functions and max pooling operations. As usual, in the convolutional layers, the spatial information is reduced while feature information is increased. The original part of the U-net architecture is the expansive pathway which combines the feature and spatial information through a sequence of up-convolutions and concatenations with high-resolution features from the contracting path.

In 2019, Perslev {\it et al.} \cite{perslev2019utime} presented a fully convolutional encoder-decoder network called U-Time inspired by the U-Net architecture. U-Time adopts basic concepts from U-Net for 1D time-series segmentation by mapping a whole sequence to a dense segmentation in a single forward pass. Originally, U-Time was used for studying time series segmentation applied to sleep staging. In this paper, U-time is used as inspiration for our study of the air-quality in Madrid. In a similar way to the study based on simple CNN architectures, the U-net study also exploits the 2D representation of the data on space and time.

\section{Results and Analysis\label{section:results}}

Since the goal of this paper is to predict the pollution level of six-hours blocks knowing the previous pollution levels, different models and initial conditions are considered. As pointed above, two different rules of pollution alert levels are considered: Rule I for the 95th, 90th and 75th percentiles, and Rule II for the 95th, 75th and 50th percentiles. Three distinct sizes of input data are considered: one day of data per station (24x12 values), three days of data per station (72x12 values) and a week of data per station (168x12 values).

Some considerations must be taken into account. Firstly, the classes are highly unbalanced: most of the days, the pollution alert level is null. Secondly, the 95th, 90th and 75th percentiles are closer together than the 95th, 75th and 50th percentiles, so we expect the models to obtain worse results for the first pollution alert levels than for the second ones. Finally, it is better to obtain false positives than false negatives, given that the ultimate objective is reducing pollution.

\subsection{Confusion Matrix based Analysis}

In this work, four different pollution alert levels are defined: Pre-Warning, Warning, Alert or No-alert. To correctly evaluate the performance of a model predicting the pollution alert level, it is necessary to obtain the accuracy of such model for each of these levels. The results are presented using confusion matrices, where the performance of the different models can be correctly visualized, and more detailed analysis is allowed. Each row of a matrix represents the instances in an actual class, while each column represents the instances in a predicted class. 

\subsubsection{LSTM}
In this first approach, a model based on LSTM cells is considered, namely a model with two LSTM layers, the first one with 50 neurons and the second one with 10 neurons. After these layers a Dropout layer \cite{JMLR:v15:srivastava14a} with index 0.05 was included. Finally, an output layer with four neurons and softmax as activation function is added. Cross-entropy loss has been used for training. After training the LSTM with different data blocks, input sizes and pollution levels, the confusion matrices are obtained. Let us remark that the classes have been balanced before training in order to avoid overfitting.

\begin{figure*}
\centering{\renewcommand{\arraystretch}{1.1}
\rotatebox{90}{
\begin{minipage}[c][][c]{\textheight}
\centering
  \subfloat[Second block. 24 hours.]{
    \includegraphics[width=0.331\textwidth]{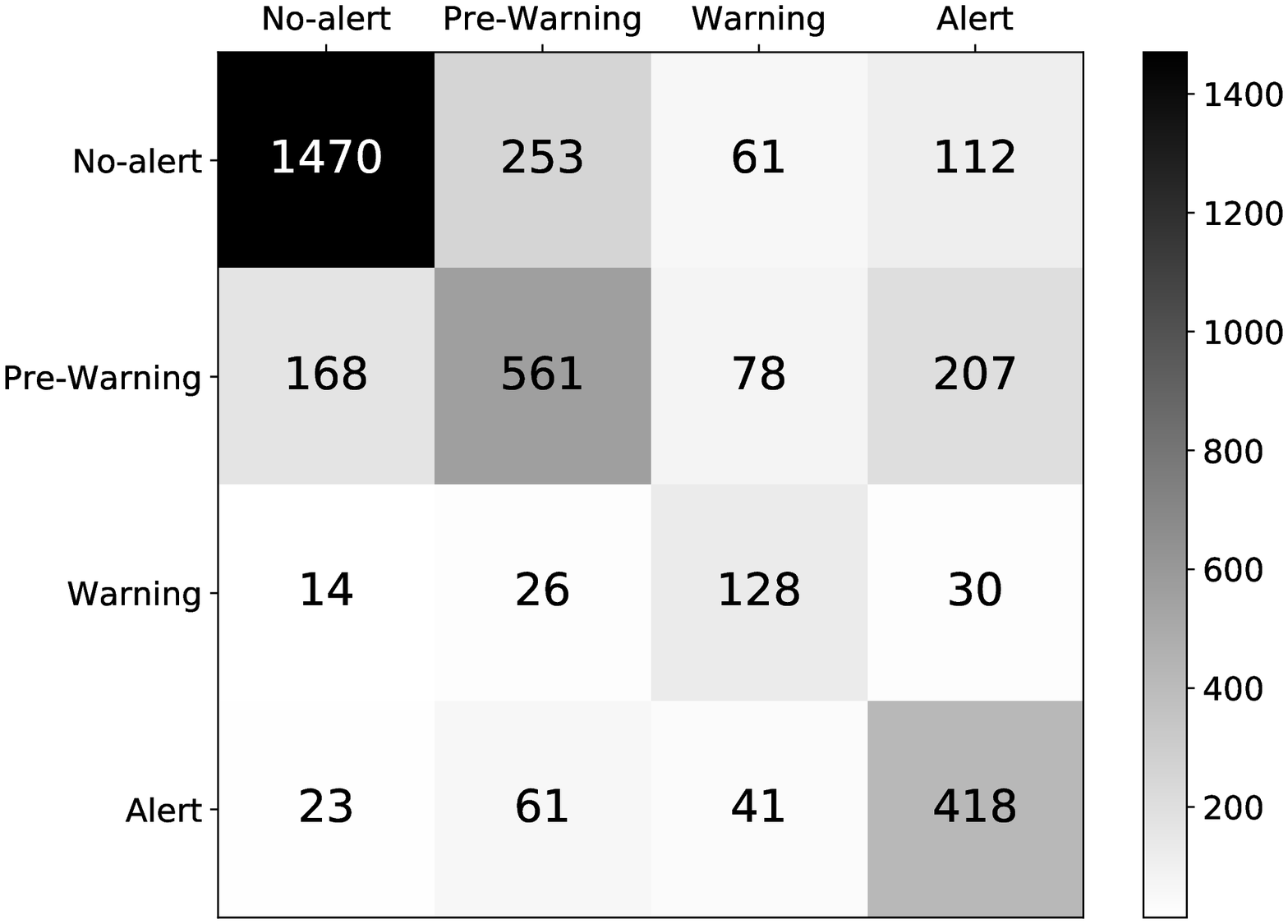}
    \label{fig:LSTM:Bloque2LSTM959075:1dia}
  }
  \subfloat[Second block. 72 hours.]{
    \includegraphics[width=0.331\textwidth]{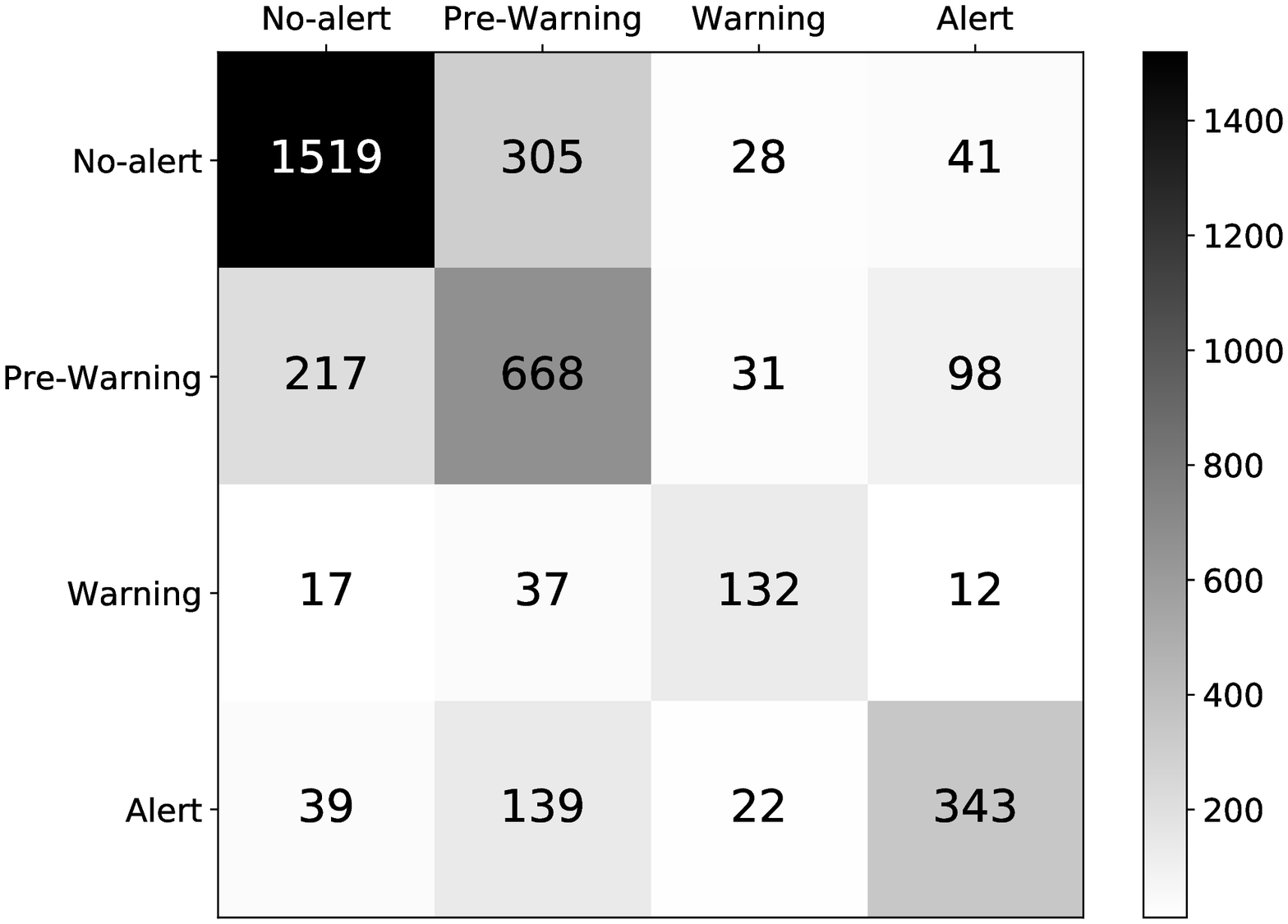}
    \label{fig:LSTM:Bloque2LSTM959075:3dia}
  }
  \subfloat[Second block. 168 hours.]{
    \includegraphics[width=0.331\textwidth]{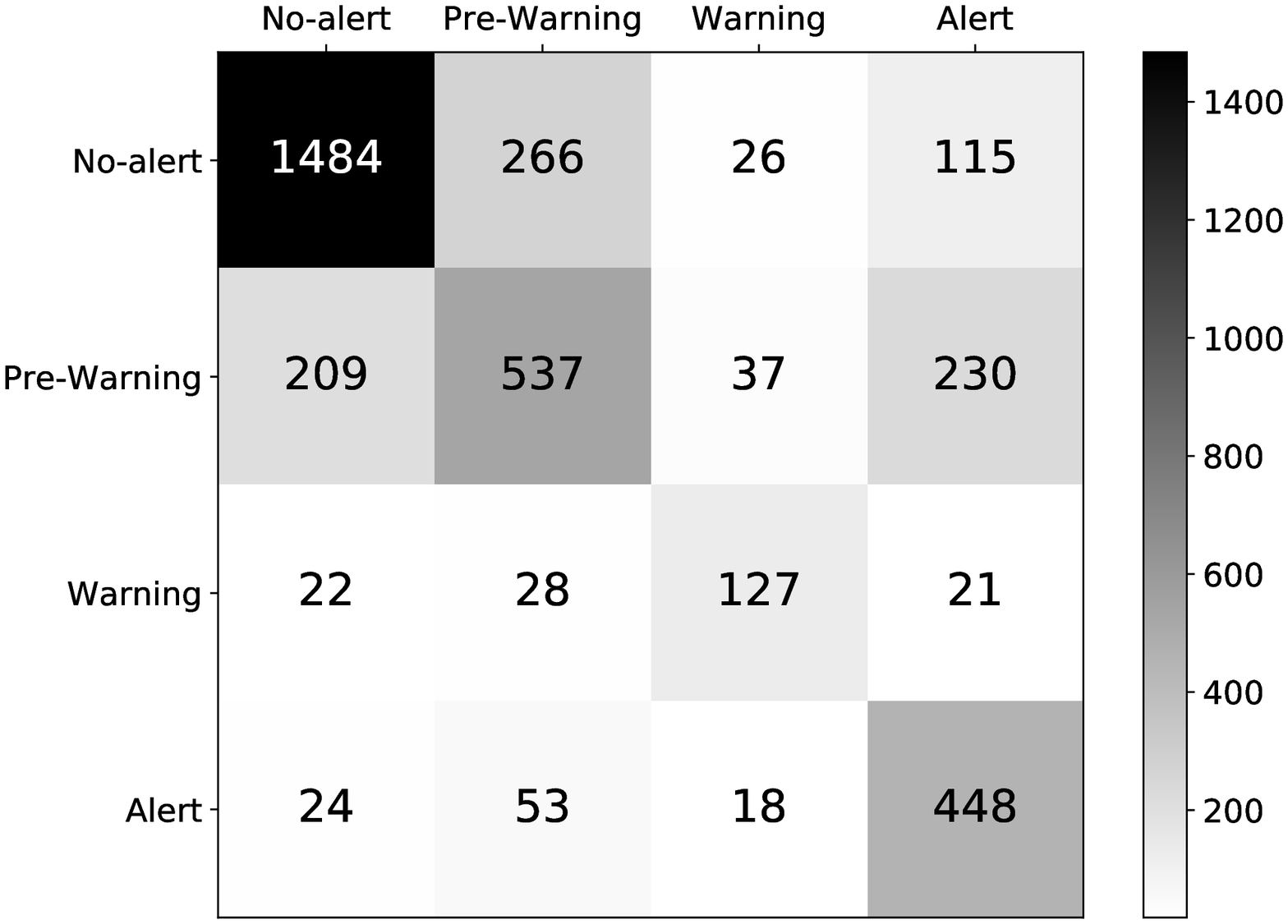}
    \label{fig:LSTM:Bloque2LSTM959075:7dia}
  }\\
  \subfloat[Third block. 24 hours.]{
    \includegraphics[width=0.331\textwidth]{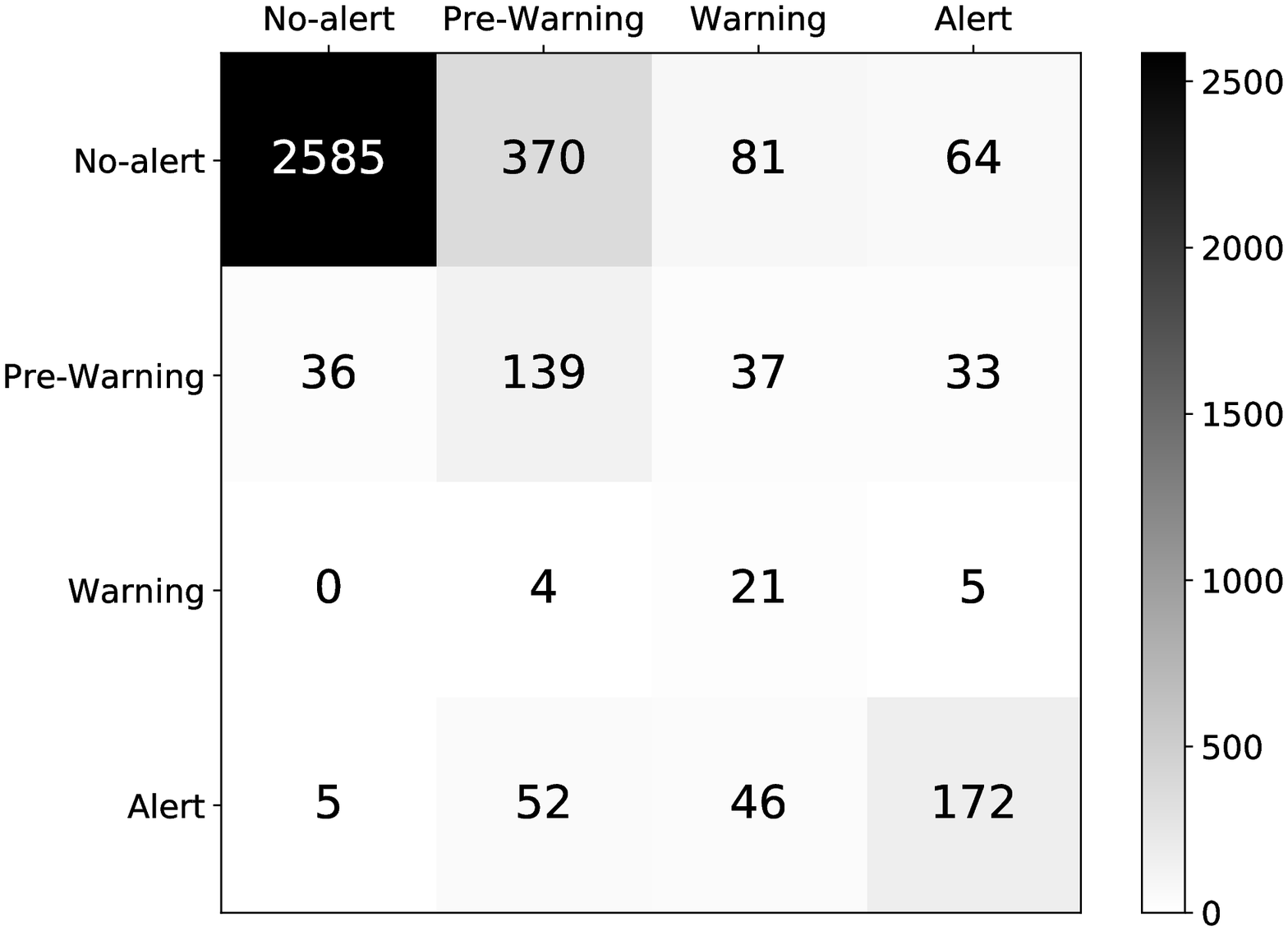}
    \label{fig:LSTM:Bloque3LSTM959075:1dia}
  }
  \subfloat[Third block. 72 hours.]{
    \includegraphics[width=0.331\textwidth]{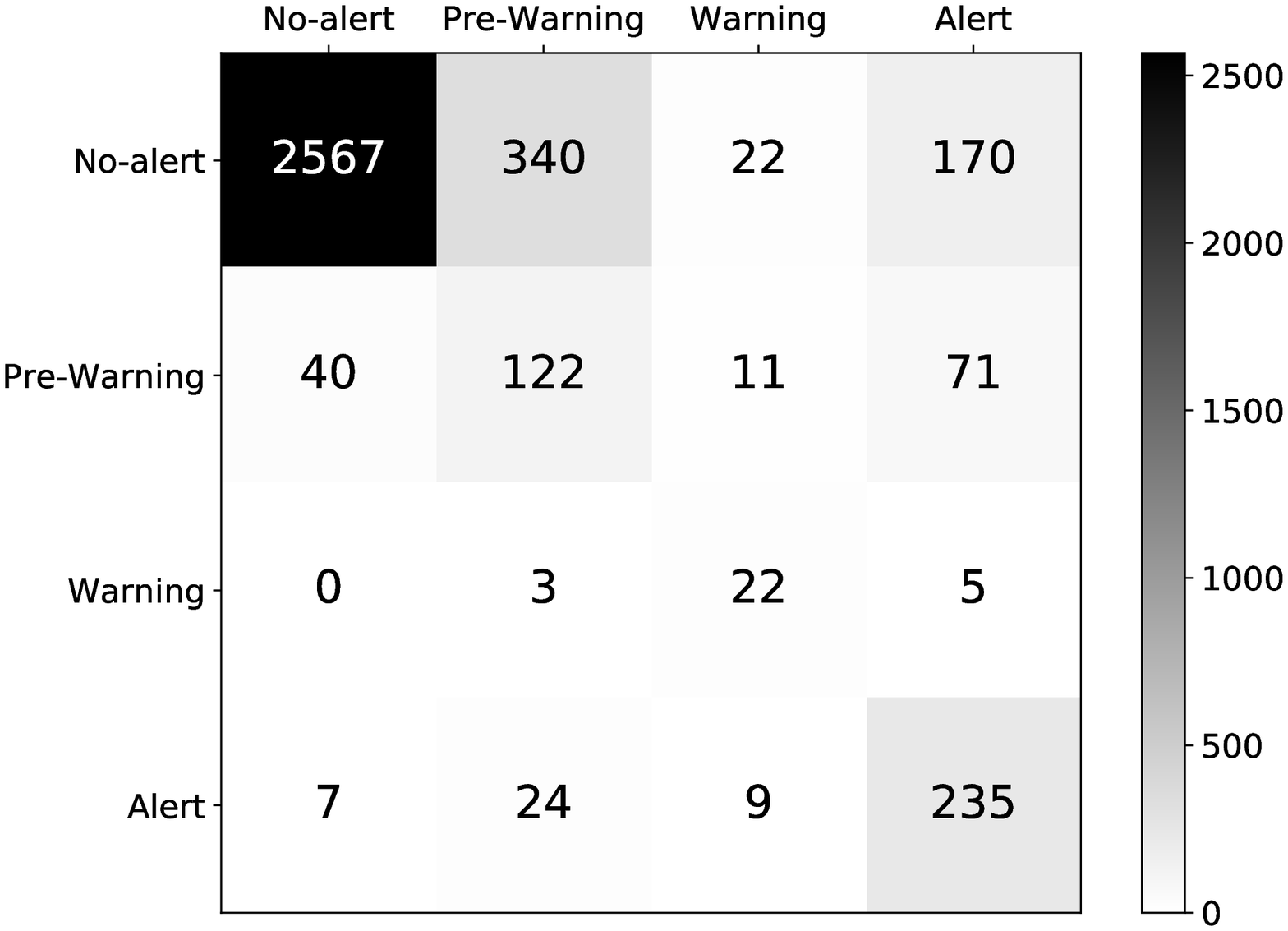}
    \label{fig:LSTM:Bloque3LSTM959075:3dia}
  }
  \subfloat[Third block. 168 hours.]{
    \includegraphics[width=0.331\textwidth]{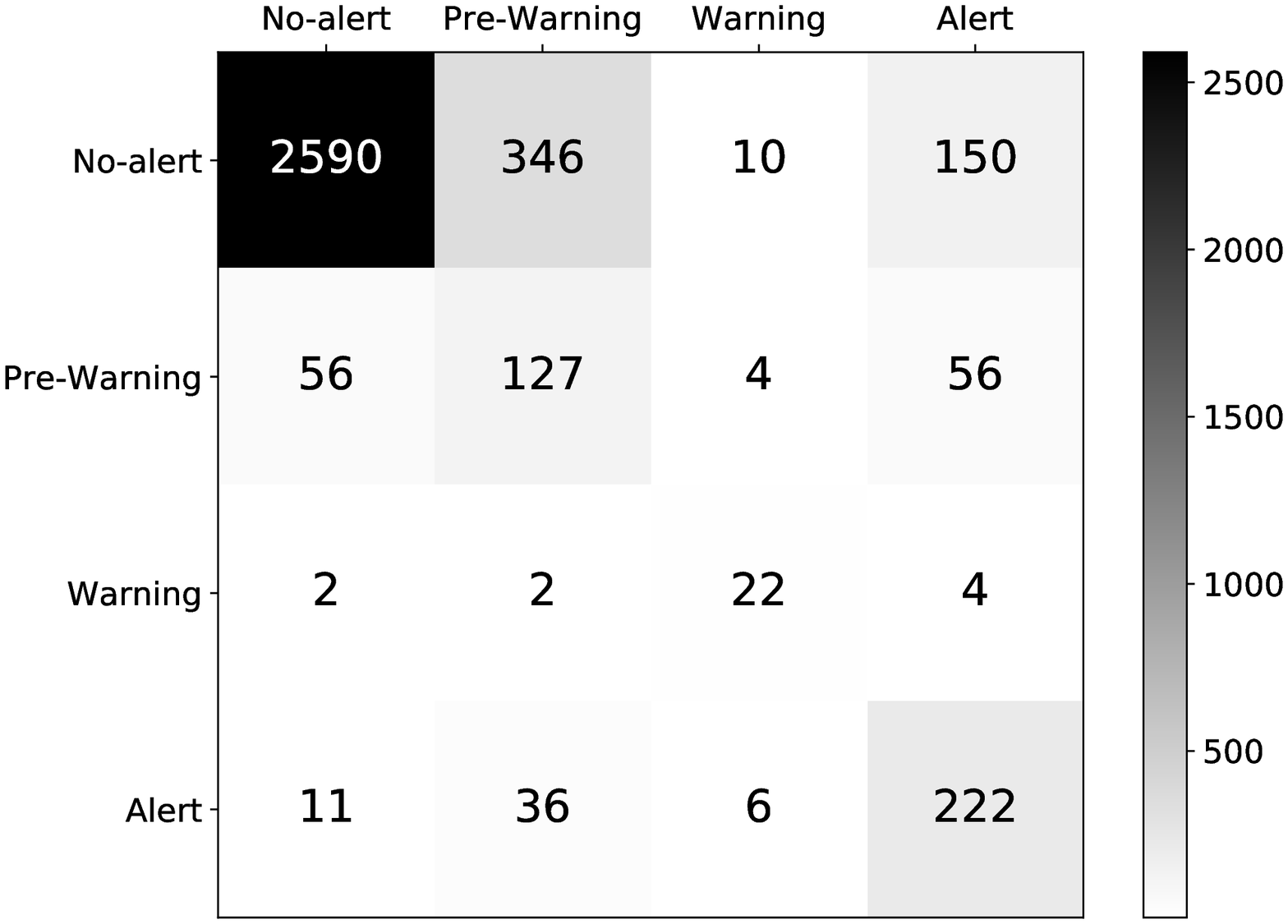}
    \label{fig:LSTM:Bloque3LSTM959075:7dia}
  }  
\caption{Confusion matrices obtained from the LSTM model for the Rule I. Three input sizes are considered: 24 hours, 72 hours and 168 hours of data per station. Two blocks are considered: the second block (6:00-12:00) and the third block (12:00-18:00). Each row represents the actual class and each column represents the predicted class.}
\label{fig:LSTM:RuleI}
\end{minipage}
}
}
\end{figure*}

\begin{figure*}
\centering
{\renewcommand{\arraystretch}{1.1}
\rotatebox{90}{
\begin{minipage}[c][][c]{\textheight}
\centering
\subfloat[Second block. 24 hours.]{
    \includegraphics[width=0.331\textwidth]{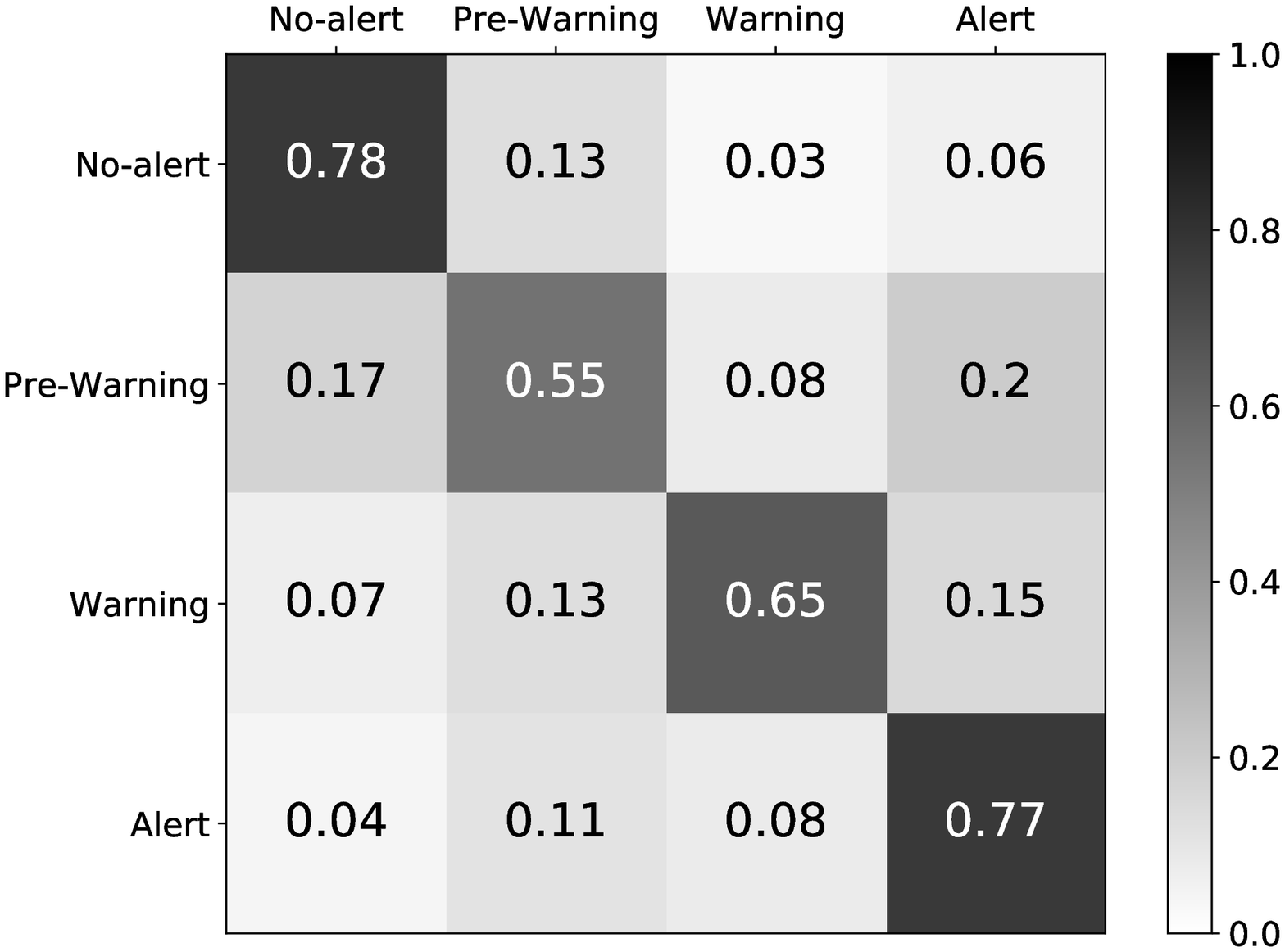}
    \label{fig:LSTMnorm:Bloque2LSTM959075:1dia}
  }
  \subfloat[Second block. 72 hours.]{
    \includegraphics[width=0.331\textwidth]{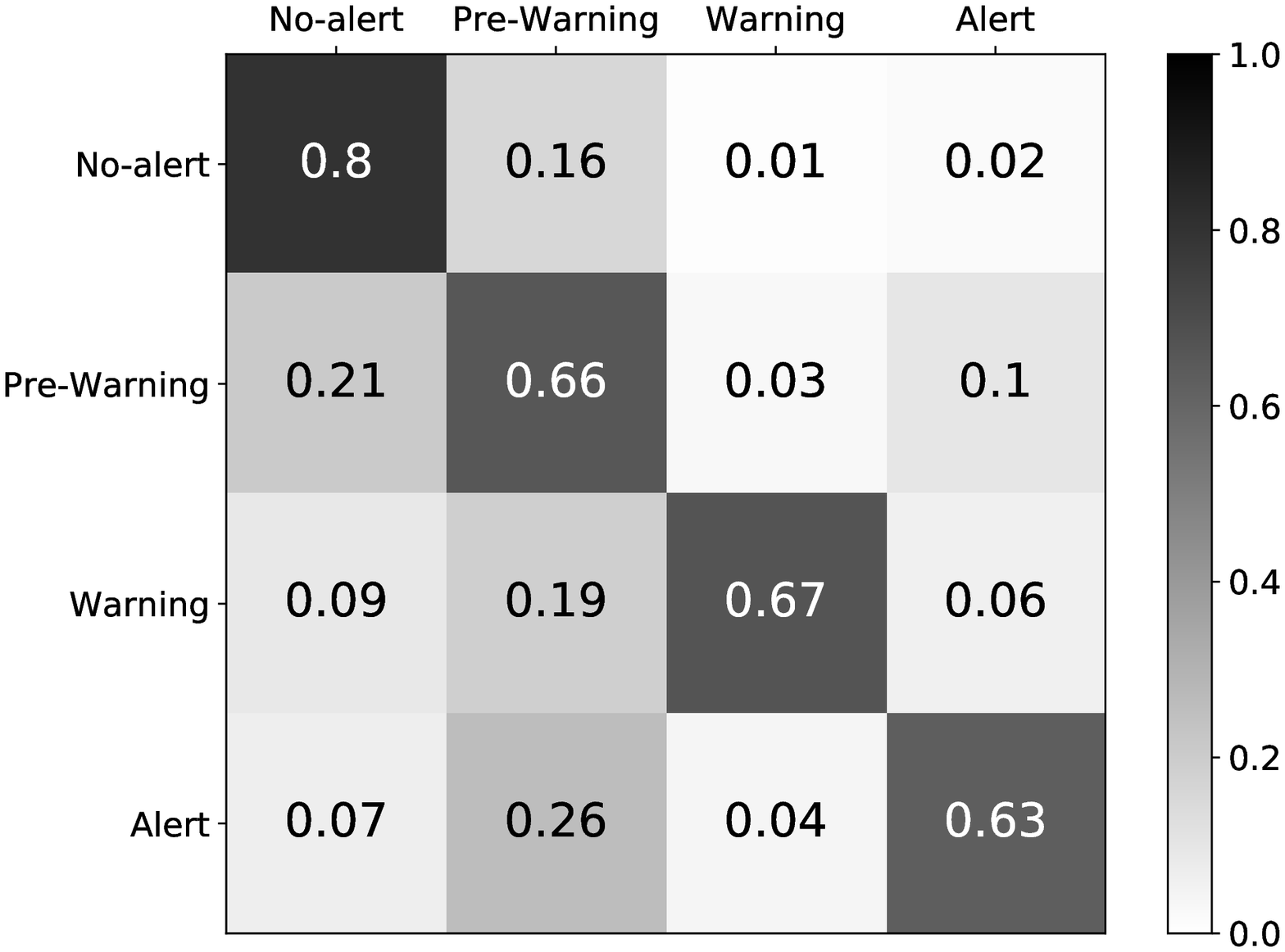}
    \label{fig:LSTMnorm:Bloque2LSTM959075:3dia}
  }
  \subfloat[Second block. 168 hours.]{
    \includegraphics[width=0.331\textwidth]{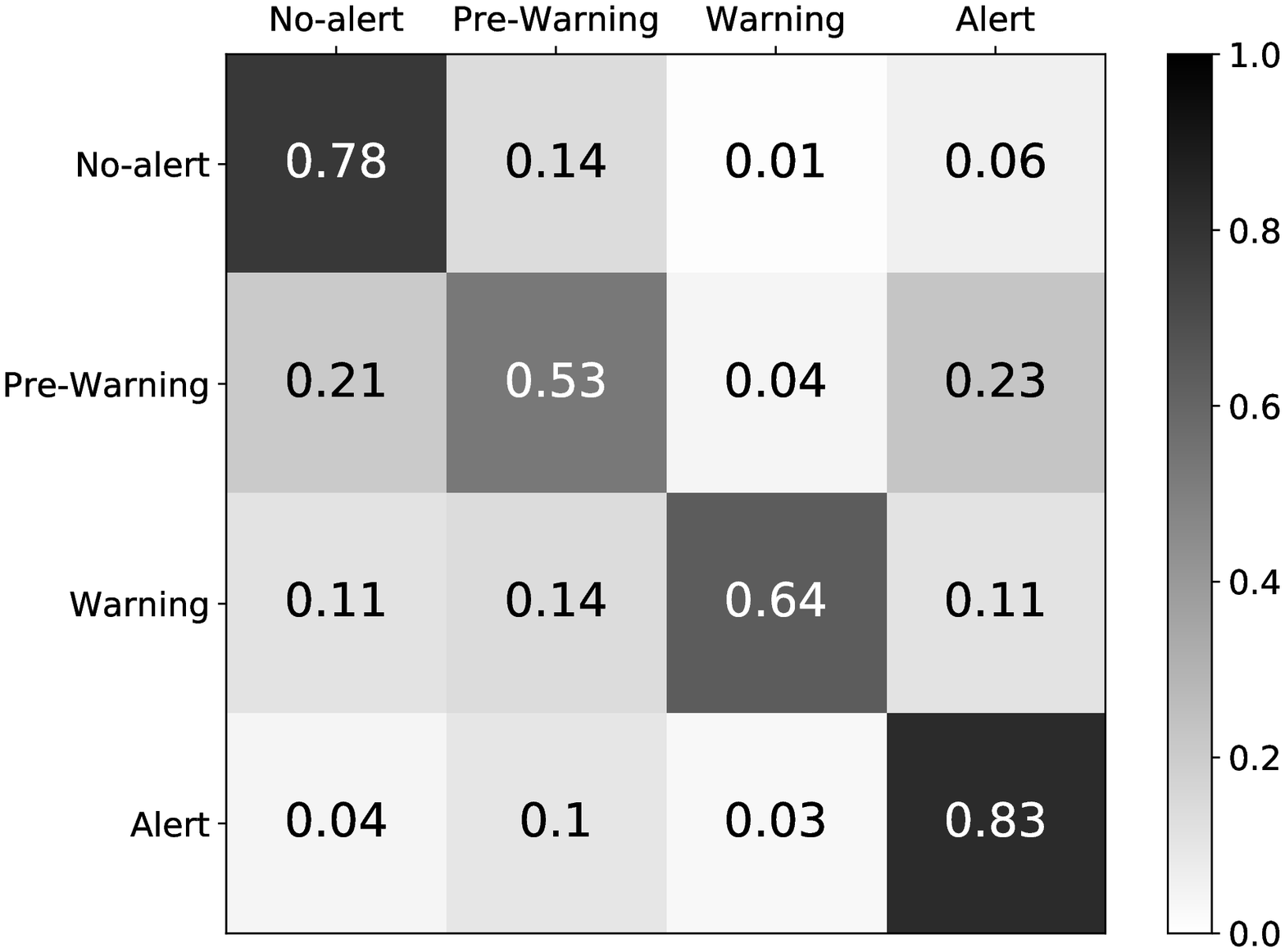}
    \label{fig:LSTMnorm:Bloque2LSTM959075:7dia}
  }\\
  \subfloat[Third block. 24 hours.]{
    \includegraphics[width=0.331\textwidth]{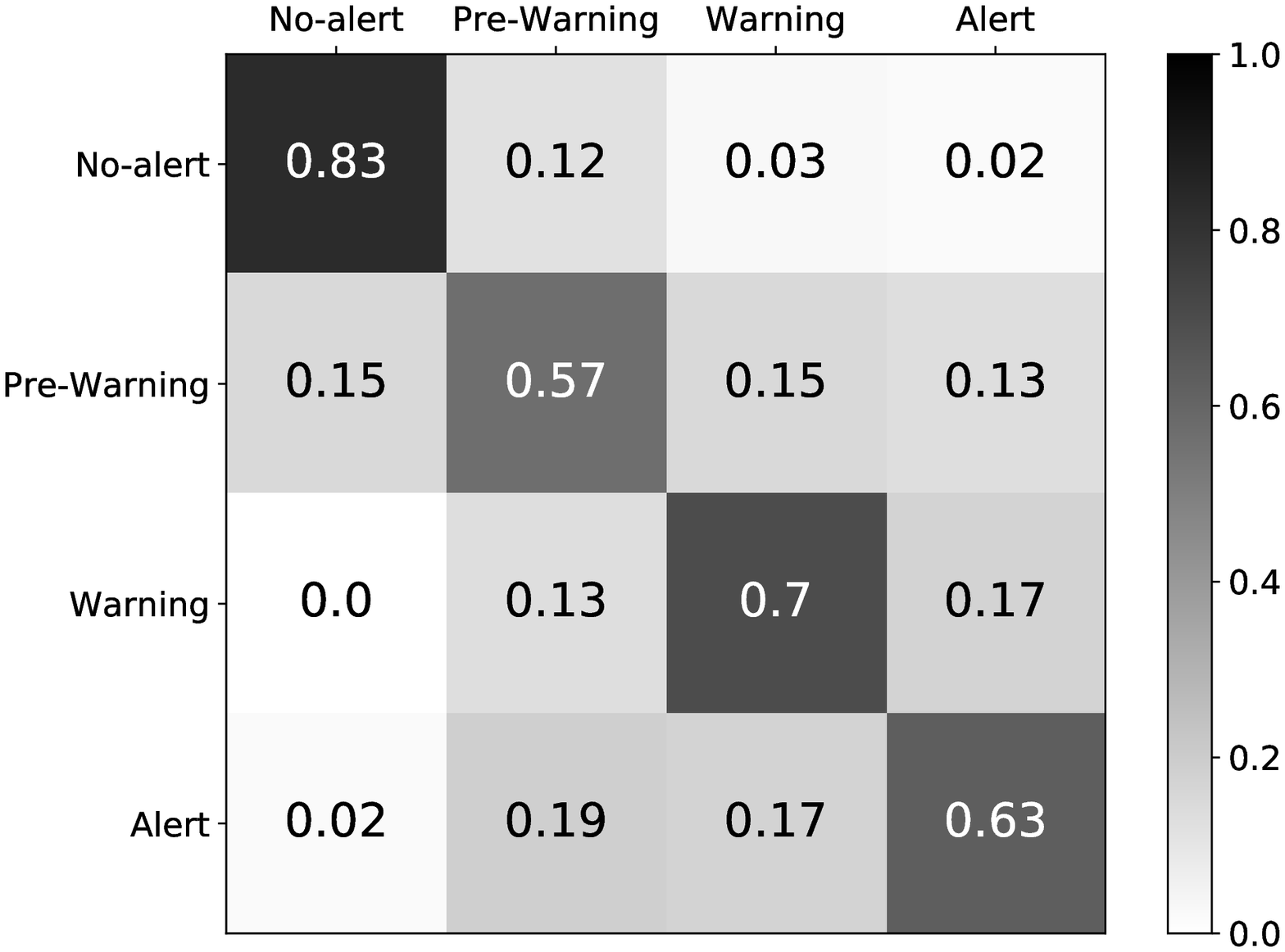}
    \label{fig:LSTMnorm:Bloque3LSTM959075:1dia}
  }
  \subfloat[Third block. 72 hours.]{
    \includegraphics[width=0.331\textwidth]{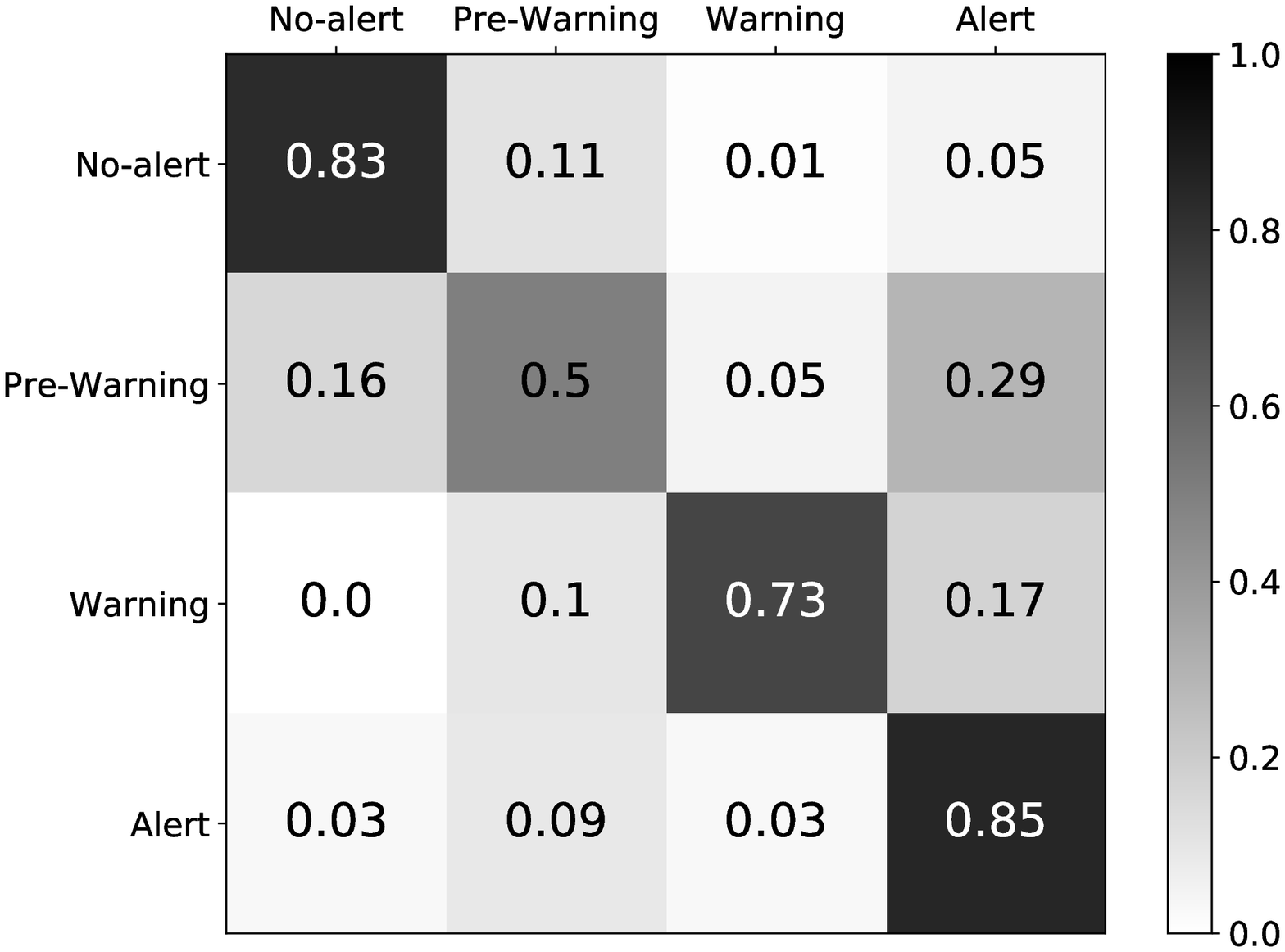}
    \label{fig:LSTMnorm:Bloque3LSTM959075:3dia}
  }
  \subfloat[Third block. 168 hours.]{
    \includegraphics[width=0.331\textwidth]{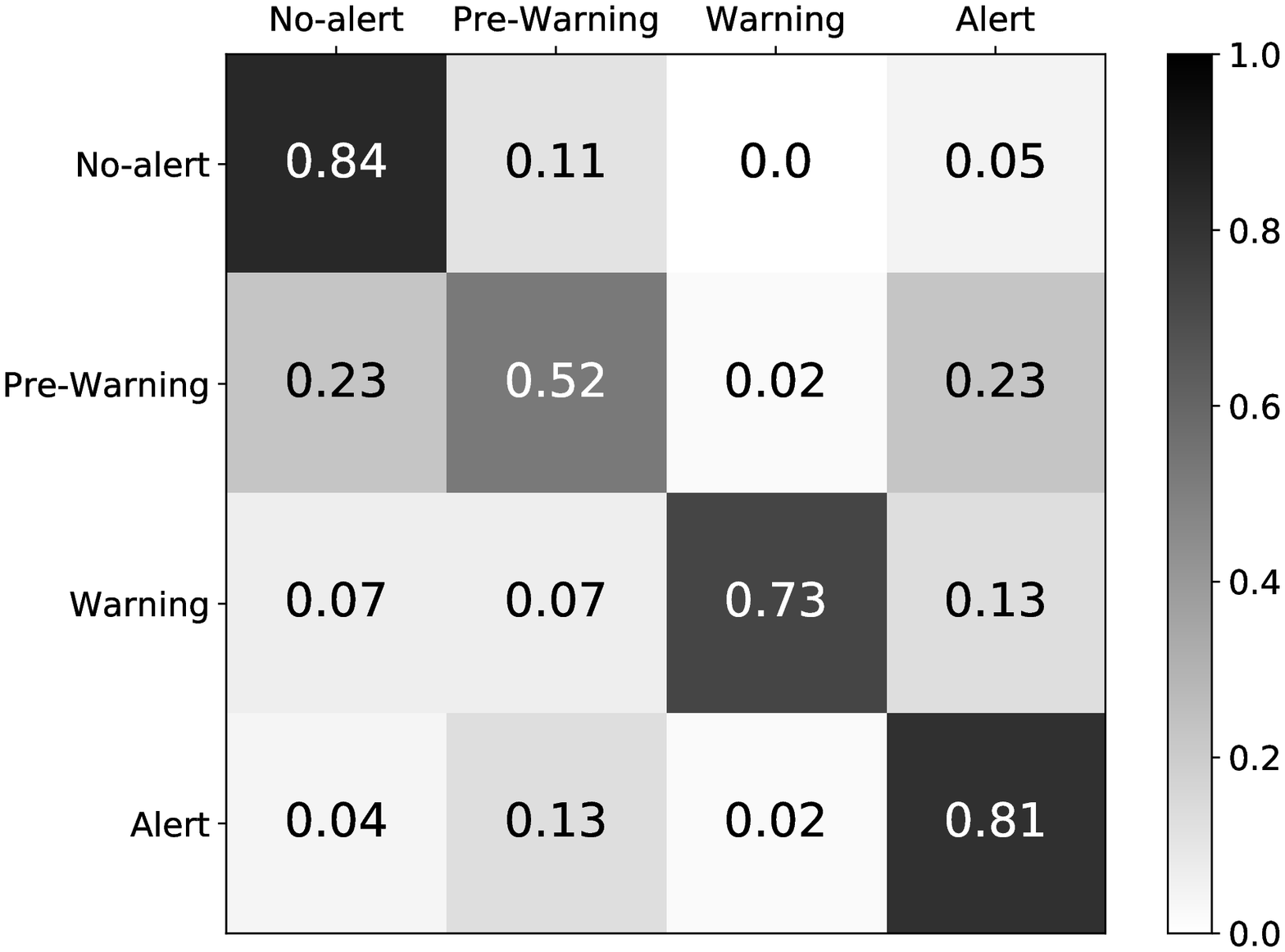}
    \label{fig:LSTMnorm:Bloque3LSTM959075:7dia}
  }
 \caption{Normalized confusion matrices obtained from the LSTM model for the Rule I. Three input sizes are considered: 24 hours, 72 hours and 168 hours of data per station. Two blocks are considered: the second block (6:00-12:00) and the third block (12:00-18:00). Each row represents the actual class and each column represents the predicted class.}
\label{fig:LSTMnorm:RuleI}
\end{minipage}
        }
}
\end{figure*}

As it is shown in Figs. \ref{fig:LSTM:Bloque2LSTM959075:1dia}, \ref{fig:LSTM:Bloque2LSTM959075:3dia}, and \ref{fig:LSTM:Bloque2LSTM959075:7dia}; and in Figs. \ref{fig:LSTMnorm:Bloque2LSTM959075:1dia}, \ref{fig:LSTMnorm:Bloque2LSTM959075:3dia}, and \ref{fig:LSTMnorm:Bloque2LSTM959075:7dia} for the normalized confusion matrices, for the Rule I and for the second block (6:00-12:00),  positive results are achieved. There is a significant difference between using 24 hours, 72 hours or 168 hours of input data, obtaining better results when using 72 hours of input data, the accuracy ranges from 63\% to 80\%; and when using 168 hours of input data, the accuracy ranges from 53\% to 83\%. In both cases, the adjacent errors ---labels erroneously predicted as a contiguous label--- and the non-adjacent errors ---labels predicted as a non-contiguous label--- are high. They rise up to 26\%.


In a similar way, as detailed in Figs. \ref{fig:LSTM:Bloque3LSTM959075:1dia}, \ref{fig:LSTM:Bloque3LSTM959075:3dia}, and \ref{fig:LSTM:Bloque3LSTM959075:7dia}; and in Figs. \ref{fig:LSTMnorm:Bloque3LSTM959075:1dia}, \ref{fig:LSTMnorm:Bloque3LSTM959075:3dia}, and \ref{fig:LSTMnorm:Bloque3LSTM959075:7dia}for the normalized confusion matrices for the Rule I and for the third block (12:00-18:00), the results are confusing, without a obvious best configuration, with low accuracy in the main diagonal, and large adjacent and non-adjacent errors. It is important to remark that for both blocks the results are not concentrated around the main diagonal, but scattered. This is probably due to the closeness of the percentiles values, and as a consequence of the imbalanced labels.

\begin{figure*}
\centering{\renewcommand{\arraystretch}{1.1}
\rotatebox{90}{
\begin{minipage}[c][][c]{\textheight}
\centering
  \subfloat[Second block. 24 hours.]{
    \includegraphics[width=0.331\textwidth]{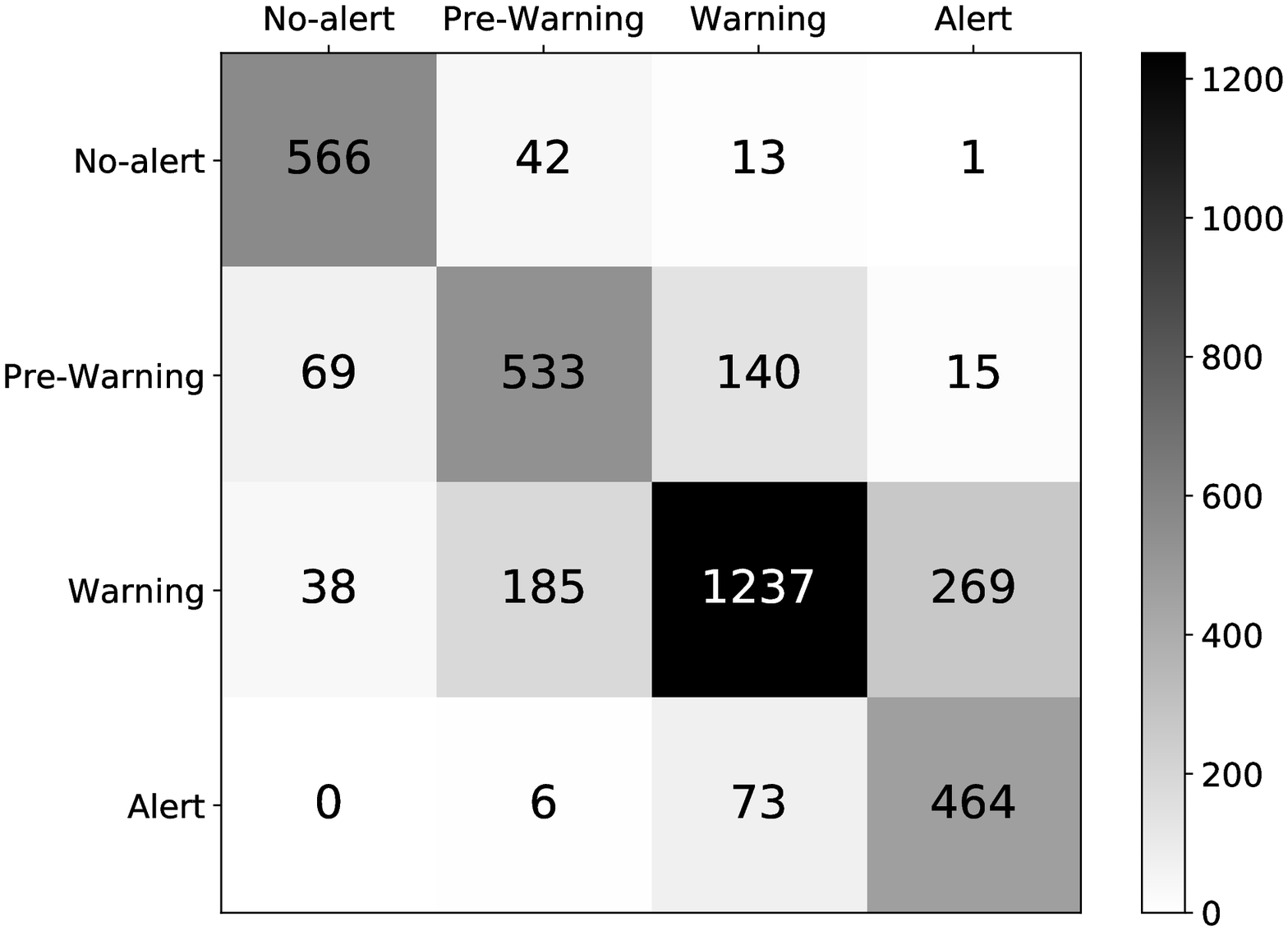}
    \label{fig:LSTM:Bloque2LSTM957550:1dia}
  }
  \subfloat[Second block. 72 hours.]{
    \includegraphics[width=0.331\textwidth]{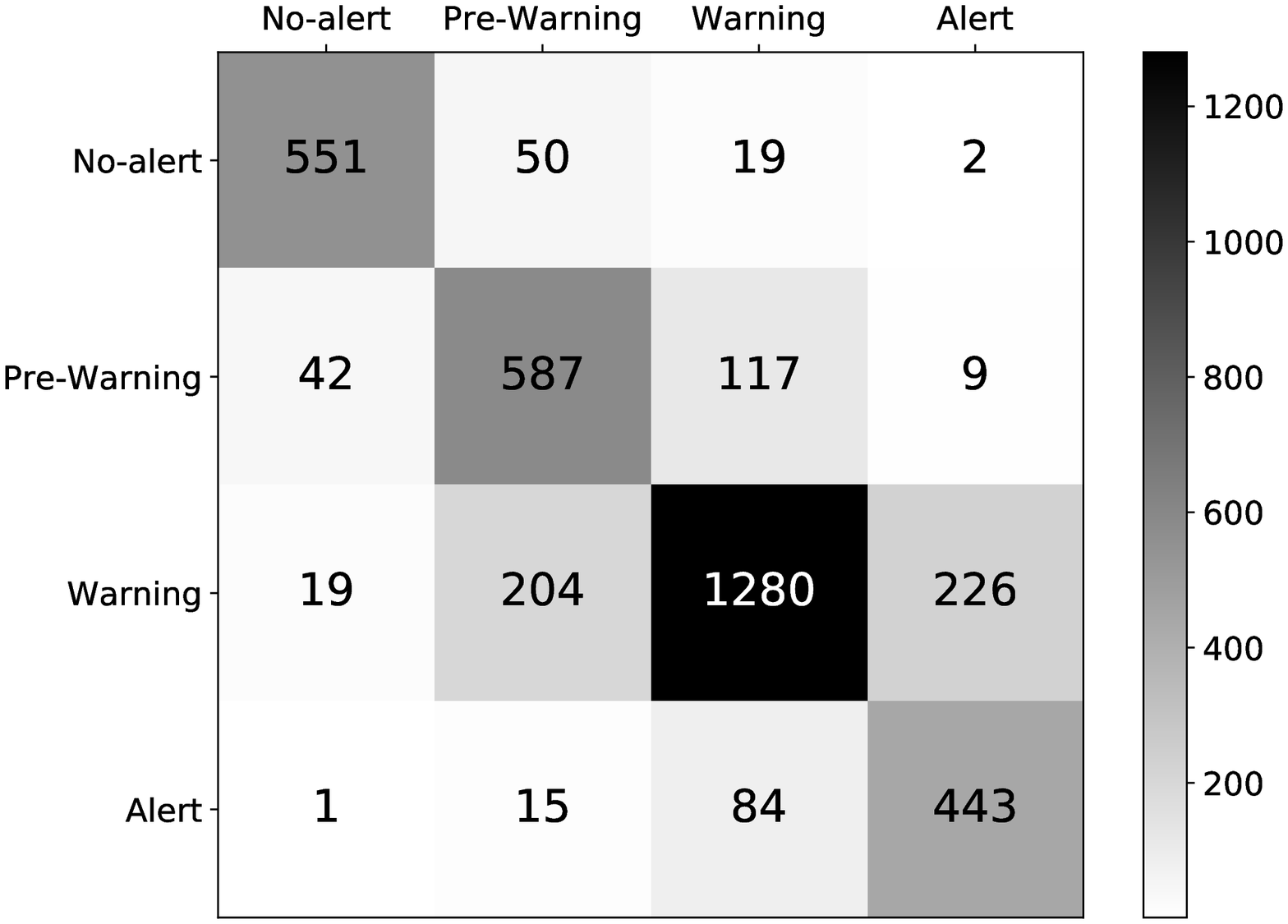}
    \label{fig:LSTM:Bloque2LSTM957550:3dia}
  }
  \subfloat[Second block. 168 hours.]{
    \includegraphics[width=0.331\textwidth]{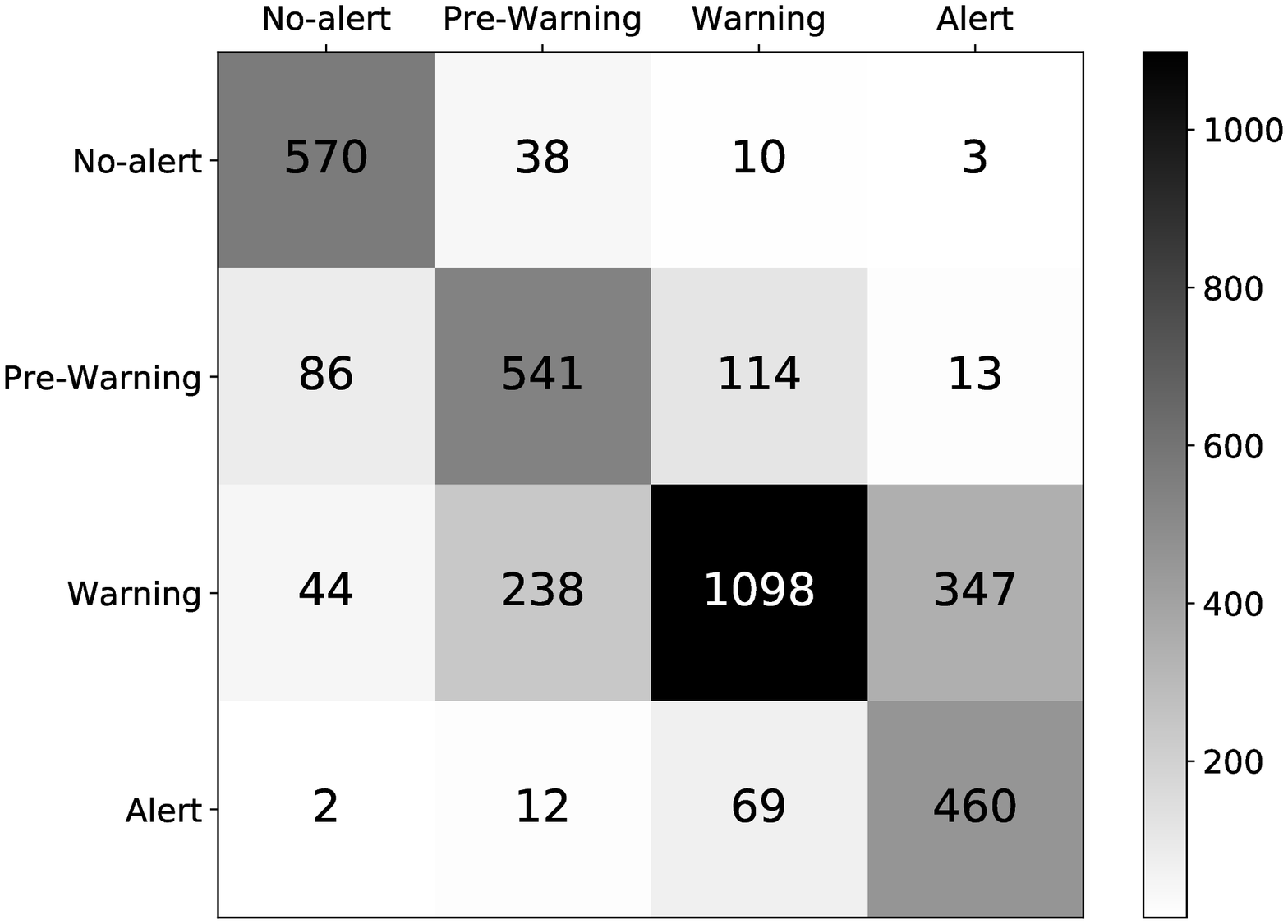}
    \label{fig:LSTM:Bloque2LSTM957550:7dia}
  }\\
  \subfloat[Third block. 24 hours.]{
    \includegraphics[width=0.331\textwidth]{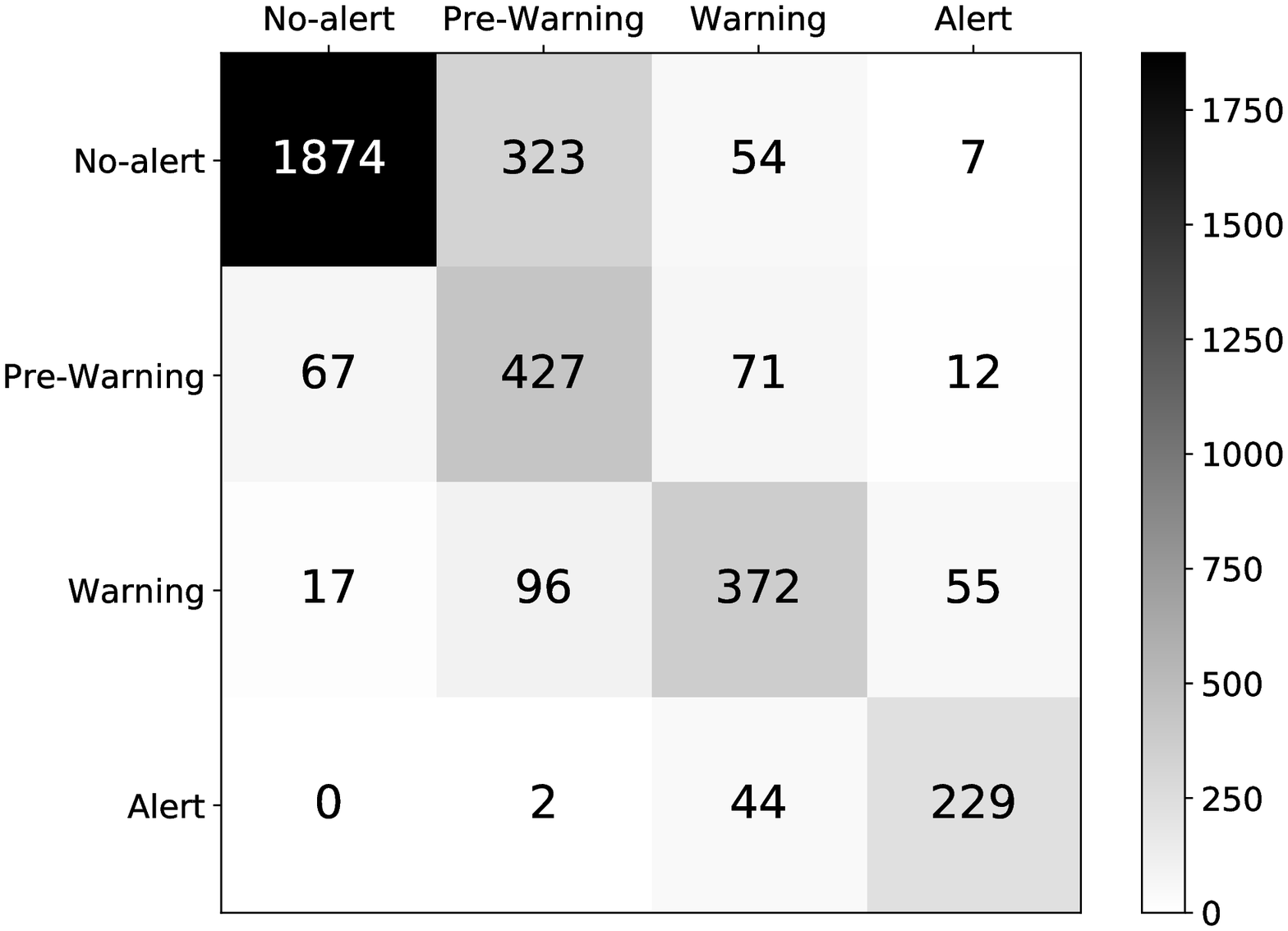}
    \label{fig:LSTM:Bloque3LSTM957550:1dia}
  }
  \subfloat[Third block. 72 hours.]{
    \includegraphics[width=0.331\textwidth]{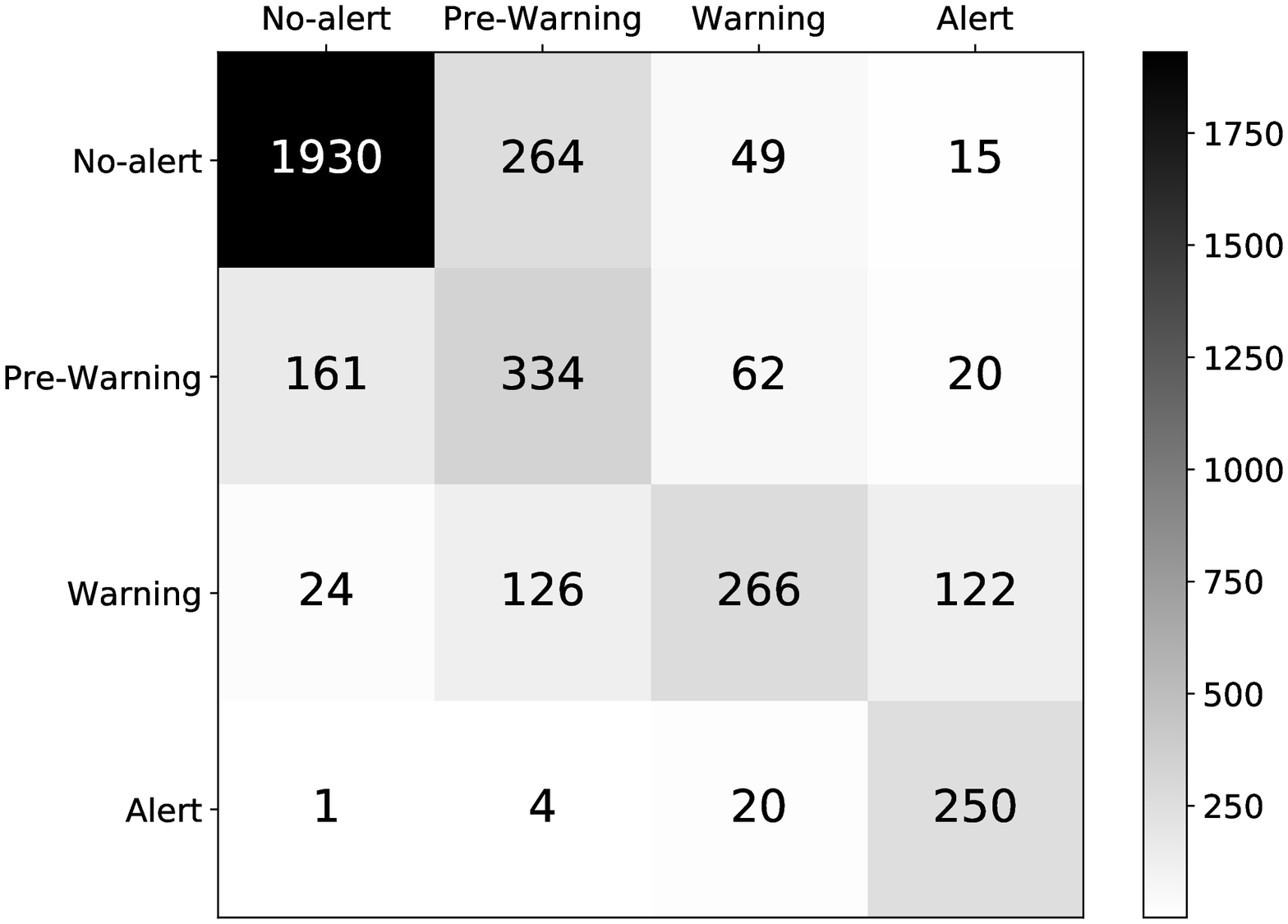}
    \label{fig:LSTM:Bloque3LSTM957550:3dia}
  }
  \subfloat[Third block. 168 hours.]{
    \includegraphics[width=0.331\textwidth]{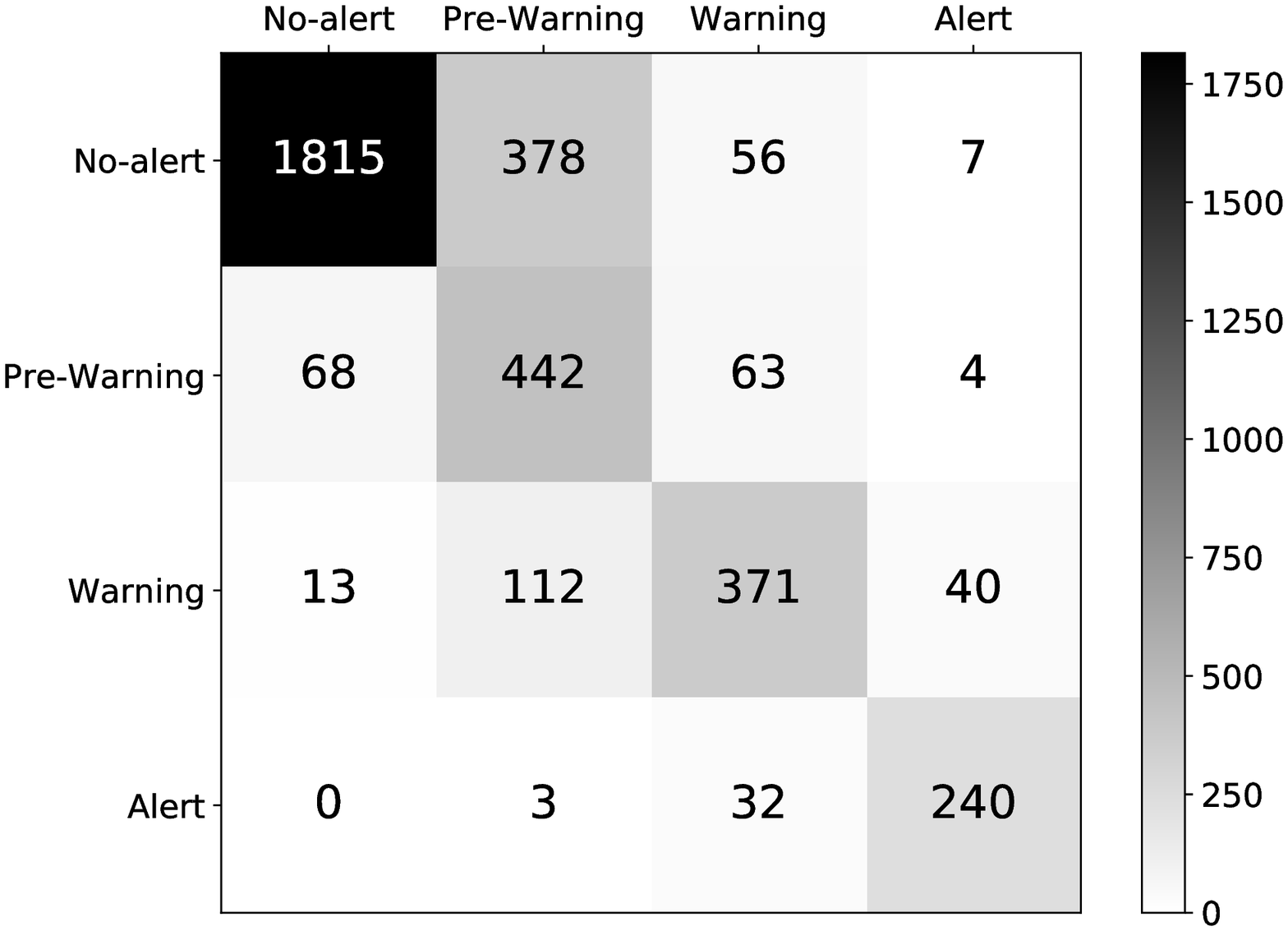}
    \label{fig:LSTM:Bloque3LSTM957550:7dia}
  }\\
\caption{Confusion matrices obtained from the LSTM model for the Rule II. Three input sizes are considered: 24 hours, 72 hours and 168 hours of data per station. Two blocks are considered: the second block (6:00-12:00) and the third block (12:00-18:00). Each row represents the actual class and each column represents the predicted class.}
\label{fig:LSTM:RuleI}
\end{minipage}
}
}
\end{figure*}

\begin{figure*}
\centering
{\renewcommand{\arraystretch}{1.1}
\rotatebox{90}{
\begin{minipage}[c][][c]{\textheight}
\centering
  \subfloat[Second block. 24 hours.]{
    \includegraphics[width=0.331\textwidth]{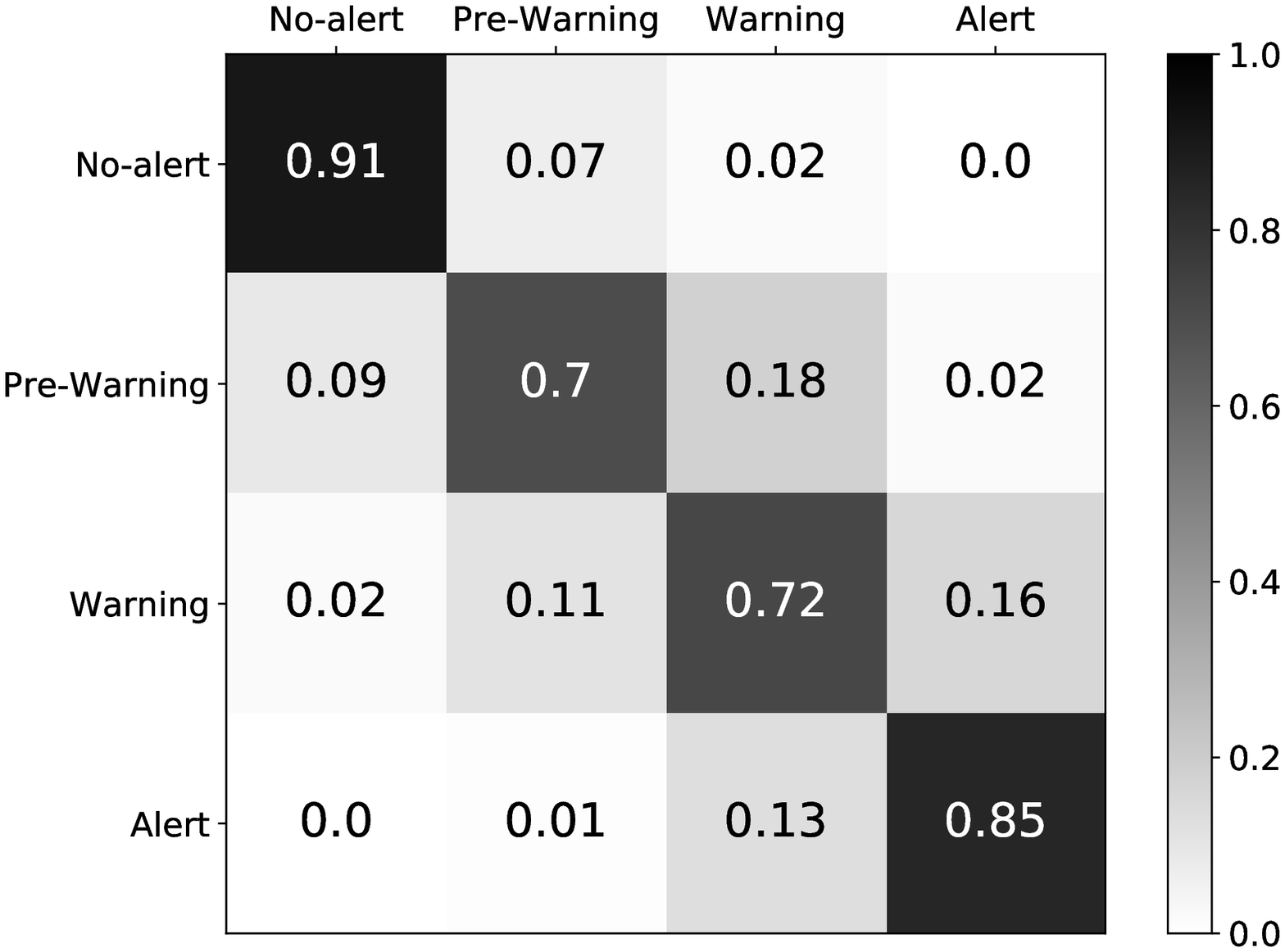}
    \label{fig:LSTMnorm:Bloque2LSTM957550:1dia}
  }
  \subfloat[Second block. 72 hours.]{
    \includegraphics[width=0.331\textwidth]{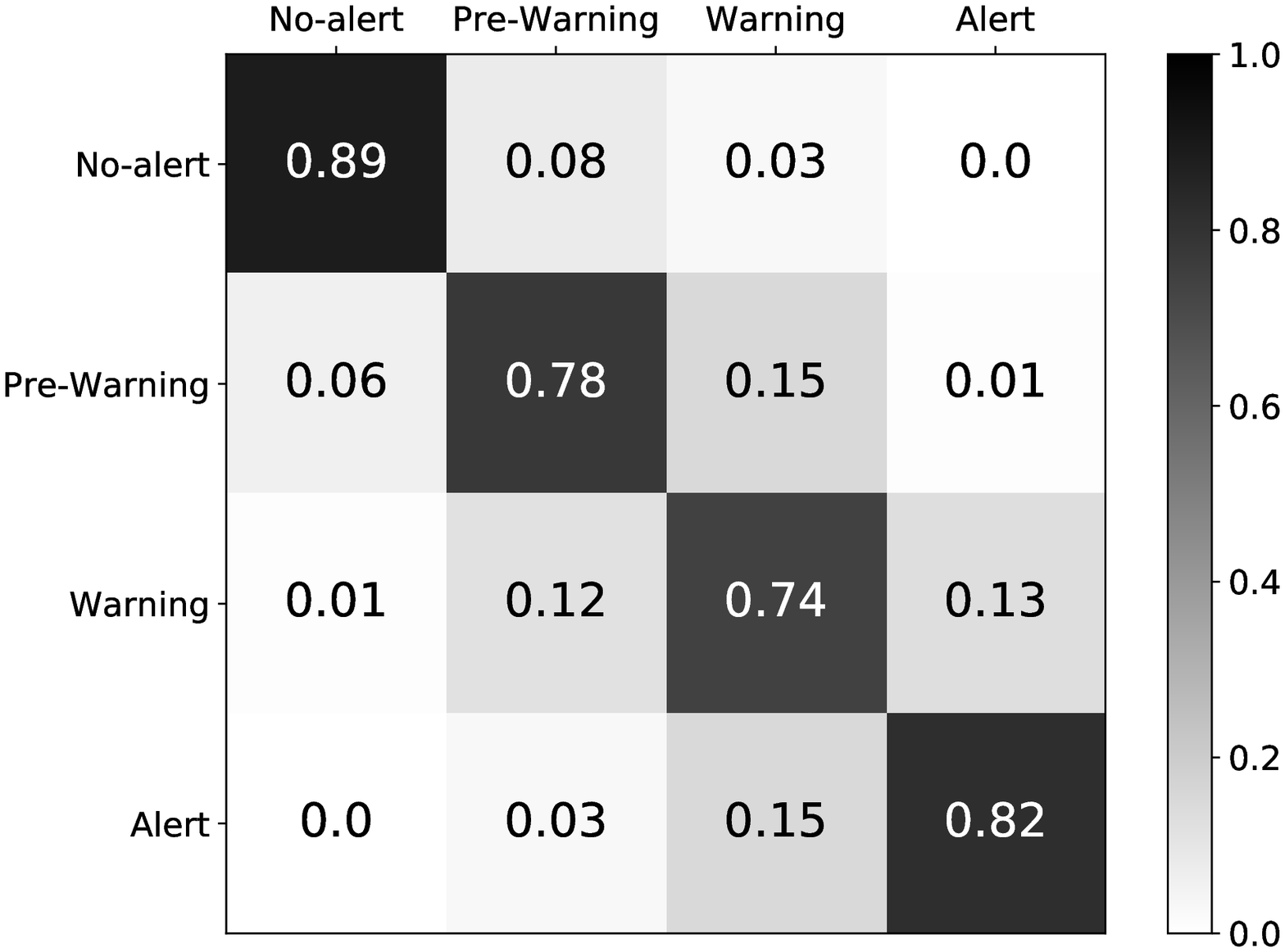}
    \label{fig:LSTMnorm:Bloque2LSTM957550:3dia}
  }
  \subfloat[Second block. 168 hours.]{
    \includegraphics[width=0.331\textwidth]{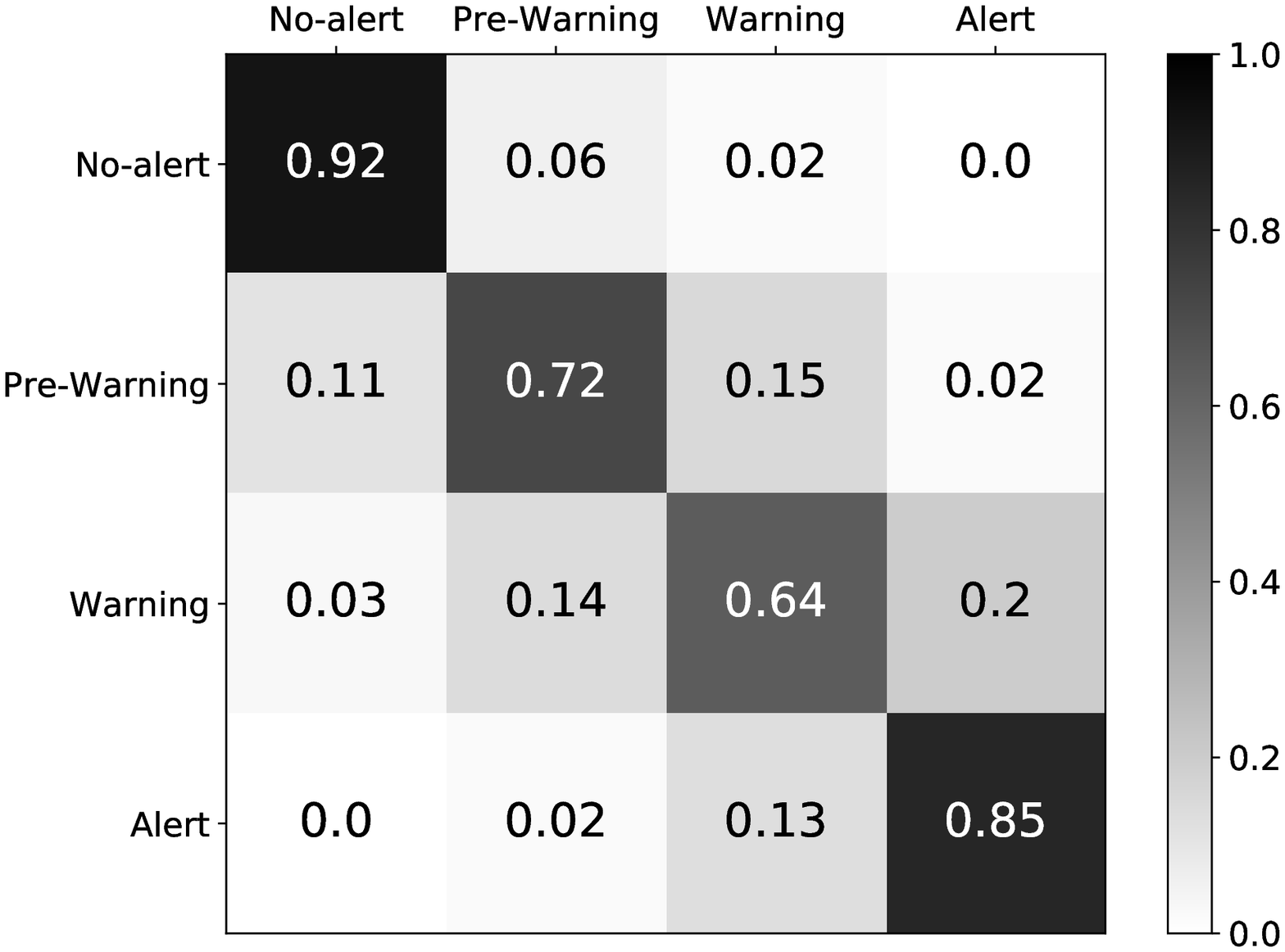}
    \label{fig:LSTMnorm:Bloque2LSTM957550:7dia}
  }\\
  \subfloat[Third block. 24 hours.]{
    \includegraphics[width=0.331\textwidth]{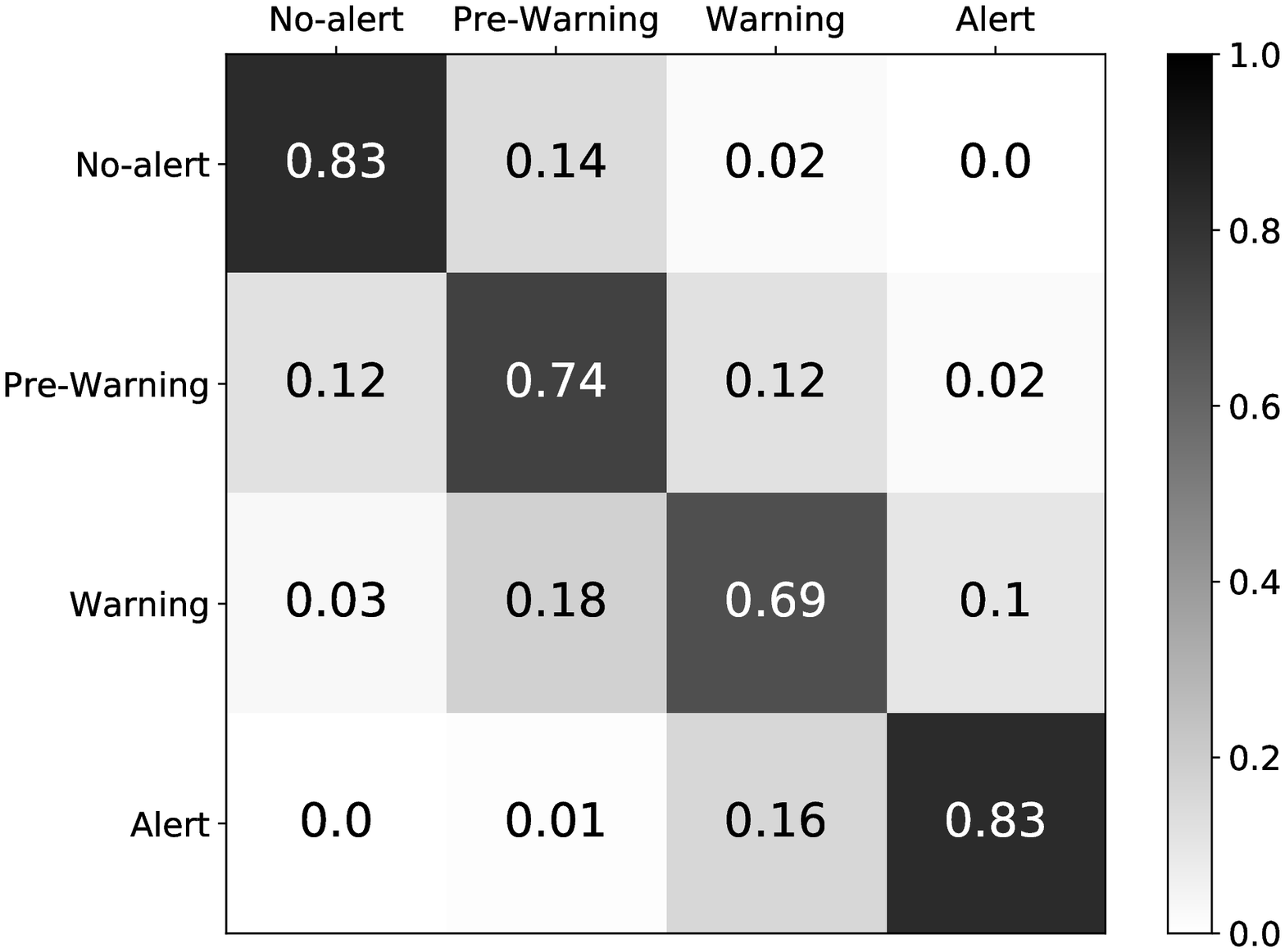}
    \label{fig:LSTMnorm:Bloque3LSTM957550:1dia}
  }
  \subfloat[Third block. 72 hours.]{
    \includegraphics[width=0.331\textwidth]{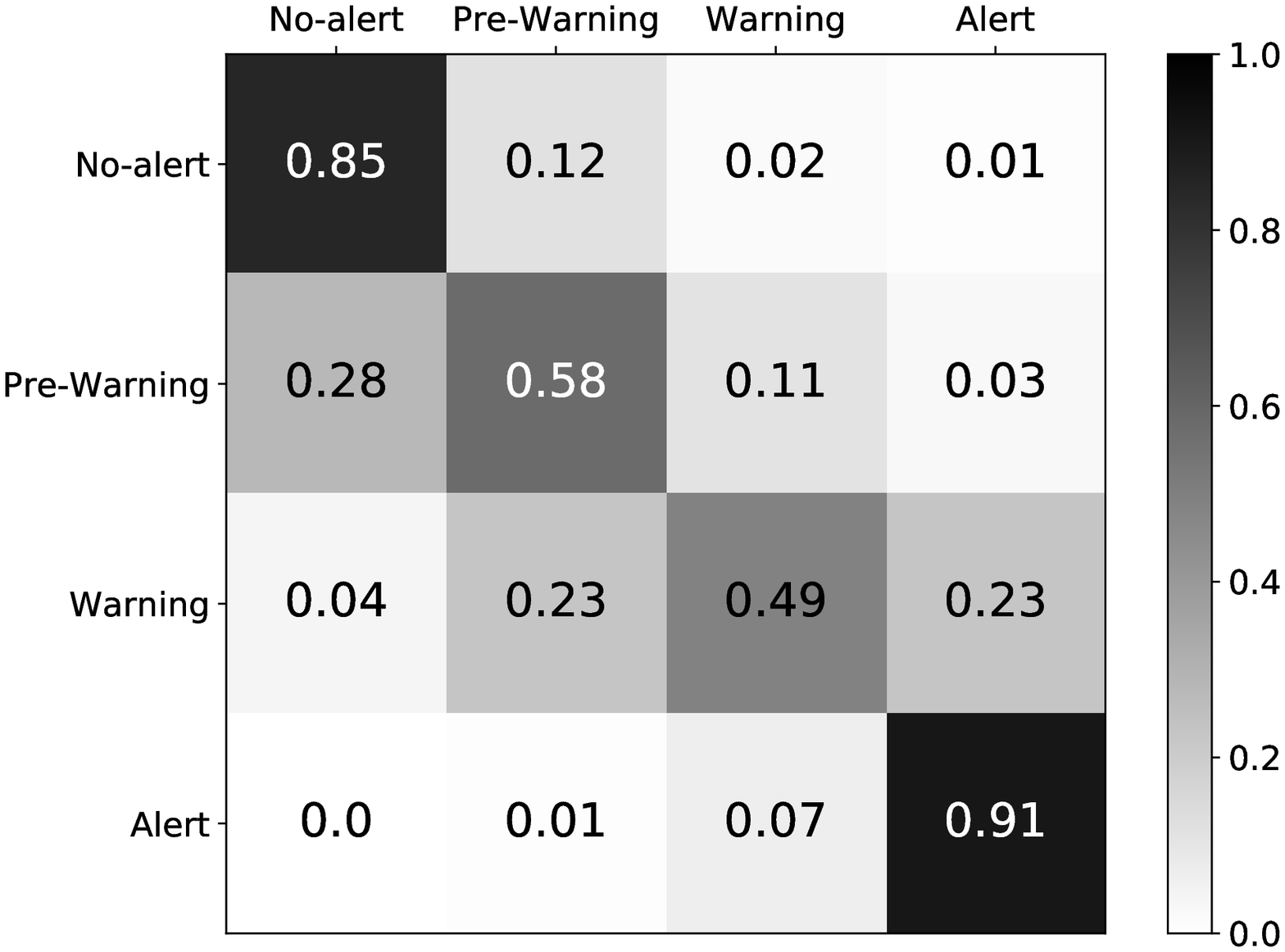}
    \label{fig:LSTMnorm:Bloque3LSTM957550:3dia}
  }
  \subfloat[Third block. 168 hours.]{
    \includegraphics[width=0.331\textwidth]{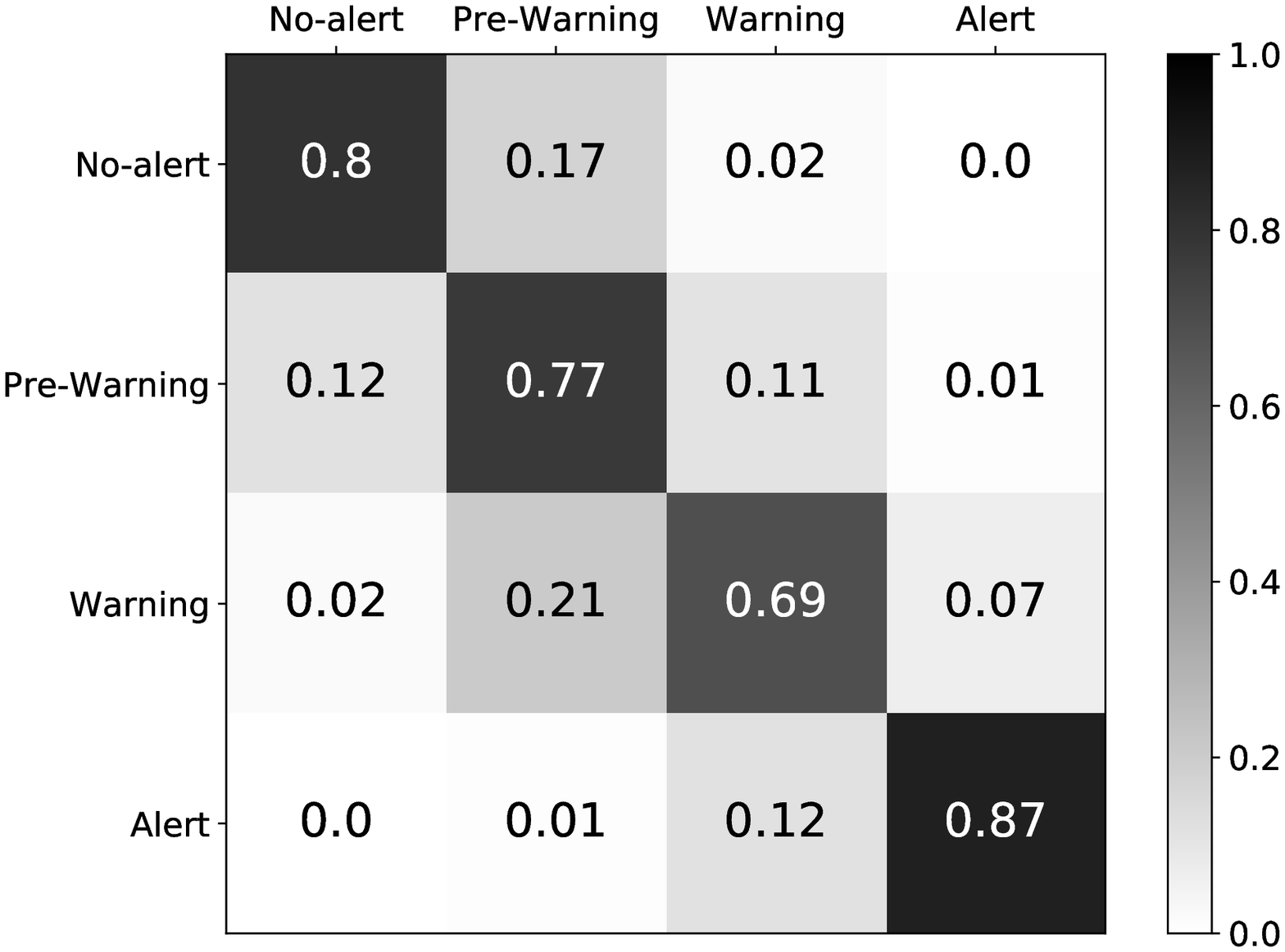}
    \label{fig:LSTMnorm:Bloque3LSTM957550:7dia}
  }
 \caption{Normalized confusion matrices obtained from the LSTM model for the  Rule II. Three input sizes are considered: 24 hours, 72 hours and 168 hours of data per station. Two blocks are considered: the second block (6:00-12:00) and the third block (12:00-18:00). Each row represents the actual class and each column represents the predicted class.}
\label{fig:LSTMnorm:RuleII}
\end{minipage}
        }
}
\end{figure*}

In the case of the pollution alert levels for the Rule II, as illustrated in Figs. \ref{fig:LSTM:Bloque2LSTM957550:1dia}, \ref{fig:LSTM:Bloque2LSTM957550:3dia}, and \ref{fig:LSTM:Bloque2LSTM957550:7dia} for the second block, and Figs. \ref{fig:LSTM:Bloque3LSTM957550:1dia}, \ref{fig:LSTM:Bloque3LSTM957550:3dia}, and \ref{fig:LSTM:Bloque3LSTM957550:7dia} for the third block; and for the normalized confusion matrices in Figs. \ref{fig:LSTMnorm:Bloque2LSTM957550:1dia}, \ref{fig:LSTMnorm:Bloque2LSTM957550:3dia}, and \ref{fig:LSTMnorm:Bloque2LSTM957550:7dia} for the second block, and Figs. \ref{fig:LSTMnorm:Bloque3LSTM957550:1dia}, \ref{fig:LSTMnorm:Bloque3LSTM957550:3dia}, and \ref{fig:LSTMnorm:Bloque3LSTM957550:7dia} for the third block, the results for both blocks are better than those obtained for Rule I. This stems from the less imbalance labels under the Rule II percentiles. For the second block and the third block, it is difficult to ascertain which configuration produces the best results.  
For the second block, the accurate results ---main diagonal--- range from 74\% to 89\% for the 72 hours configuration to the range from 64\% to 92\% for the 128 hours configuration. 
For the third block, the accurate results overlap: from 69\% to 83\% when using the configuration of 24 hours, from 49\% to 91\% when using the configuration of 72 hours, and from 69\% to 87\% when using the configuration of 168 hours. At the same time large adjacent errors are also produced, for instance the 28\% of Pre-Warning are incorrectly predicted as No alert when using the configuration of 72 hours.


\subsubsection{CNN}
\begin{figure*}
\centering
{\renewcommand{\arraystretch}{1.1}
\rotatebox{90}{
\begin{minipage}[c][][c]{\textheight}
\centering
  \subfloat[Second block. 24 hours.]{
    \includegraphics[ width=0.331\textwidth]{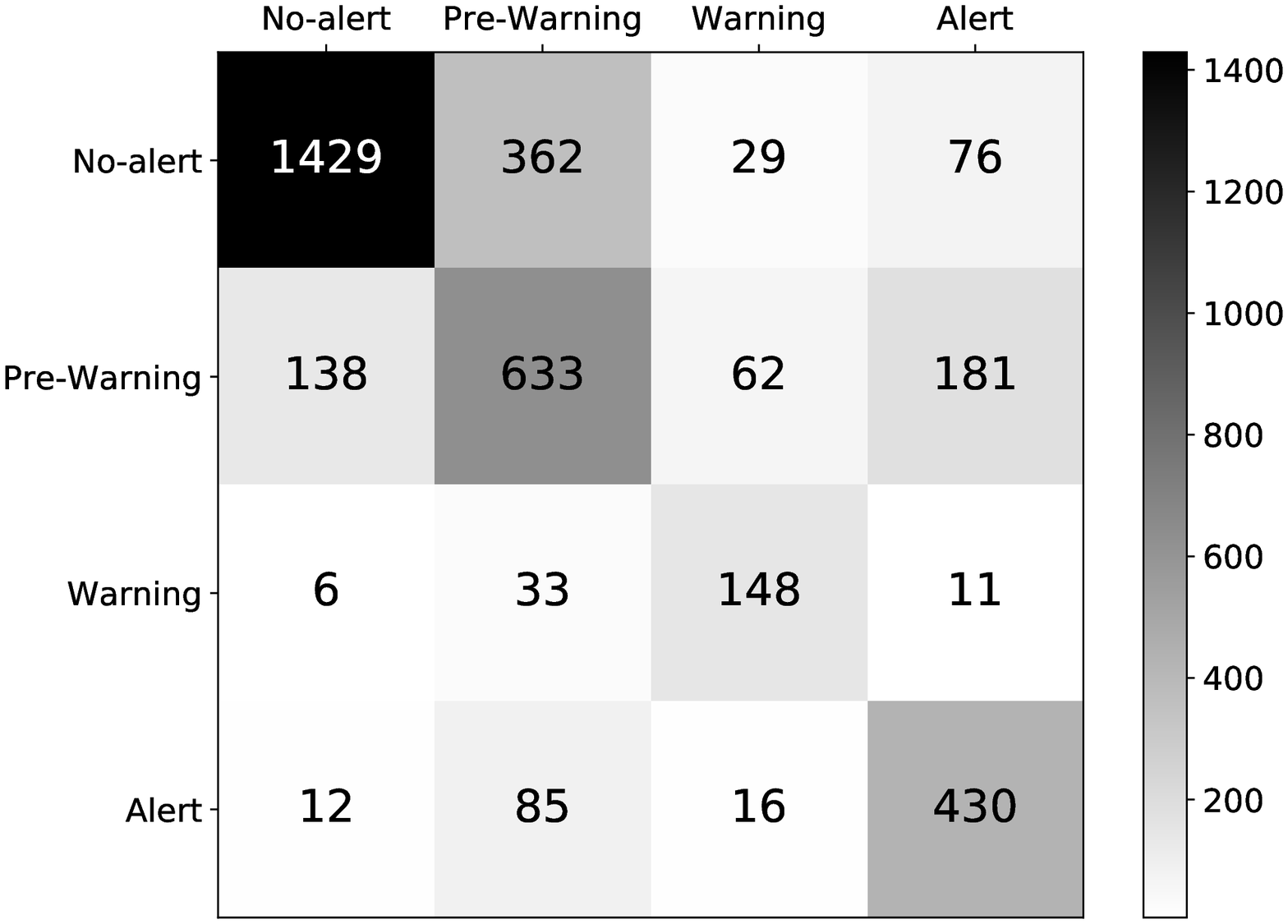}
    \label{fig:CNN:Bloque2CNN959075:1dia}
  }
  \subfloat[Second block. 72 hours.]{
    \includegraphics[ width=0.331\textwidth]{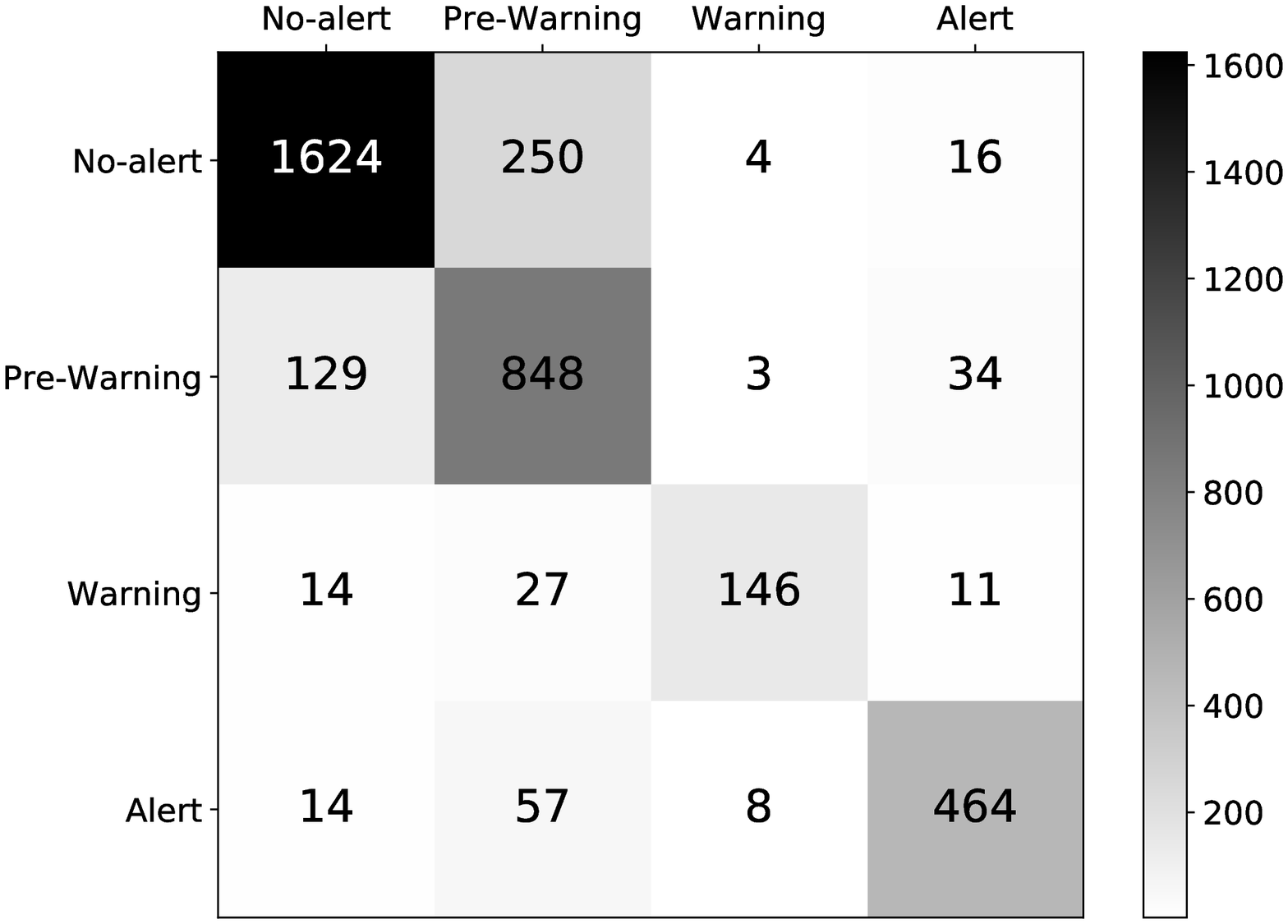}
    \label{fig:CNN:Bloque2CNN959075:3dia}
  }
  \subfloat[Second block. 168 hours.]{
    \includegraphics[ width=0.331\textwidth]{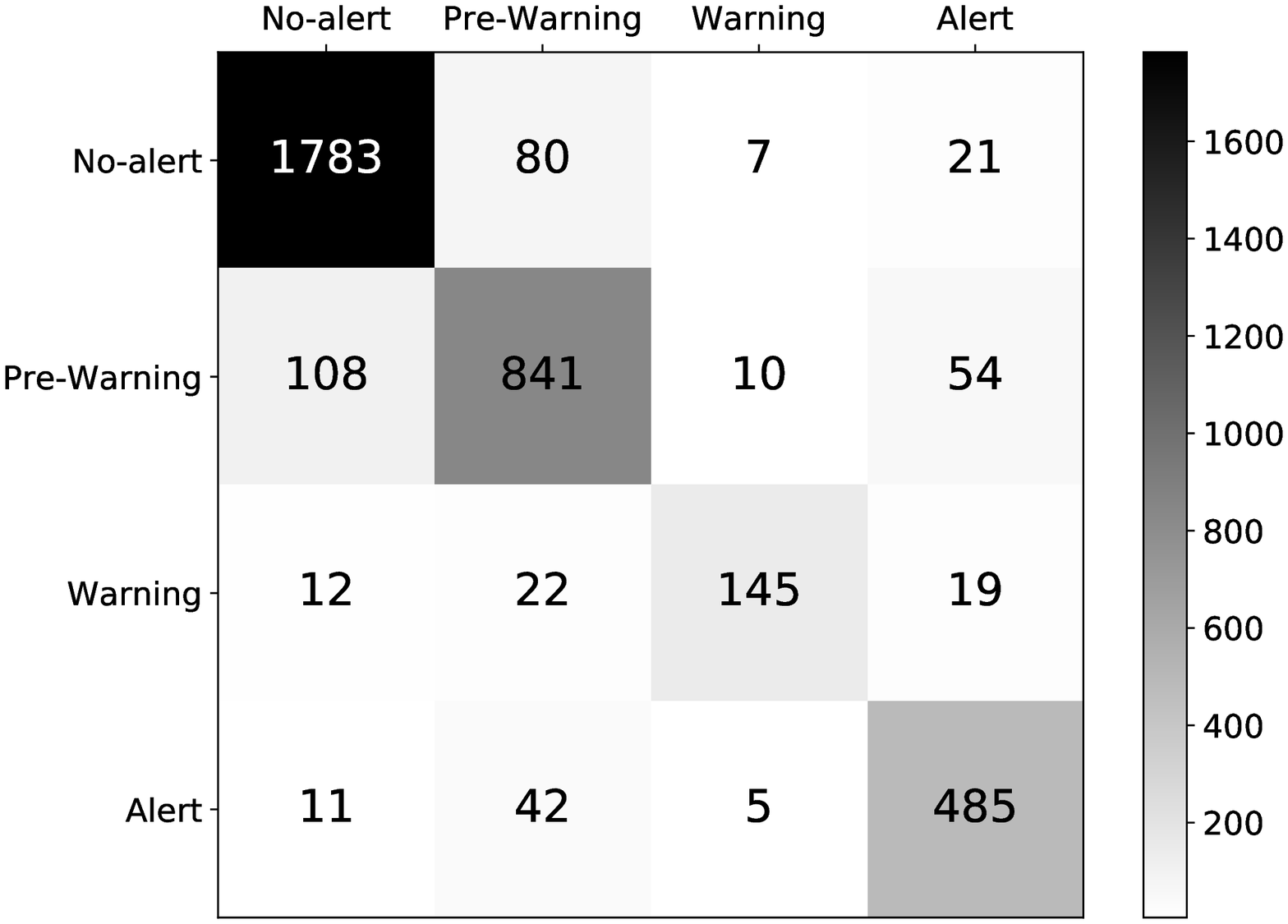}
    \label{fig:CNN:Bloque2CNN959075:7dia}
  }\\
  \subfloat[Third block. 24 hours.]{
    \includegraphics[ width=0.331\textwidth]{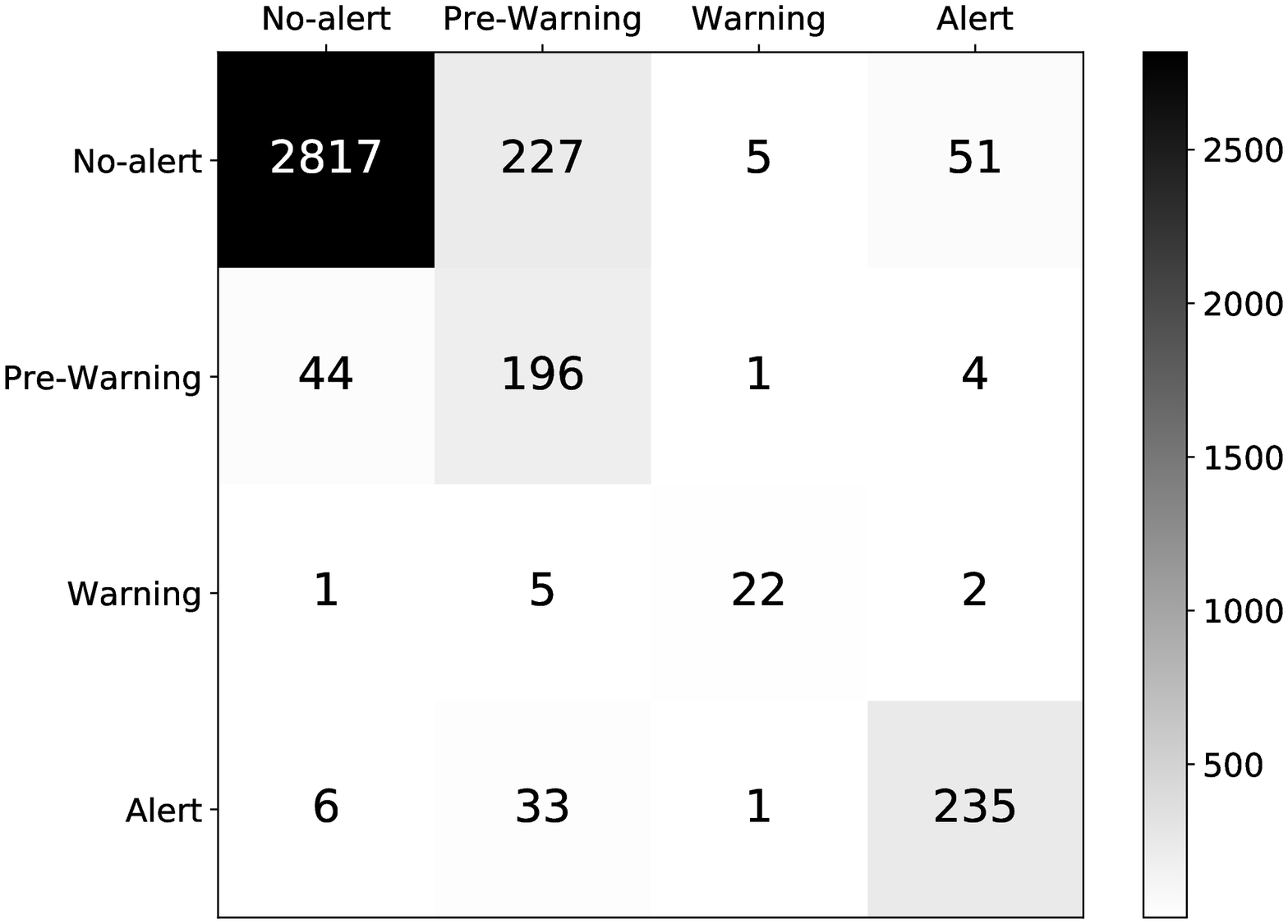}
    \label{fig:CNN:Bloque3CNN959075:1dia}
  }
  \subfloat[Third block. 72 hours.]{
    \includegraphics[ width=0.331\textwidth]{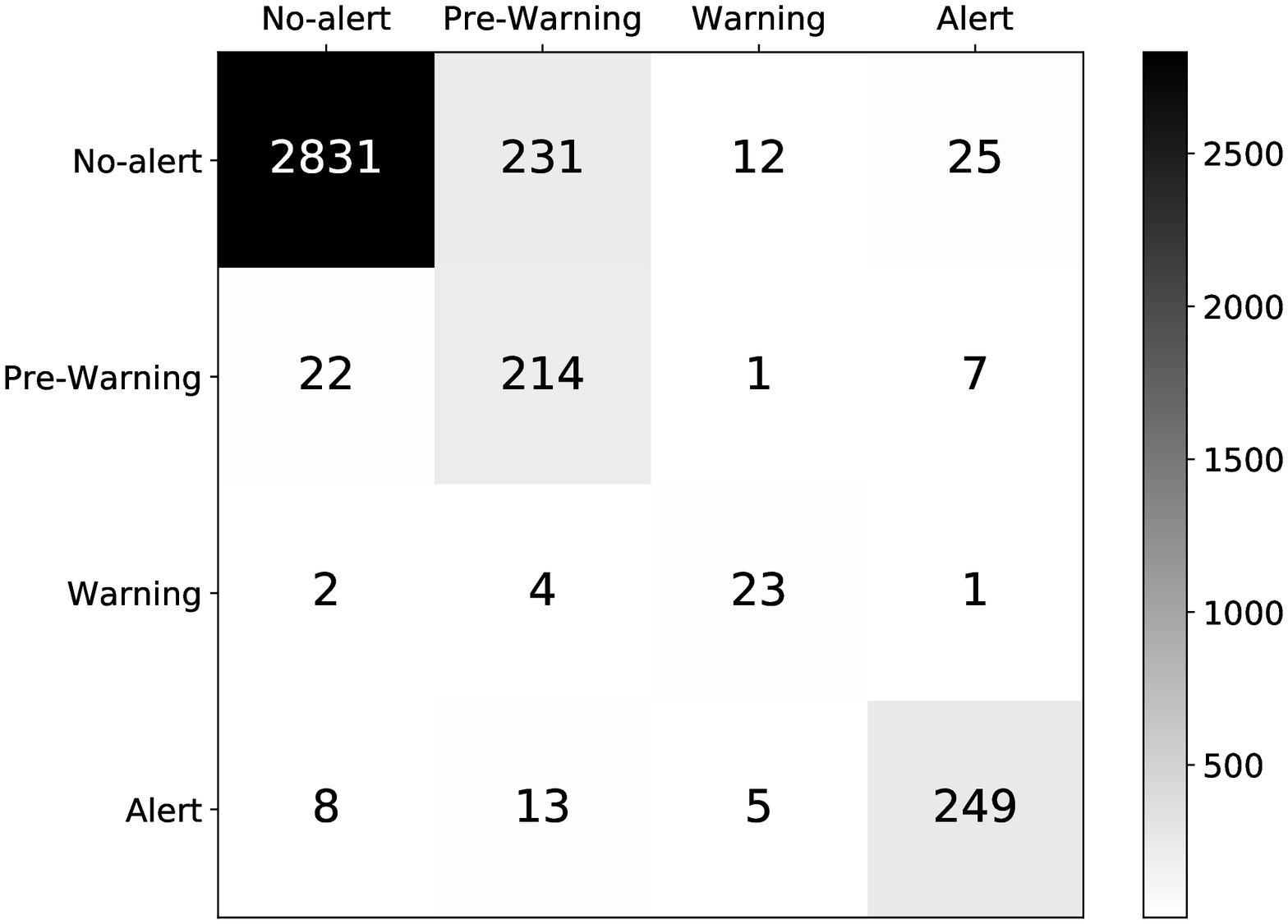}
    \label{fig:CNN:Bloque3CNN959075:3dia}
  }
  \subfloat[Third block. 168 hours.]{
    \includegraphics[ width=0.331\textwidth]{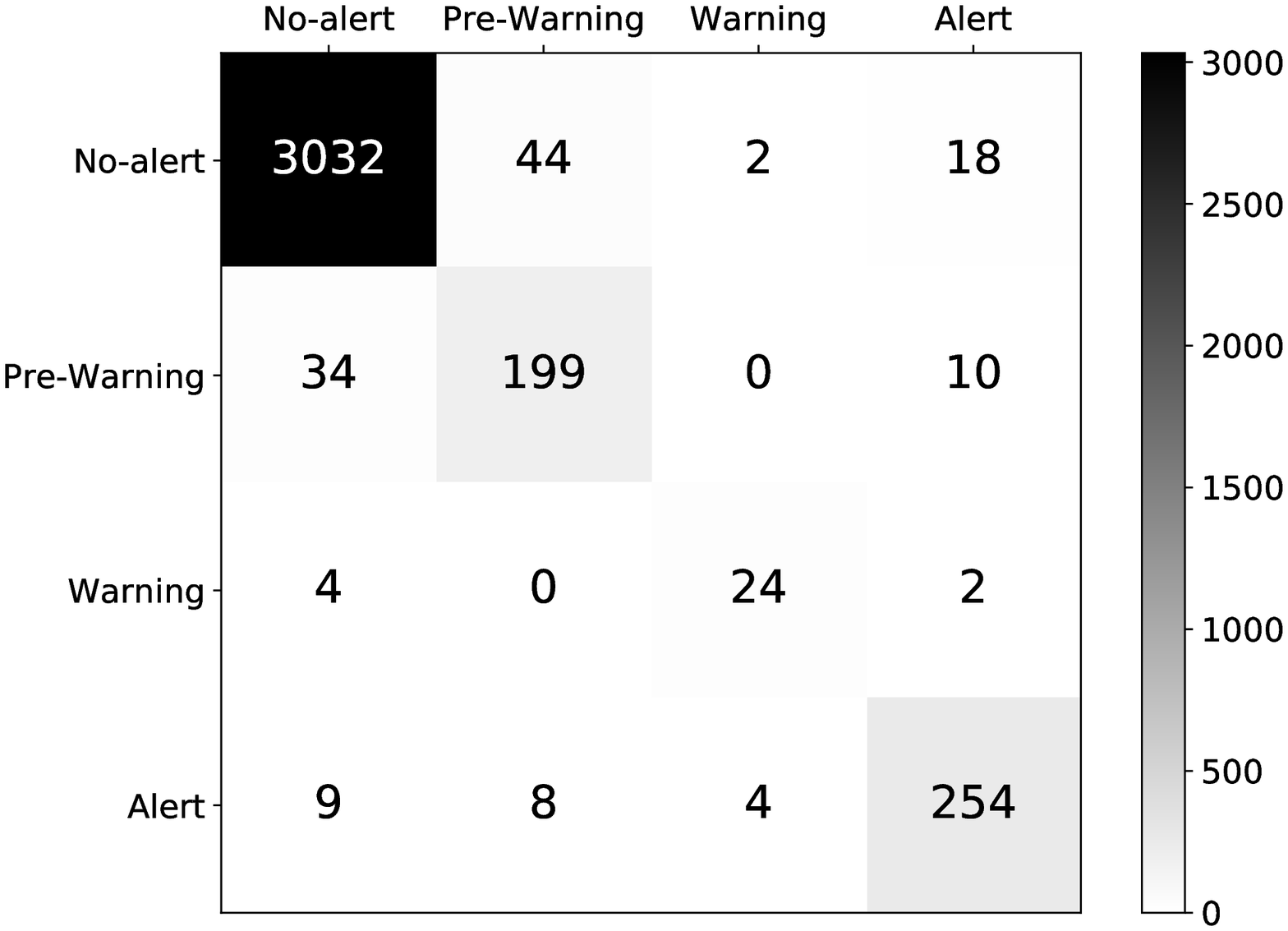}
    \label{fig:CNN:Bloque3CNN959075:7dia}
  }
\caption{Confusion matrices obtained from the CNN model for the Rule I. Three input sizes are considered: 24 hours, 72 hours and 168 hours of data per station. Two blocks are considered: the second block (6:00-12:00) and the third block (12:00-18:00). Each row represents the actual class and each column represents the predicted class.}
\label{fig:CNN:Rule_I}
\end{minipage}
 }
}
\end{figure*}

\begin{figure}
\centering
{\renewcommand{\arraystretch}{1.1}
\rotatebox{90}{
\begin{minipage}[c][][c]{\textheight}
\centering
  \subfloat[Second block. 24 hours.]{
    \includegraphics[ width=0.331\textwidth]{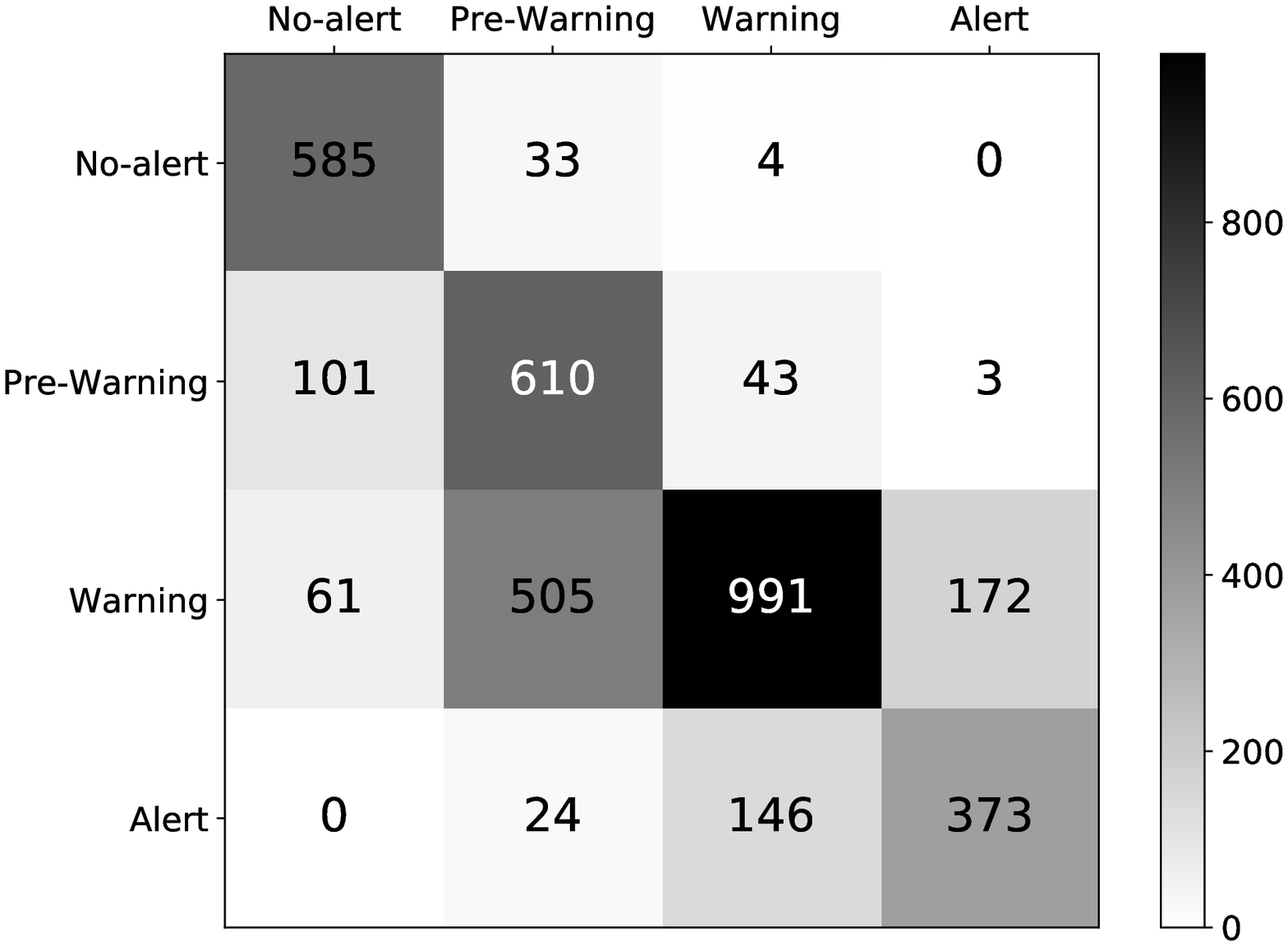}
    \label{fig:CNN:Bloque2LSTM957550:1dia}
  }
  \subfloat[Second block. 72 hours.]{
    \includegraphics[ width=0.331\textwidth]{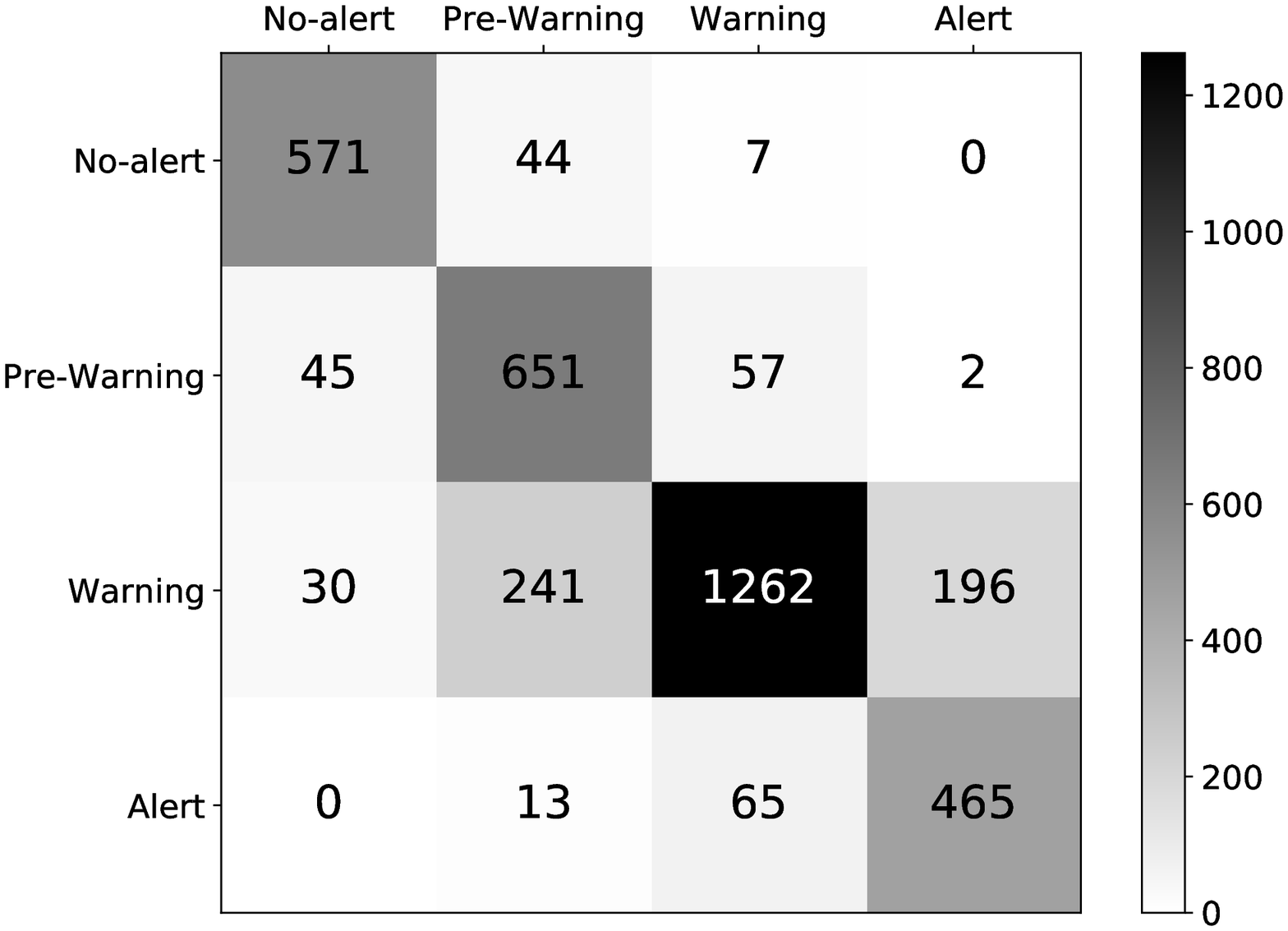}
    \label{fig:CNN:Bloque2LSTM957550:3dia}
  }
  \subfloat[Second block. 168 hours.]{
    \includegraphics[ width=0.331\textwidth]{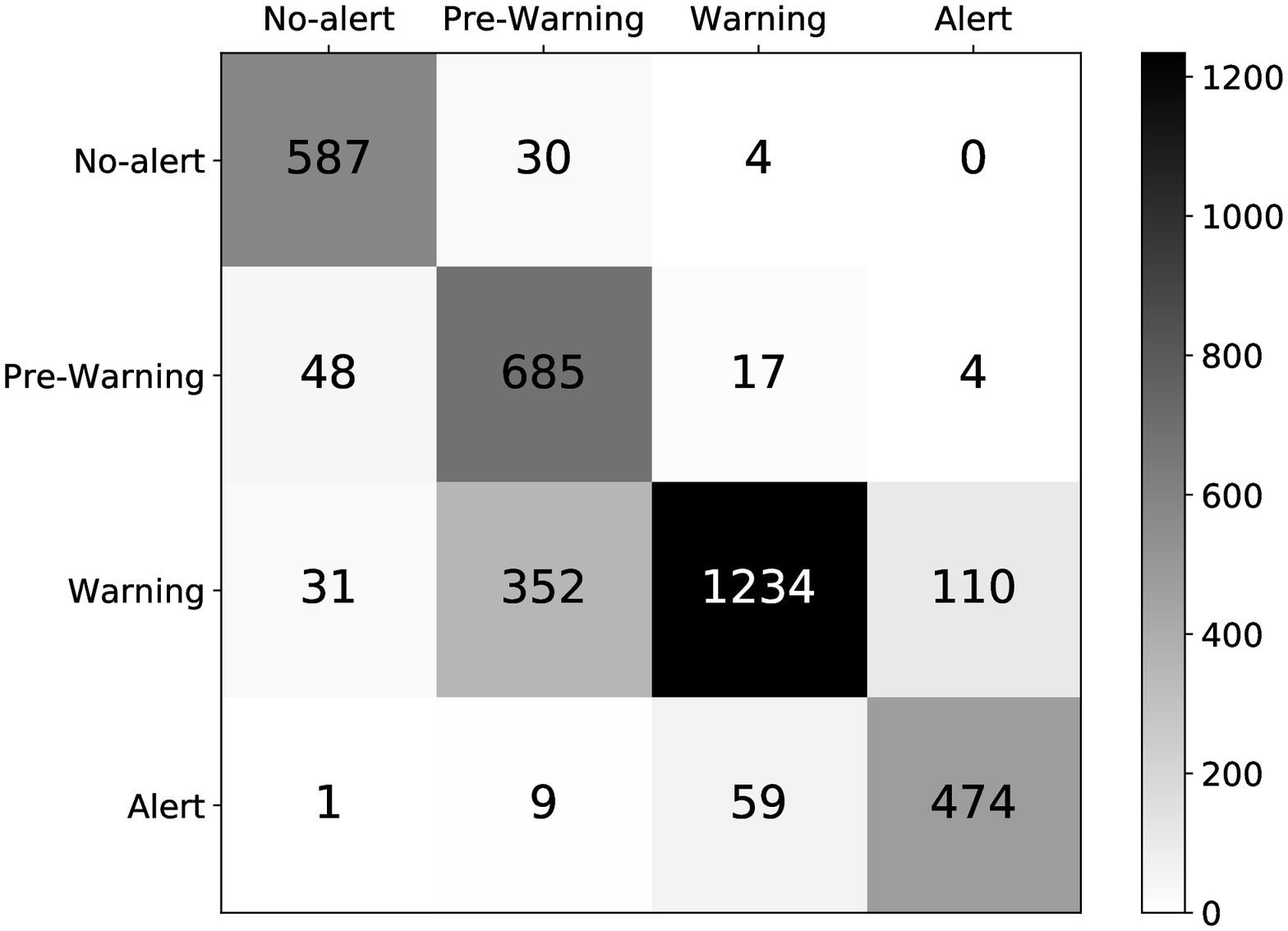}
    \label{fig:CNN:Bloque2LSTM957550:7dia}
  }\\
  \subfloat[Third block. 24 hours.]{
    \includegraphics[ width=0.331\textwidth]{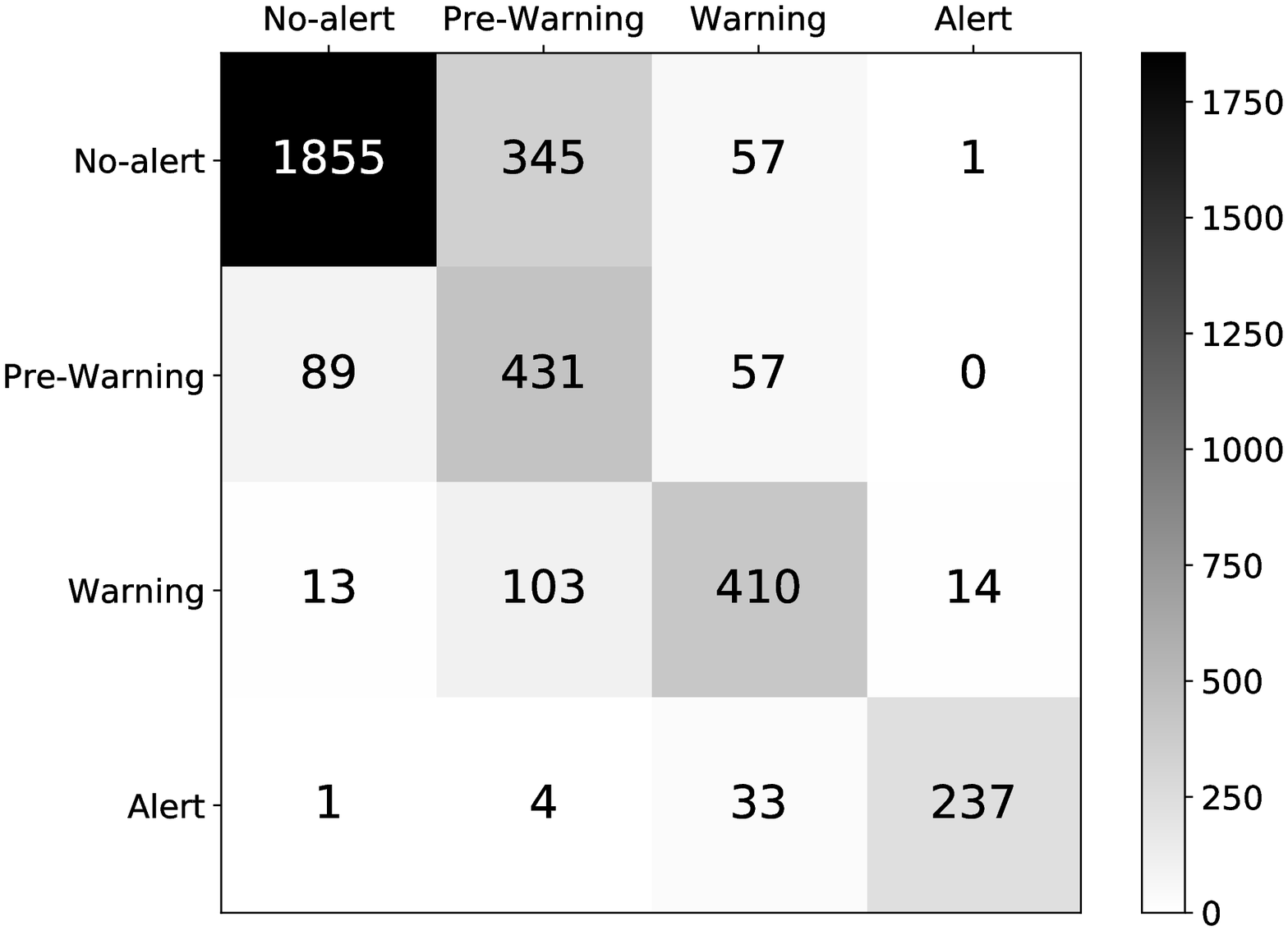}
    \label{fig:CNN:Bloque3LSTM957550:1dia}
  }
  \subfloat[Third block. 72 hours.]{
    \includegraphics[width=0.331\textwidth]{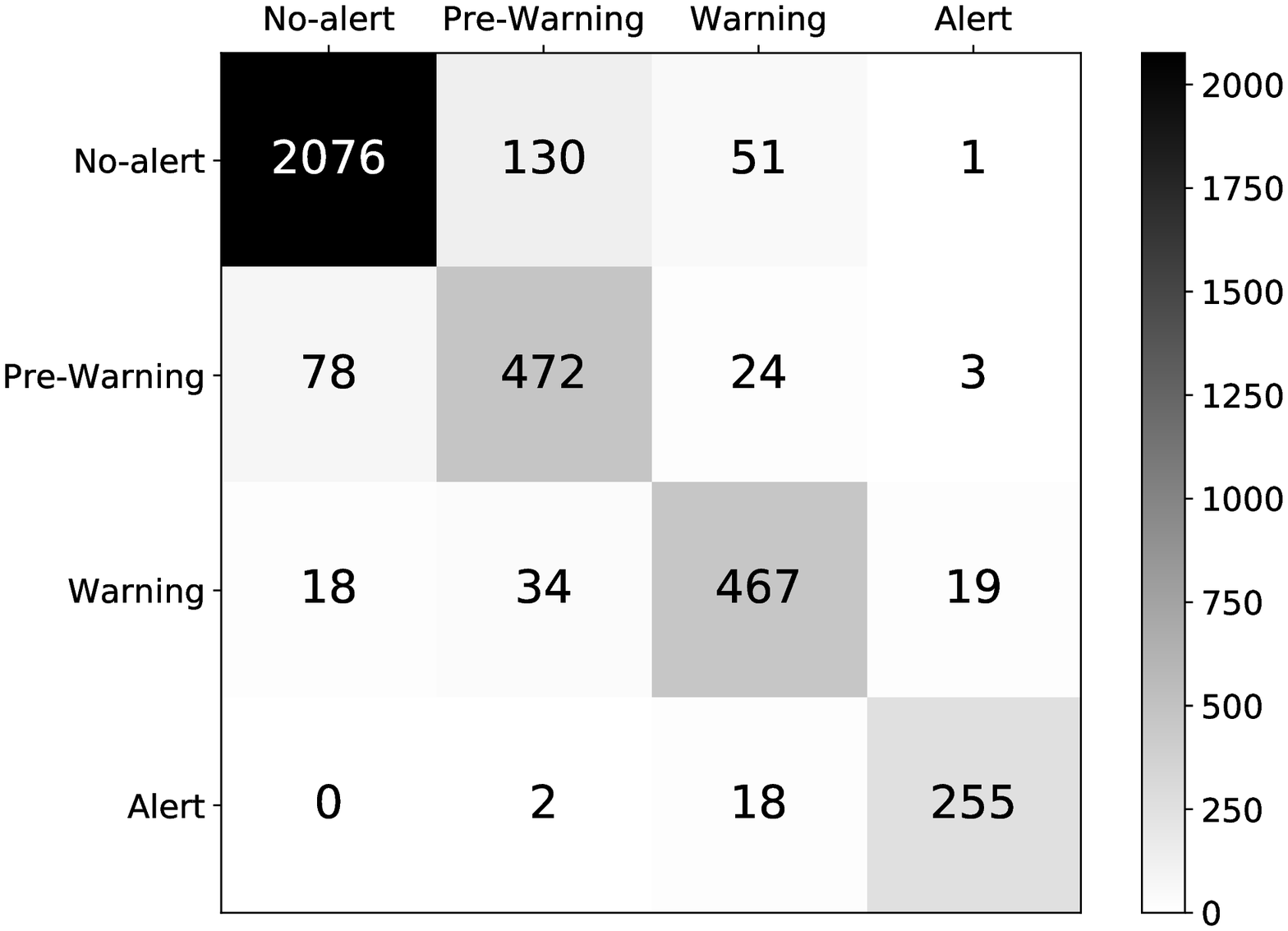}
    \label{fig:CNN:Bloque3LSTM957550:3dia}
  }
  \subfloat[Third block. 168 hours.]{
    \includegraphics[ width=0.331\textwidth]{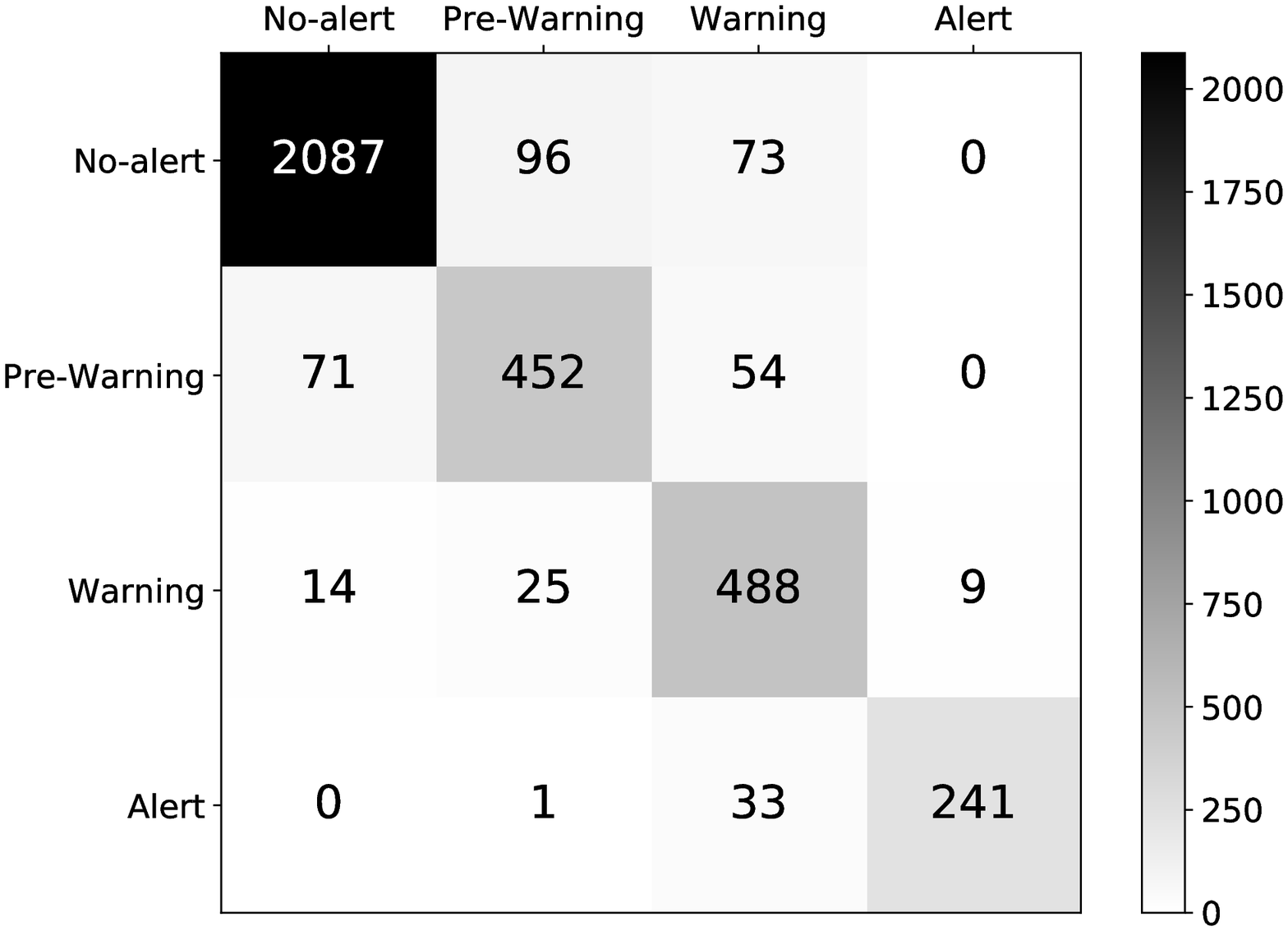}
    \label{fig:CNN:Bloque3LSTM957550:7dia}
  }
\caption{Confusion matrices obtained from the CNN model for the Rule II. Three input sizes are considered: 24 hours, 72 hours and 168 hours of data per station. Two blocks are considered: the second block (6:00-12:00) and the third block (12:00-18:00). Each row represents the actual class and each column represents the predicted class.}
\label{fig:CNN:Rule_II}
\end{minipage}
}
}
\end{figure}

In this approach, a model of CNN networks is considered. The architecture of the model can be summarized as follows: (1) a first convolutional layer with 16 kernels, kernel size 5 and ReLU as activation function; (2) a max-pooling layer; (3) a second convolutional layer with 64 kernels, kernel size 3 and ReLU as activation function; (4) another max-pooling layer; (5) A dense layer with 20 neurons and ReLU as activation function; (6) a dropout layer with index 0.05; (7) a final dense output layer with 20 neurons and softmax as activation function. Cross-entropy loss has been used for training. The confusion matrices are shown in Figs. \ref{fig:CNN:Rule_I}, and \ref{fig:CNN:Rule_II};  while the next comparisons are undertaken by the normalized ones.

\begin{figure*}
\centering
{\renewcommand{\arraystretch}{1.1}
\rotatebox{90}{
\begin{minipage}[c][][c]{\textheight}
\centering
\subfloat[Second block. 24 hours.]{
    \includegraphics[ width=0.331\textwidth]{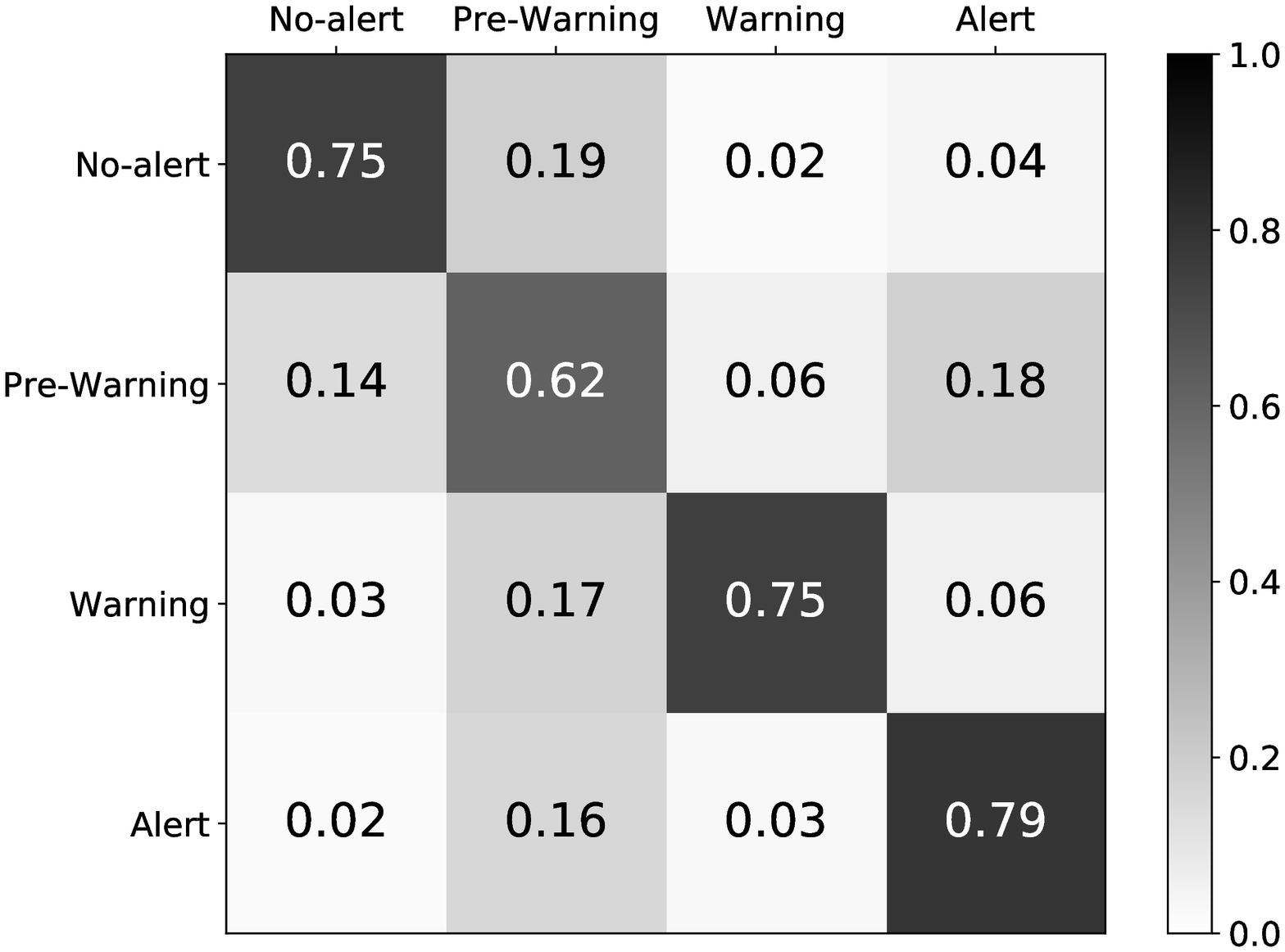}
    \label{fig:CNNnorm:Bloque2CNN959075:1dia}
  }
  \subfloat[Second block. 72 hours.]{
    \includegraphics[ width=0.331\textwidth]{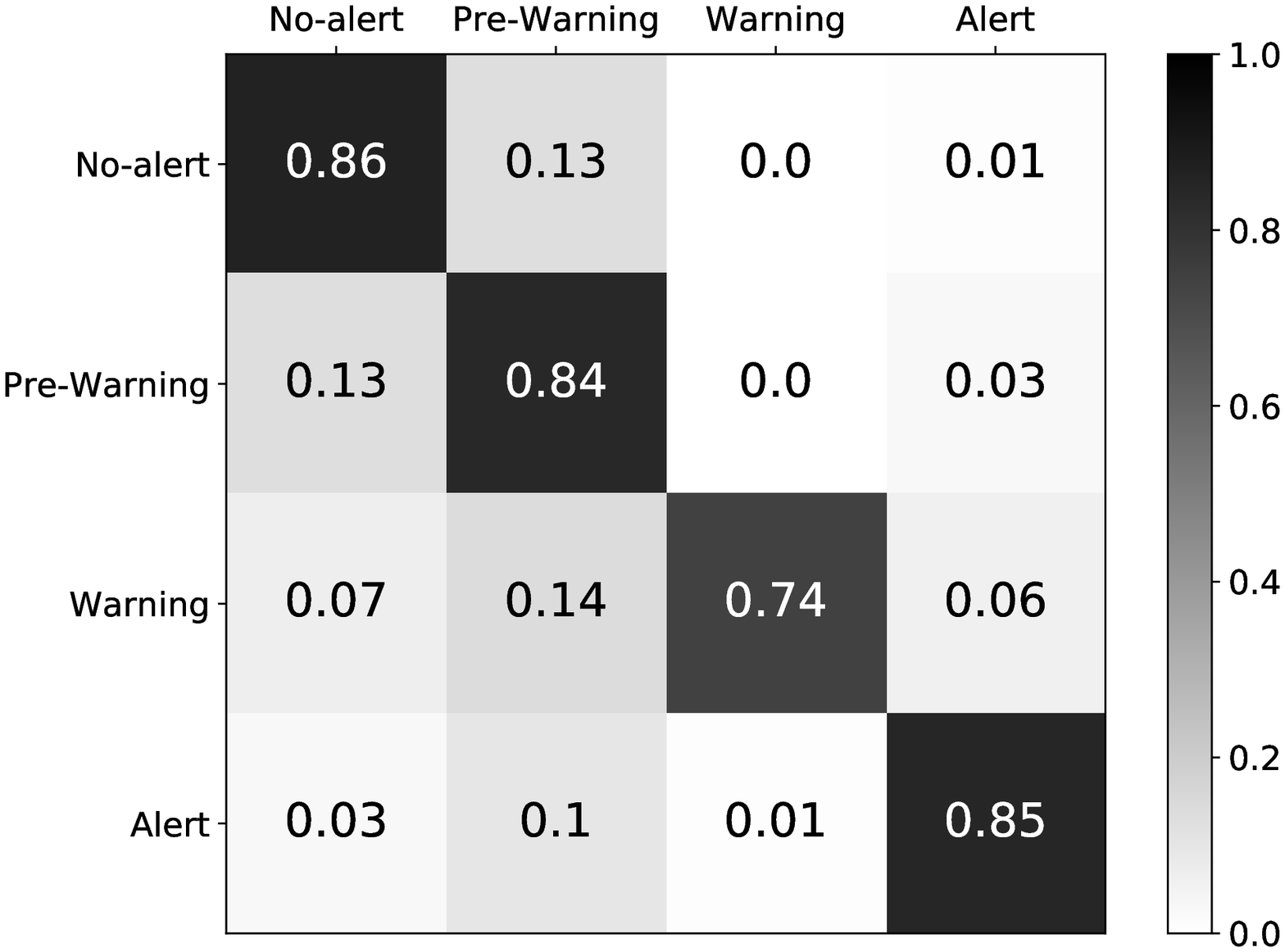}
    \label{fig:CNNnorm:Bloque2CNN959075:3dia}
  }
  \subfloat[Second block. 168 hours.]{
    \includegraphics[ width=0.331\textwidth]{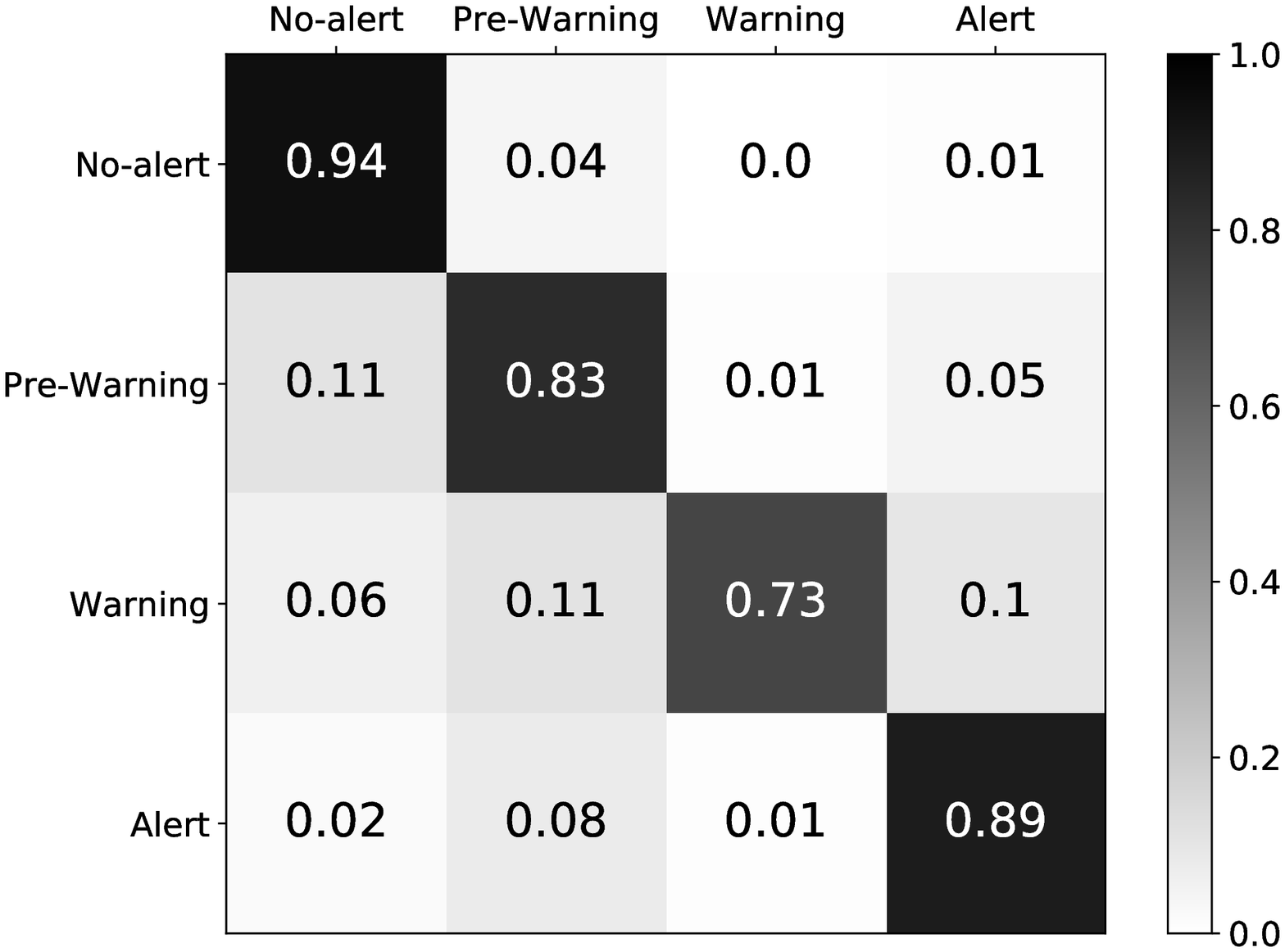}
    \label{fig:CNNnorm:Bloque2CNN959075:7dia}
  }\\
  \subfloat[Third block. 24 hours.]{
    \includegraphics[ width=0.331\textwidth]{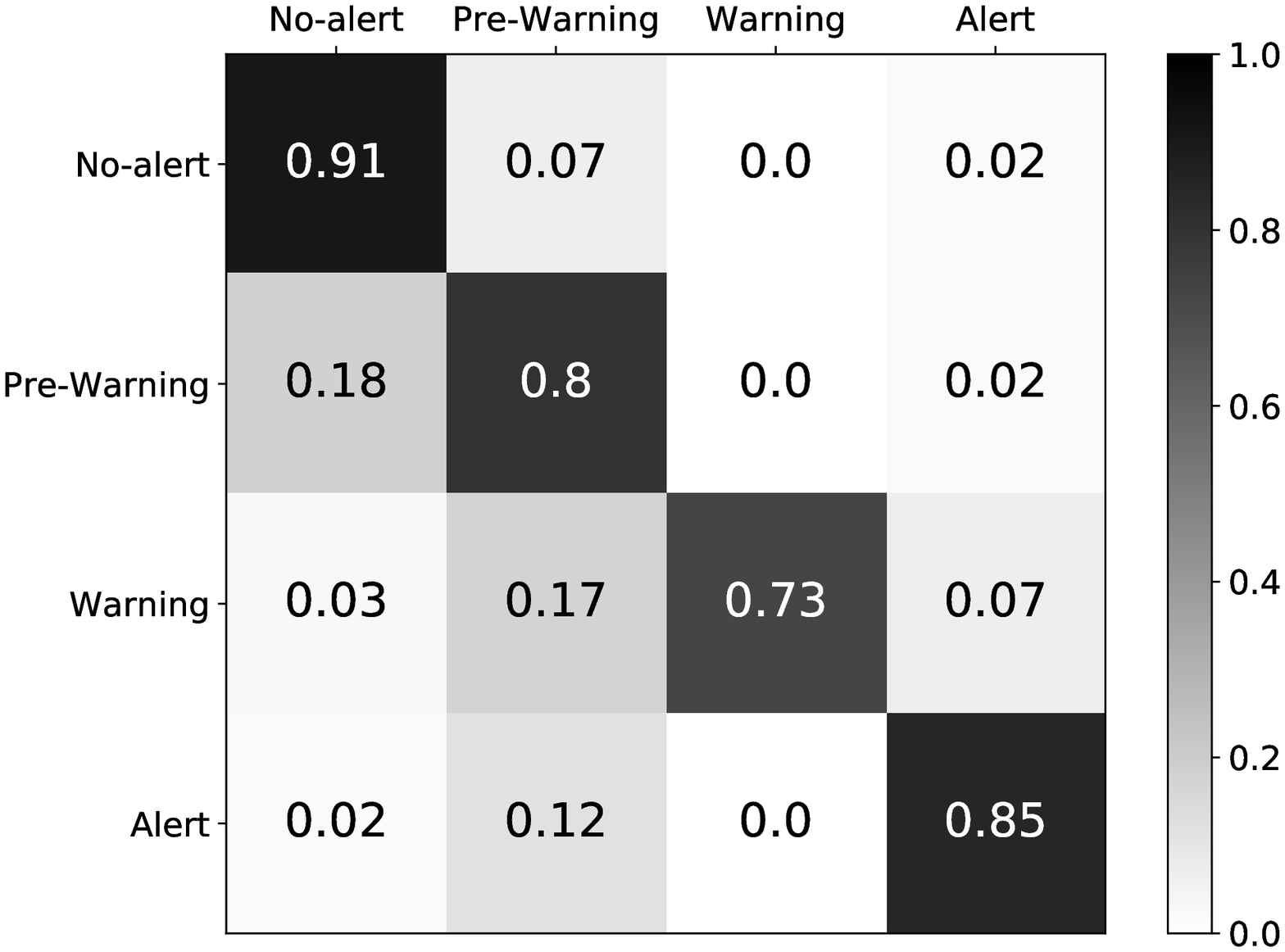}
    \label{fig:CNNnorm:Bloque3CNN959075:1dia}
  }
  \subfloat[Third block. 72 hours.]{
    \includegraphics[ width=0.331\textwidth]{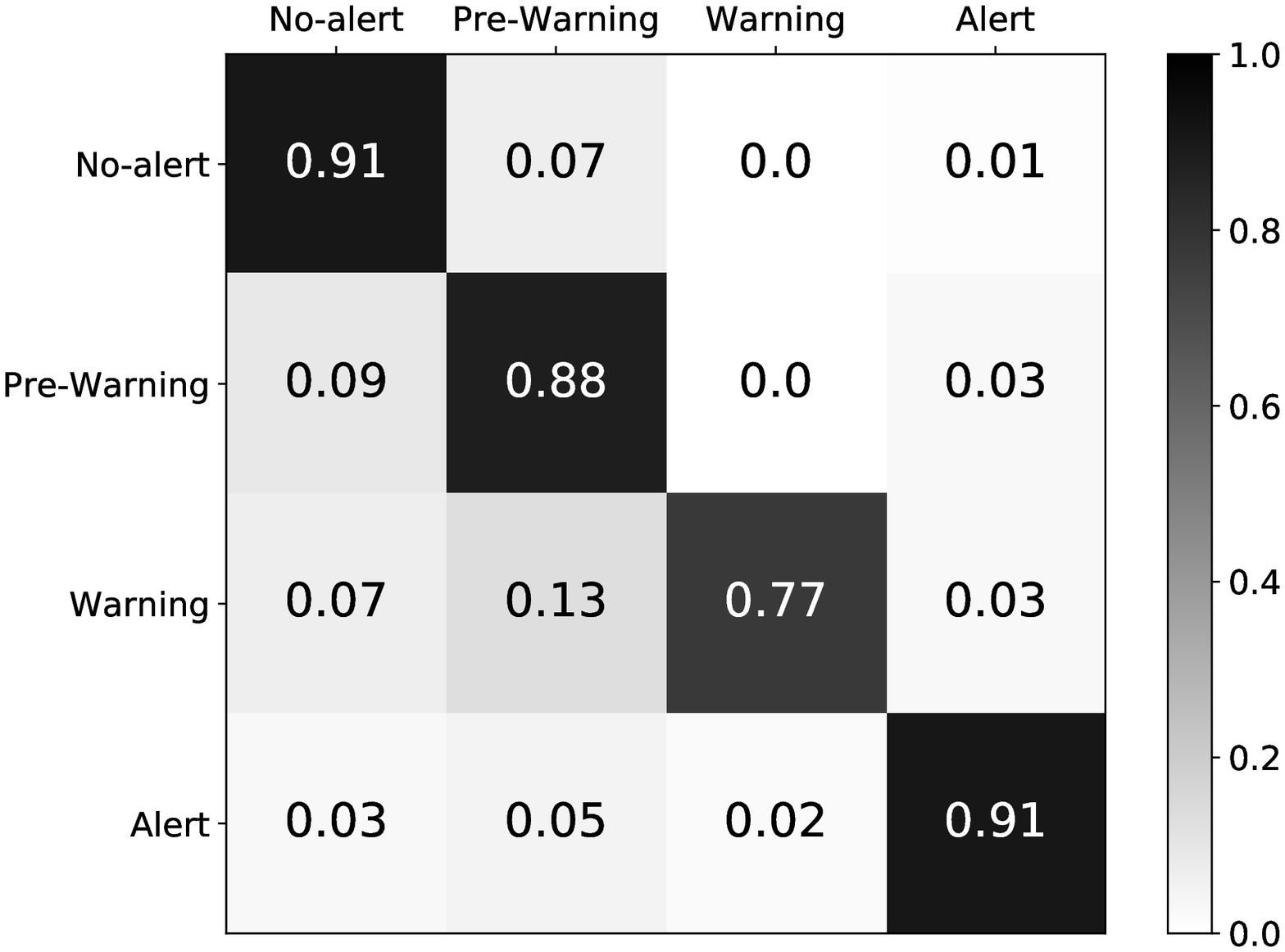}
    \label{fig:CNNnorm:Bloque3CNN959075:3dia}
  }
  \subfloat[Third block. 168 hours.]{
    \includegraphics[ width=0.331\textwidth]{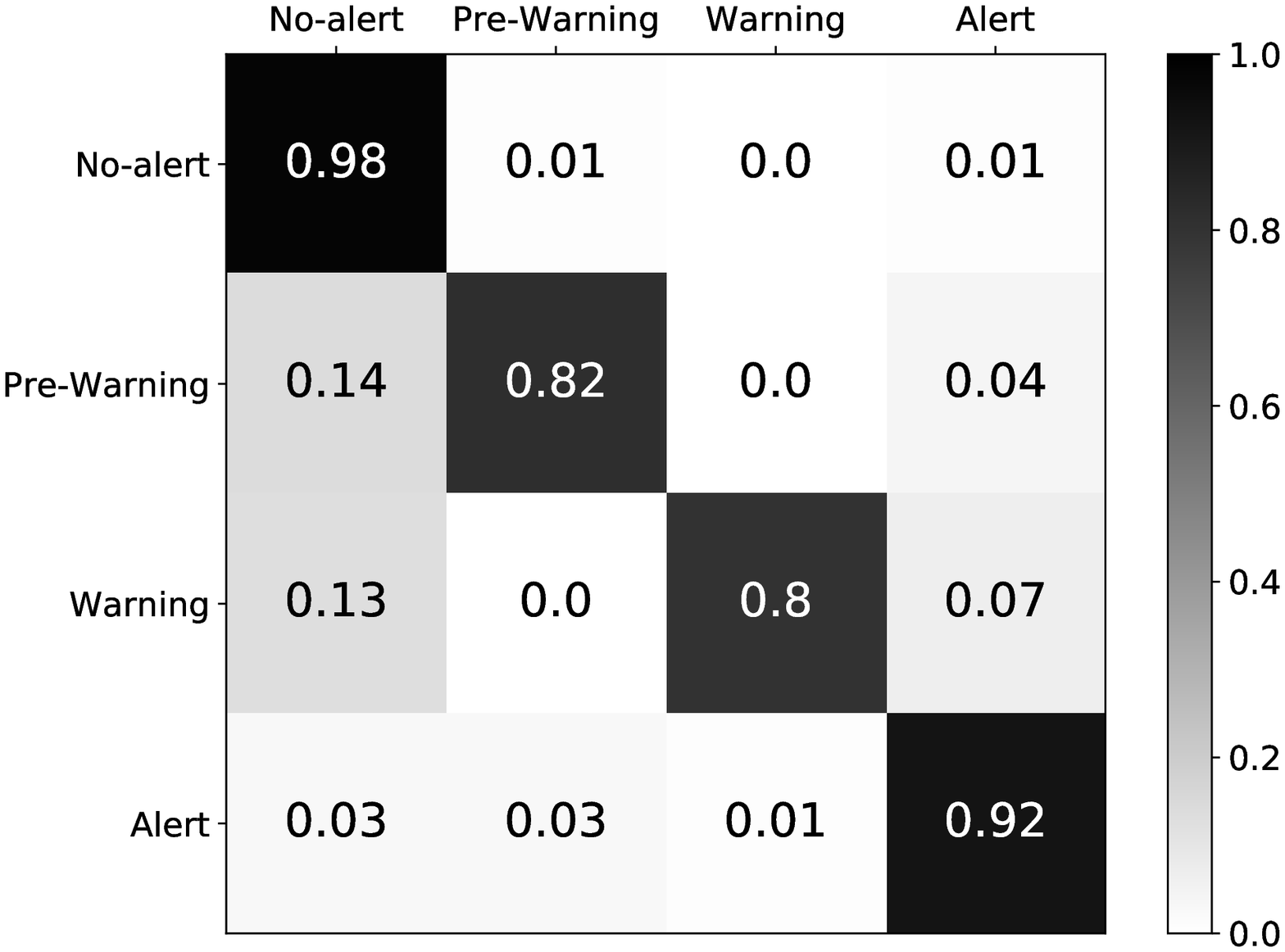}
    \label{fig:CNNnorm:Bloque3CNN959075:7dia}
  }
 \caption{Normalized confusion matrices obtained from the CNN model for the Rule I and the Rule II. Three input sizes are considered: 24 hours, 72 hours and 168 hours of data per station. Two blocks are considered: the second block (6:00-12:00) and the third block (12:00-18:00). Each row represents the actual class and each column represents the predicted class.}
\label{fig:CNNnorm:Rule_I}
\end{minipage}
        }
}
\end{figure*}

Following the same line of experiments, the CNN model is trained for the same conditions that are established in the case of the LSTM model. Comparing Figs. \ref{fig:LSTMnorm:Bloque2LSTM959075:1dia}, \ref{fig:LSTMnorm:Bloque2LSTM959075:3dia}, \ref{fig:LSTMnorm:Bloque2LSTM959075:7dia} and Figs. \ref{fig:CNNnorm:Bloque2CNN959075:1dia}, \ref{fig:CNNnorm:Bloque2CNN959075:3dia}, \ref{fig:CNNnorm:Bloque2CNN959075:7dia}, it is shown that better results are obtained in the case of the CNN than in the LSTM one for the second block (6:00-12:00) and the Rule I. When using CNN model, there is a significant difference between using 24, 72 or 168 hours of input data, obtaining better results for greater amounts of hours. For 168 hours of input data, the accuracy ranges from 73\% to 94\%. For 71 hours, the accuracy has a higher inferior accuracy 74\%, but also exhibits a much lower higher accuracy 86\%. For the three configurations of number of hours, CNN outperforms the equivalent LSTM model for the second block and the Rule I. 

Comparing now Figs. \ref{fig:LSTMnorm:Bloque3LSTM959075:1dia}, \ref{fig:LSTMnorm:Bloque3LSTM959075:3dia}, \ref{fig:LSTMnorm:Bloque3LSTM959075:7dia} and Figs. \ref{fig:CNNnorm:Bloque3CNN959075:1dia}, \ref{fig:CNNnorm:Bloque3CNN959075:3dia}, \ref{fig:CNNnorm:Bloque3CNN959075:7dia}, we obtain considerably better results for the Rule I and for the third block (12:00-18:00) from the CNN model. As for the second block, there is a significant difference between using 24, 72 or 168 hours of input data, obtaining better results for those configurations with larger amounts of hours. The accuracy for 168 hours configuration ranges from 80\% to 98\%. Besides, the CNN model shows the lowest high-adjacent errors and non-adjacent errors than their equivalents in the LSTM model.

\begin{figure*}
\centering
{\renewcommand{\arraystretch}{1.1}
\rotatebox{90}{
\begin{minipage}[c][][c]{\textheight}
\centering
  \subfloat[Second block. 24 hours.]{
    \includegraphics[width=0.331\textwidth]{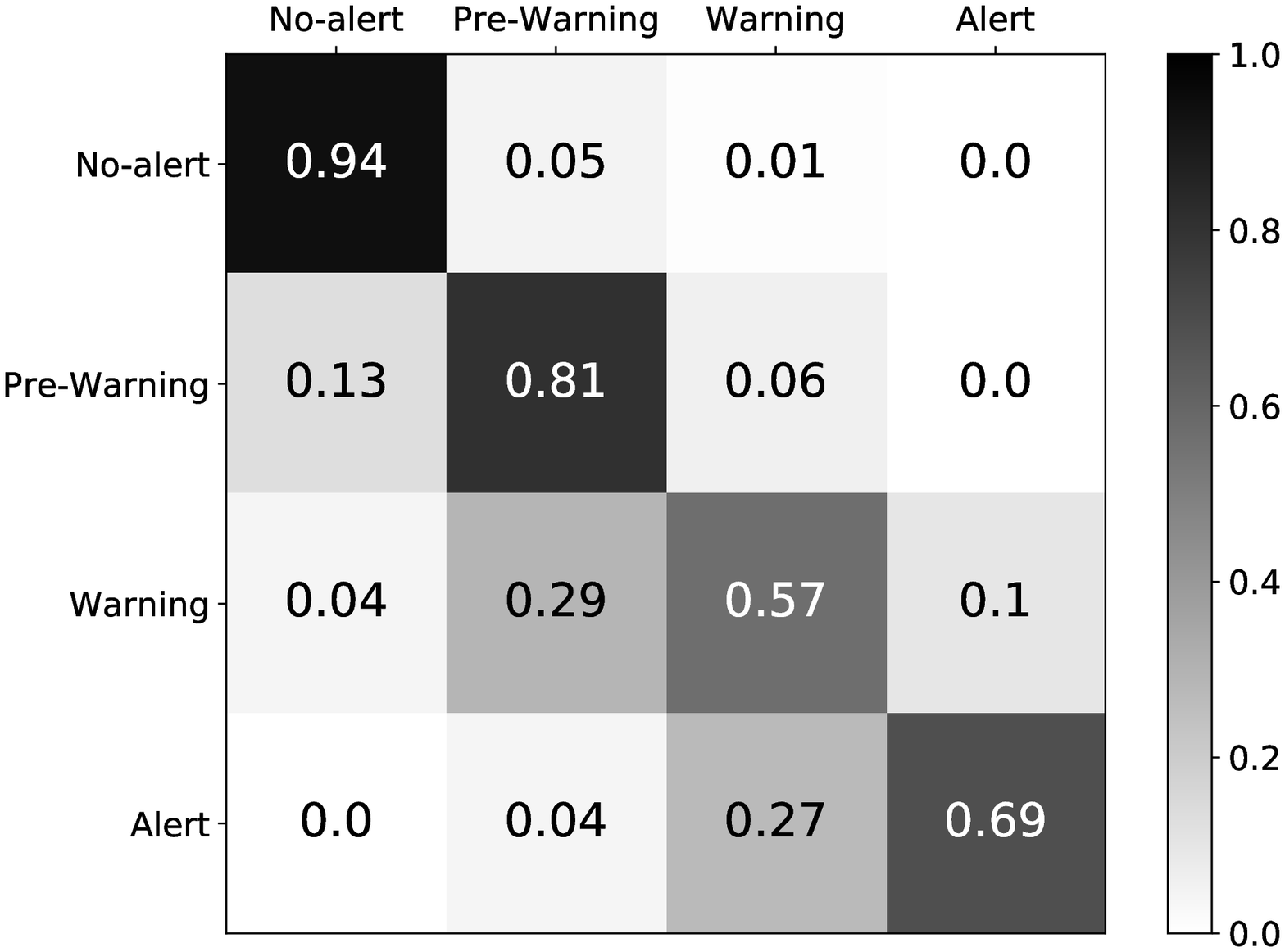}
    \label{fig:CNNnorm:Bloque2CNN957550:1dia}
  }
  \subfloat[Second block. 72 hours.]{
    \includegraphics[width=0.331\textwidth]{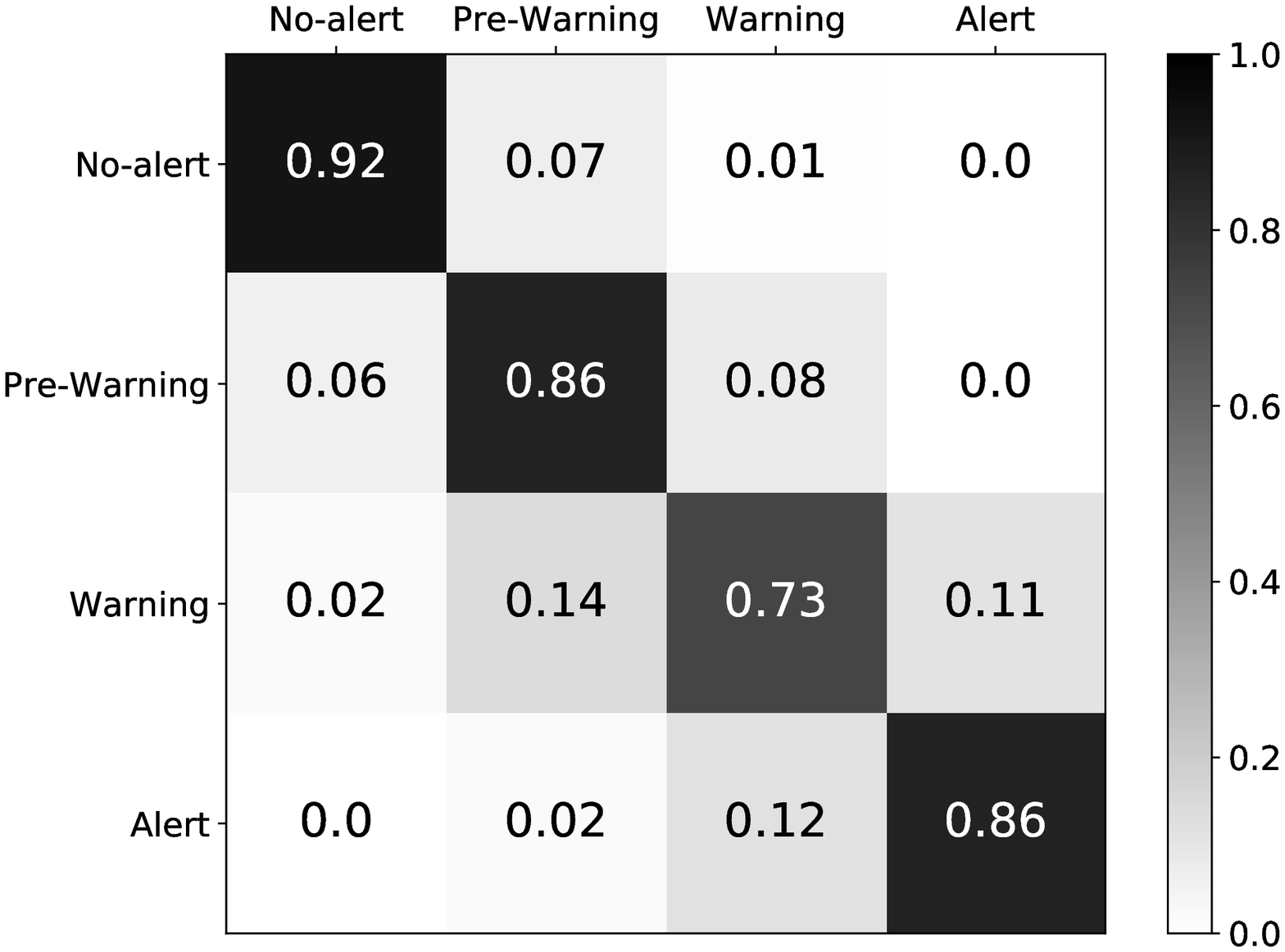}
    \label{fig:CNNnorm:Bloque2CNN957550:3dia}
  }
  \subfloat[Second block. 168 hours.]{
    \includegraphics[width=0.331\textwidth]{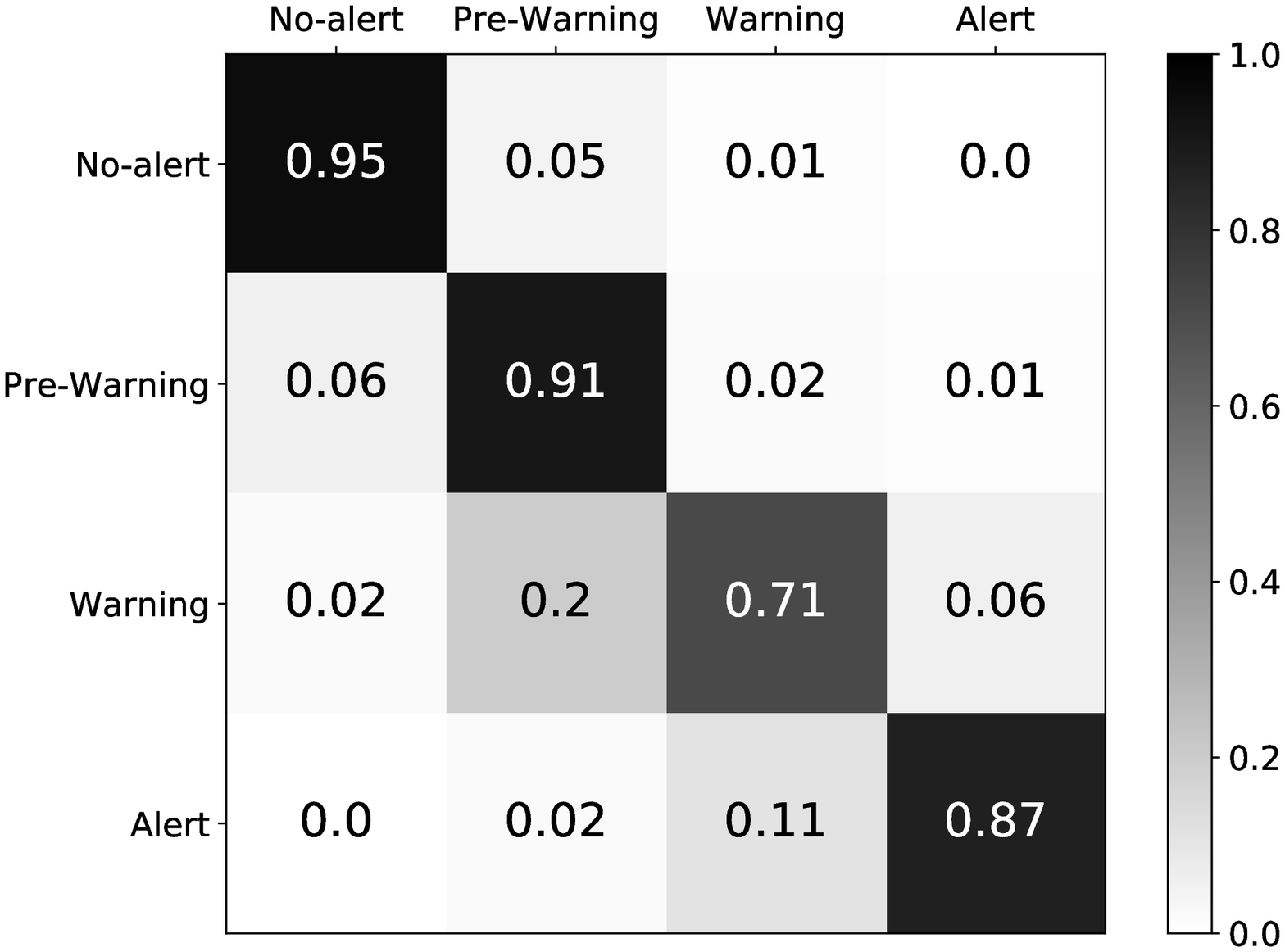}
    \label{fig:CNNnorm:Bloque2CNN957550:7dia}
  }\\
  \subfloat[Third block. 24 hours.]{
    \includegraphics[width=0.331\textwidth]{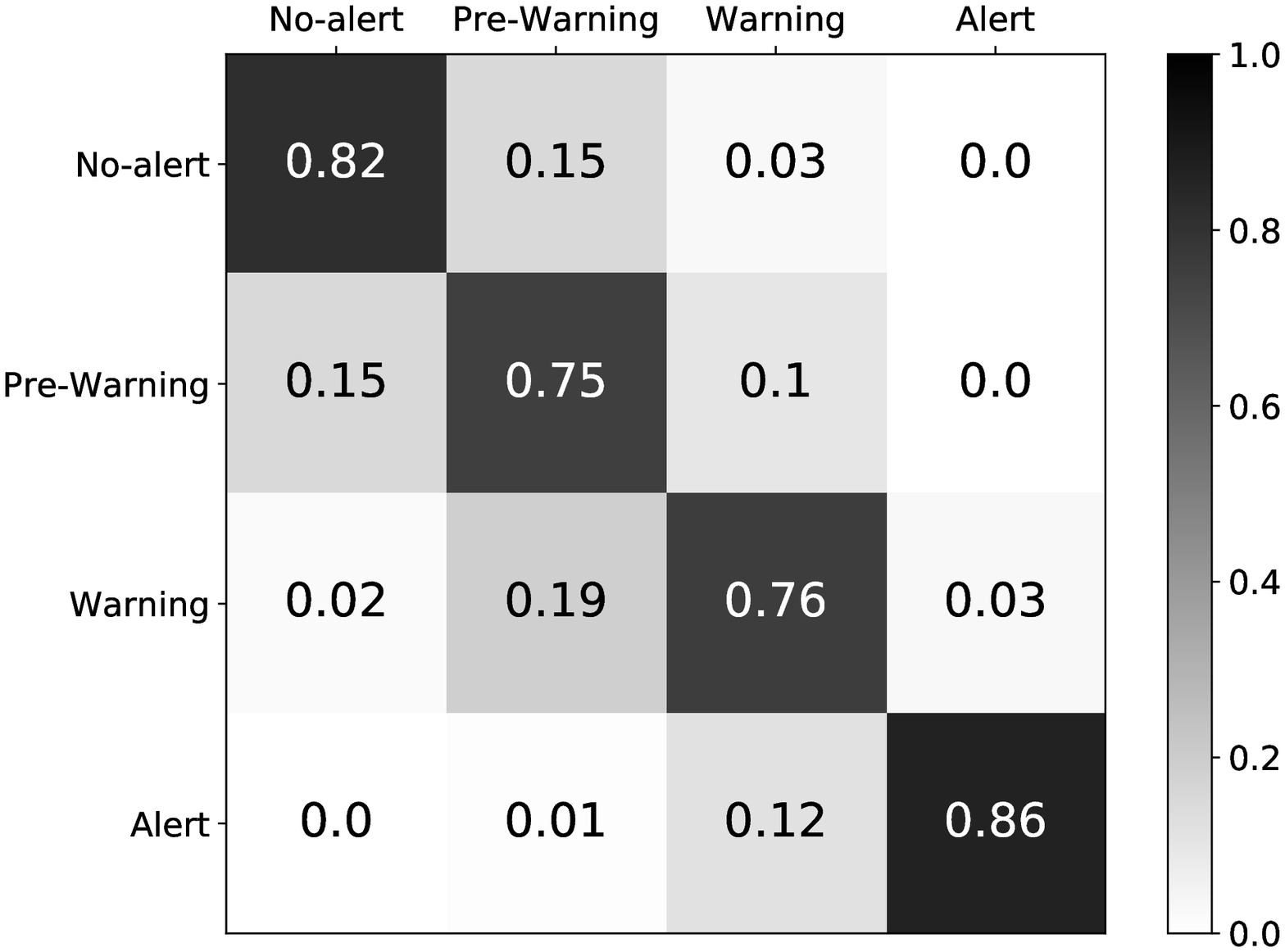}
    \label{fig:CNNnorm:Bloque3CNN957550:1dia}
  }
  \subfloat[Third block. 72 hours.]{
    \includegraphics[width=0.331\textwidth]{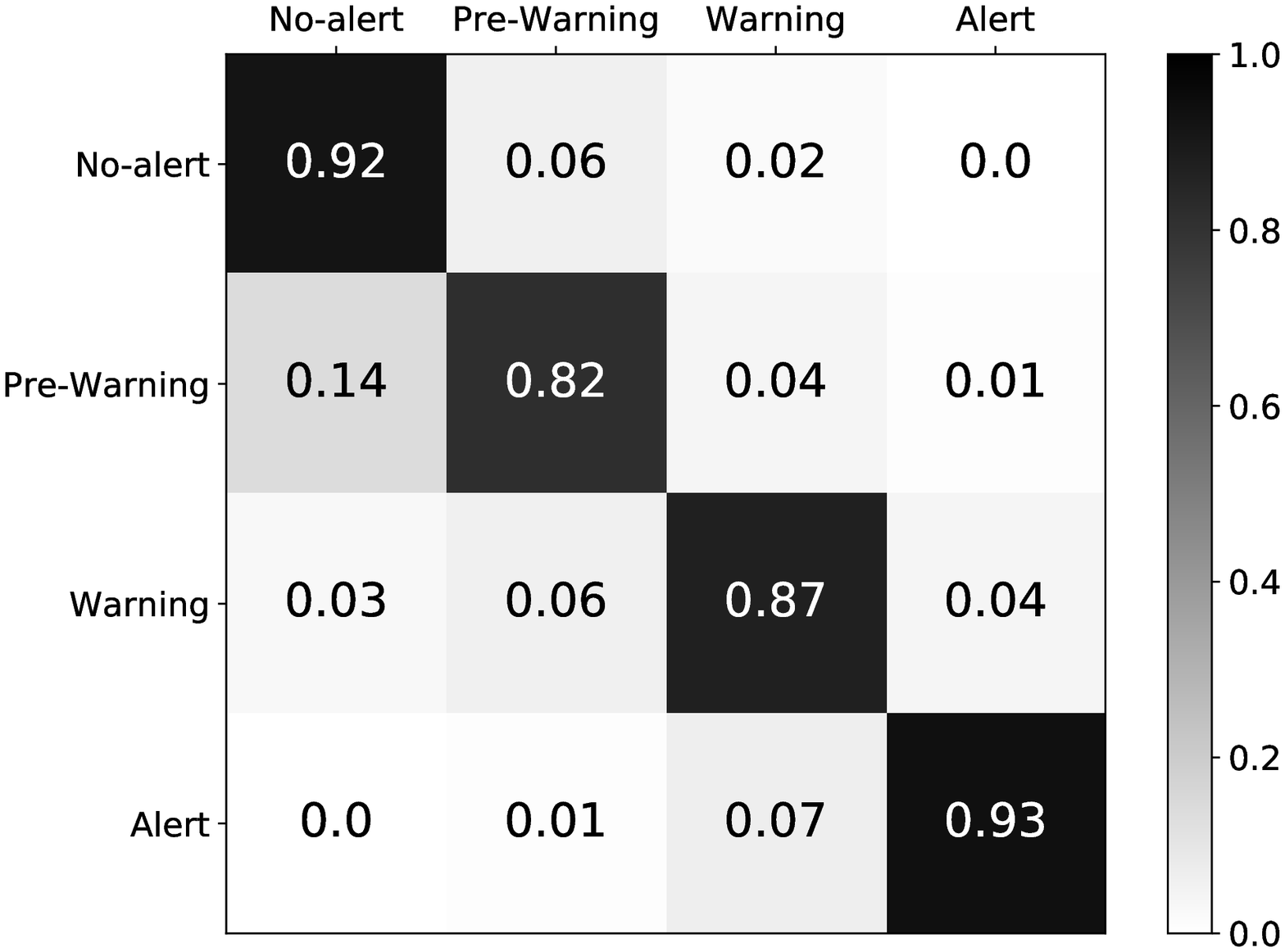}
    \label{fig:CNNnorm:Bloque3CNN957550:3dia}
  }
  \subfloat[Third block. 168 hours.]{
    \includegraphics[width=0.331\textwidth]{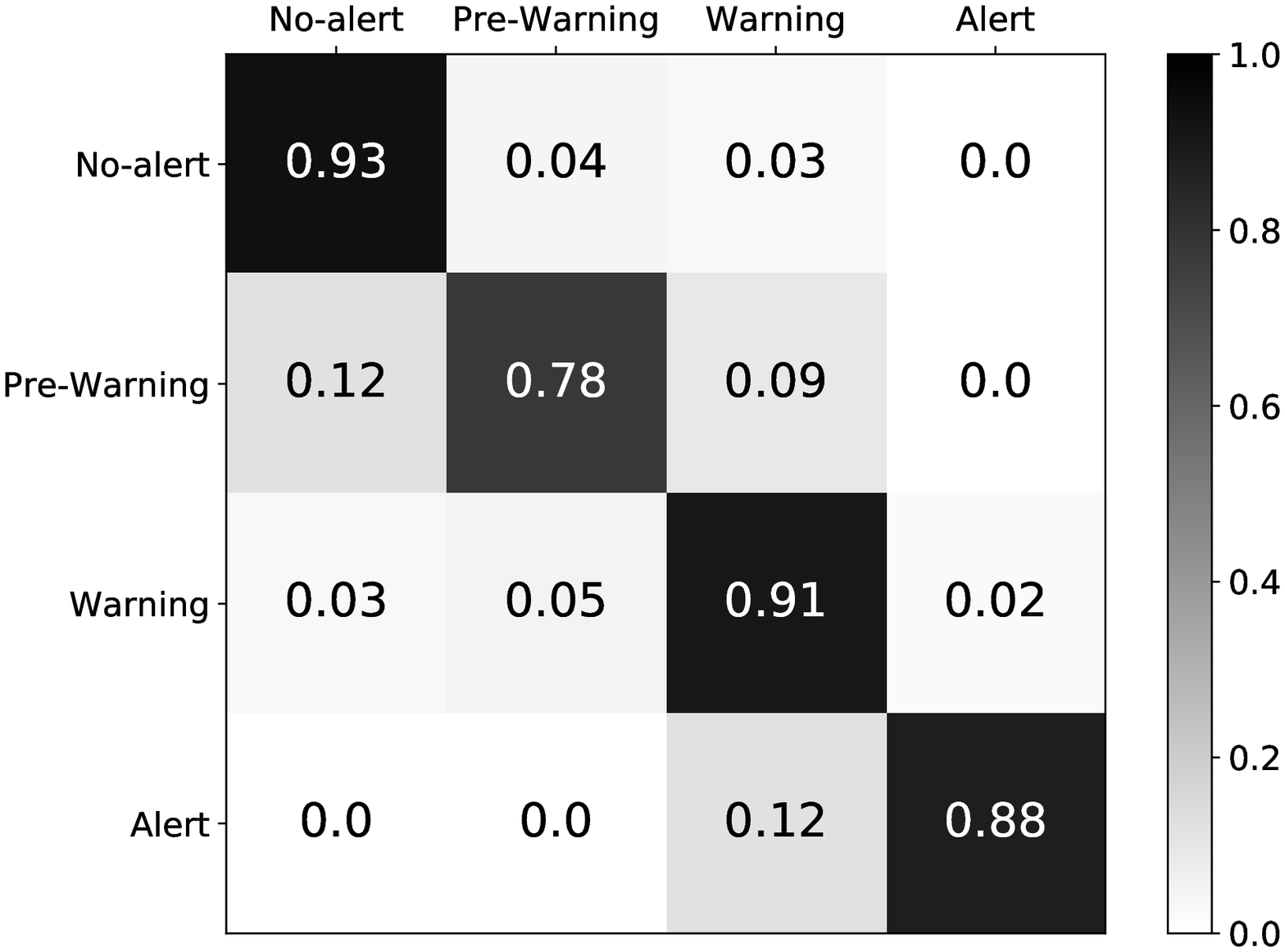}
    \label{fig:CNNnorm:Bloque3CNN957550:7dia}
  }
 \caption{Normalized confusion matrices obtained from the CNN model for the Rule II. Three input sizes are considered: 24 hours, 72 hours and 168 hours of data per station. Two blocks are considered: the second block (6:00-12:00) and the third block (12:00-18:00). Each row represents the actual class and each column represents the predicted class.}
\label{fig:CNNnorm:Rule_II}
\end{minipage}
        }
}
\end{figure*}

Considering now the Rule II and second block, the results of LSTM (Figs. \ref{fig:LSTMnorm:Bloque2LSTM957550:1dia}, \ref{fig:LSTMnorm:Bloque2LSTM957550:3dia}, \ref{fig:LSTMnorm:Bloque2LSTM957550:7dia}) and CNN (Figs.  \ref{fig:CNNnorm:Bloque2CNN957550:1dia}, \ref{fig:CNNnorm:Bloque2CNN957550:3dia}, \ref{fig:CNNnorm:Bloque2CNN957550:7dia}) models are difficult to ascertain a best configuration and model. Ranges of accuracy overlapping, rawly being better the high accuracy of CNN model ---up to 95\% for the configuration of 168 hours---, but being in many cases the low accuracy worse than their counterpart configuration at LSTM model. In this case, the exploration of diverse models takes a great value, and the evaluation of an ensemble model could be proposed as future work. The lowest high-errors, both adjacent and non-adjacent, are achieved for the configuration of 168 hours, being respectively 11\% for the Alert labels predicted as Warning, and 2\% for two cases.


Finally, for the Rule II and for the third block (12:00-18:00), we can observe that the CNN model (Figs. \ref{fig:CNNnorm:Bloque3CNN957550:1dia}, \ref{fig:CNNnorm:Bloque3CNN957550:3dia}, \ref{fig:CNNnorm:Bloque3CNN957550:7dia}) obtains again better results than the LSTM model (Figs. \ref{fig:LSTMnorm:Bloque3LSTM957550:1dia}, \ref{fig:LSTMnorm:Bloque3LSTM957550:3dia}, \ref{fig:LSTMnorm:Bloque3LSTM957550:7dia}). The accuracy for the CNN model for 72 hours configuration ranges from 82\% to 93\%, at the same time that it keeps low the highest errors: adjacent and non-adjacent. 


\subsubsection{U-Time}
\begin{figure*}[!ht]
\centering
  \subfloat[Second block. 168 hours.]{
    \includegraphics[width=0.450\textwidth]{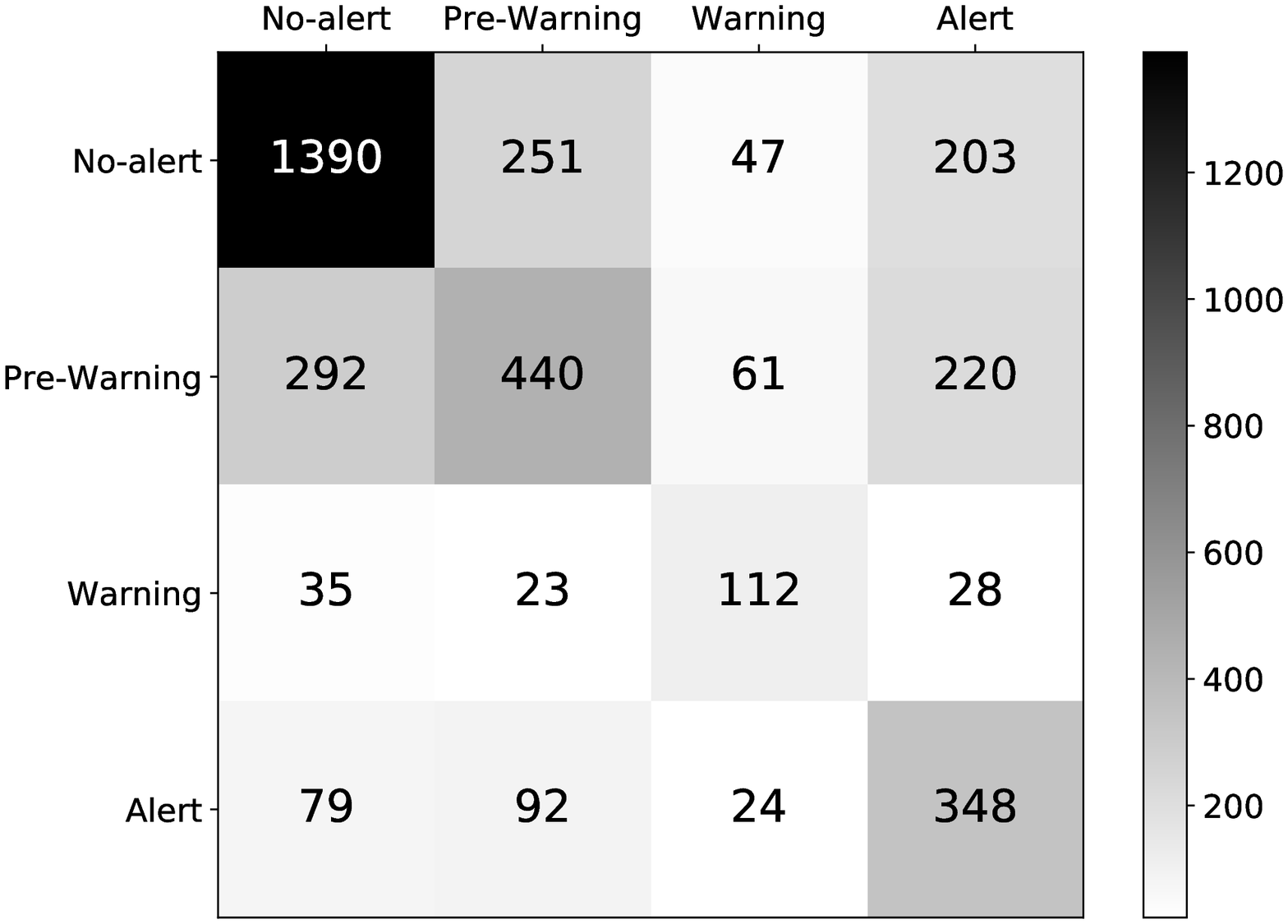}
    \label{fig:UTime:Bloque2UTime959075}
  }
  \subfloat[Third block. 168 hours.]{
    \includegraphics[width=0.450\textwidth]{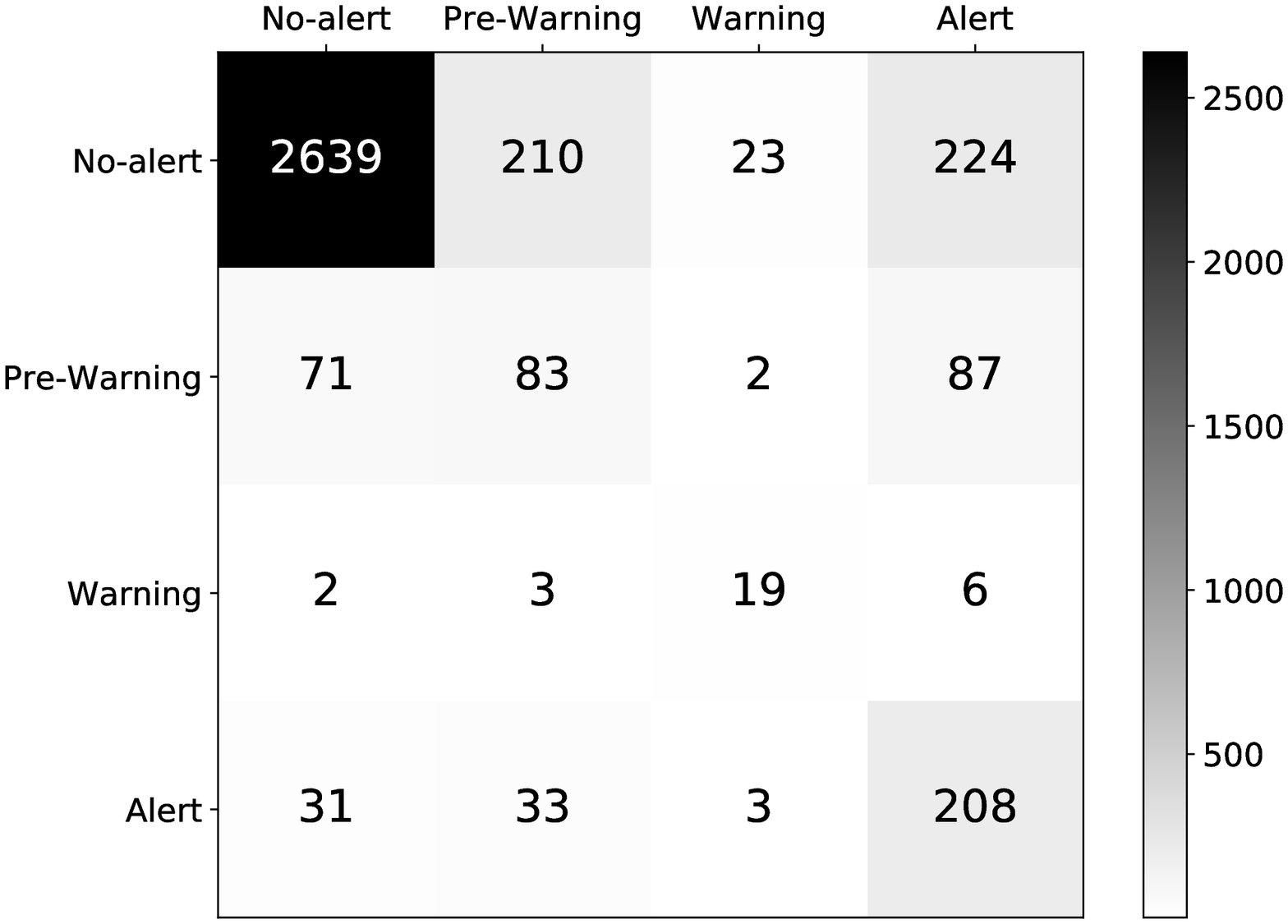}
    \label{fig:UTime:Bloque3UTime959075}
  }\\
  \subfloat[Second block. 168 hours. Normalized.]{
    \includegraphics[width=0.450\textwidth]{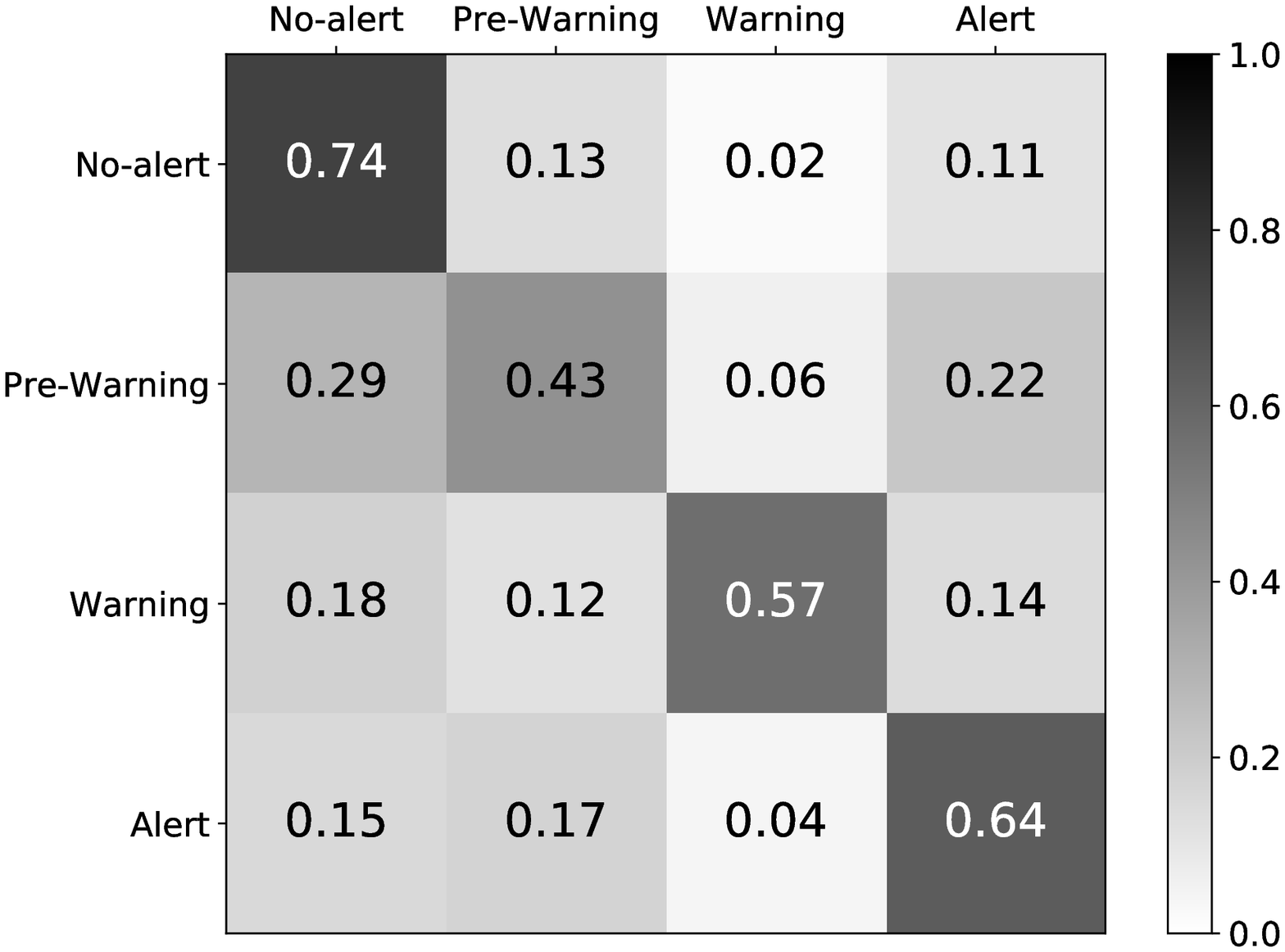}
    \label{fig:UTime:Bloque2UTime959075_norm}
  }
  \subfloat[Third block. 168 hours. Normalized.]{
    \includegraphics[width=0.450\textwidth]{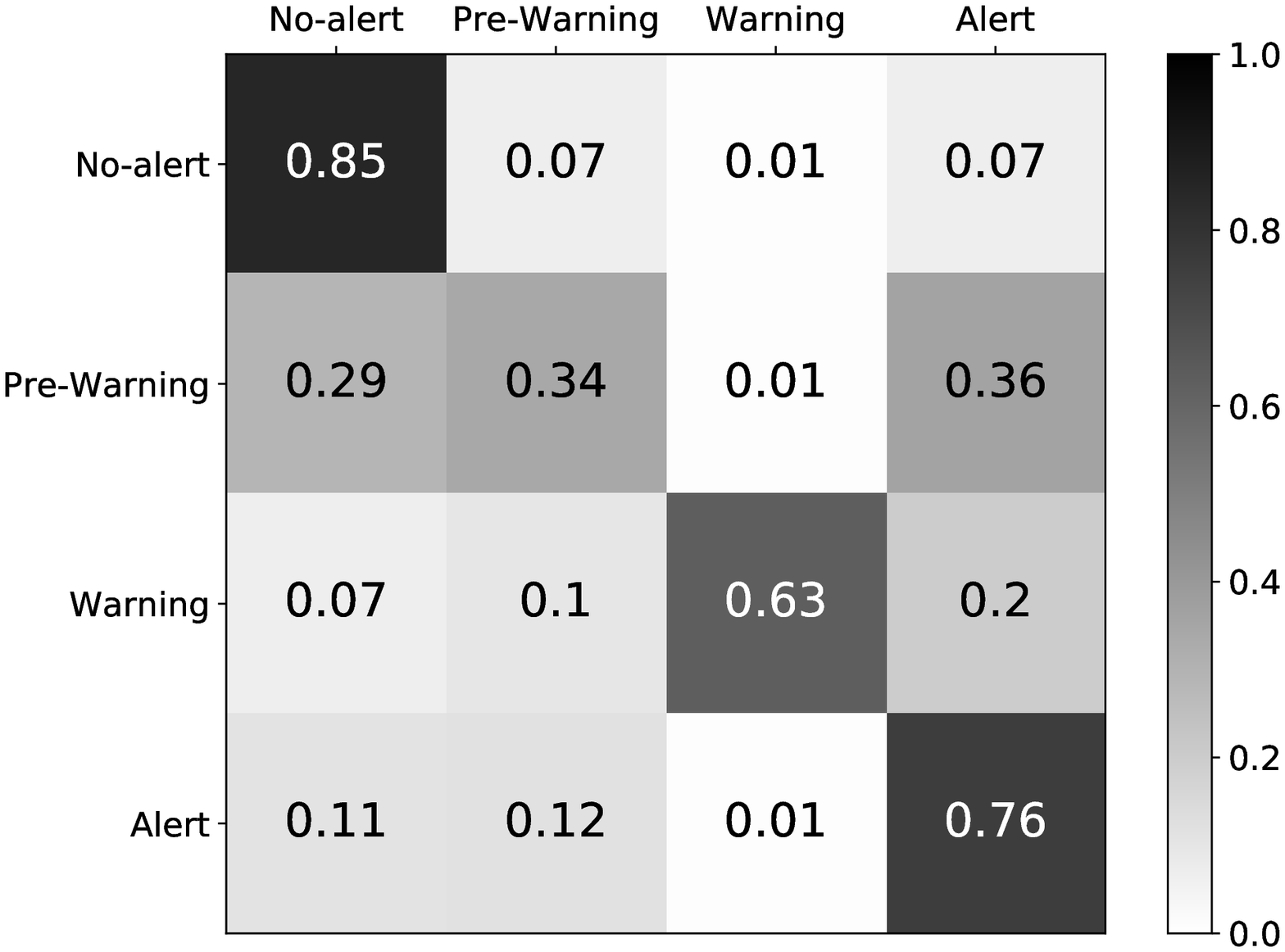}
    \label{fig:UTime:Bloque3UTime959075_norm}
  }
\caption{Confusion matrices obtained from the U-Time model for the Rule I. Only one input size is considered: 168 hours of data per station. Two blocks are considered: the second block (6:00-12:00) and the third block (12:00-18:00). Each row represents the actual class and each column represents the predicted class.}
\label{fig:UTime:RuleI}
\end{figure*}

\begin{figure*}[!ht]
\centering
  \subfloat[Second block. 168 hours.]{
    \includegraphics[width=0.450\textwidth]{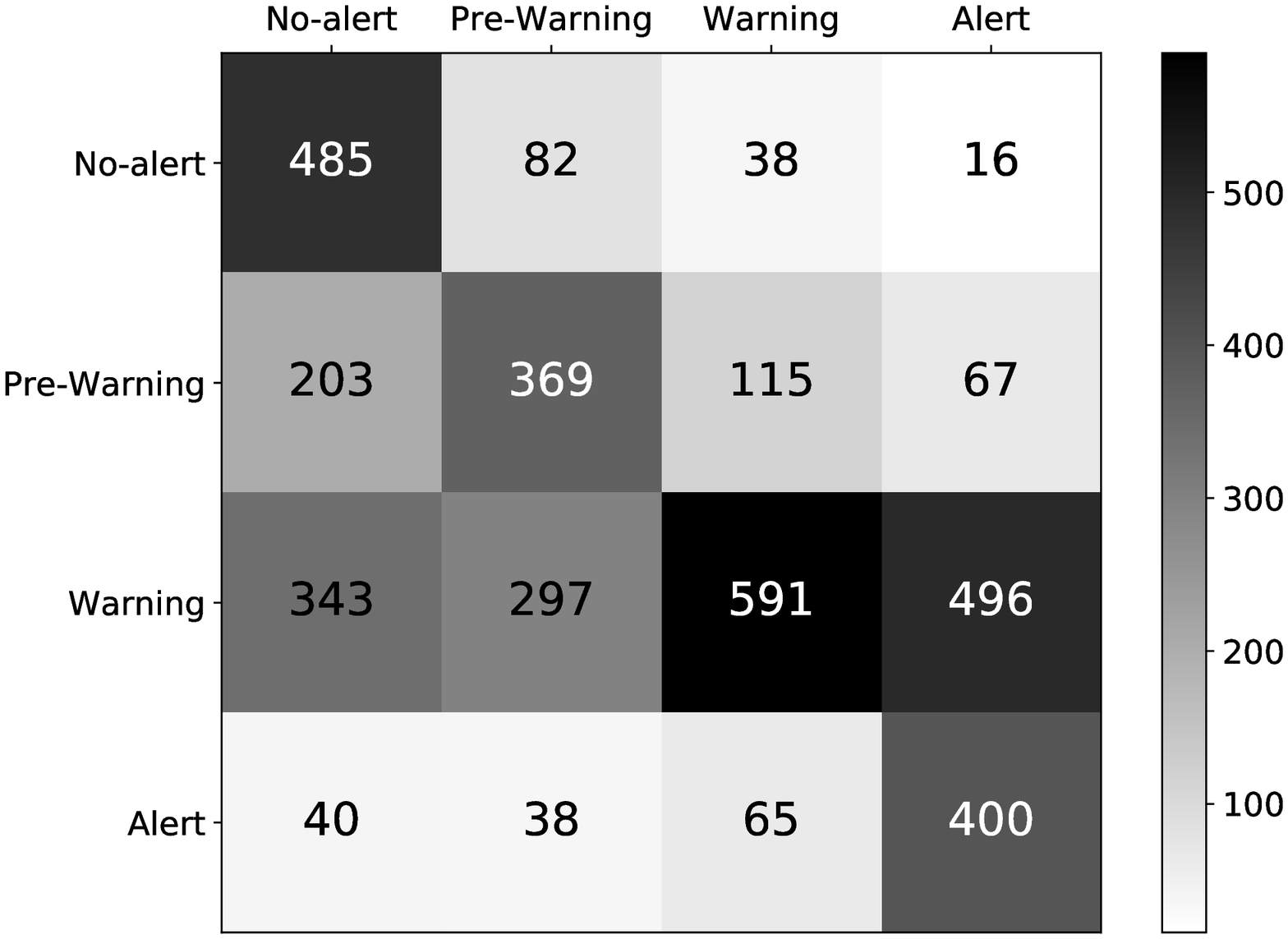}
    \label{fig:UTime:Bloque2UTime957550}
  }
  \subfloat[Third block. 168 hours.]{
    \includegraphics[width=0.450\textwidth]{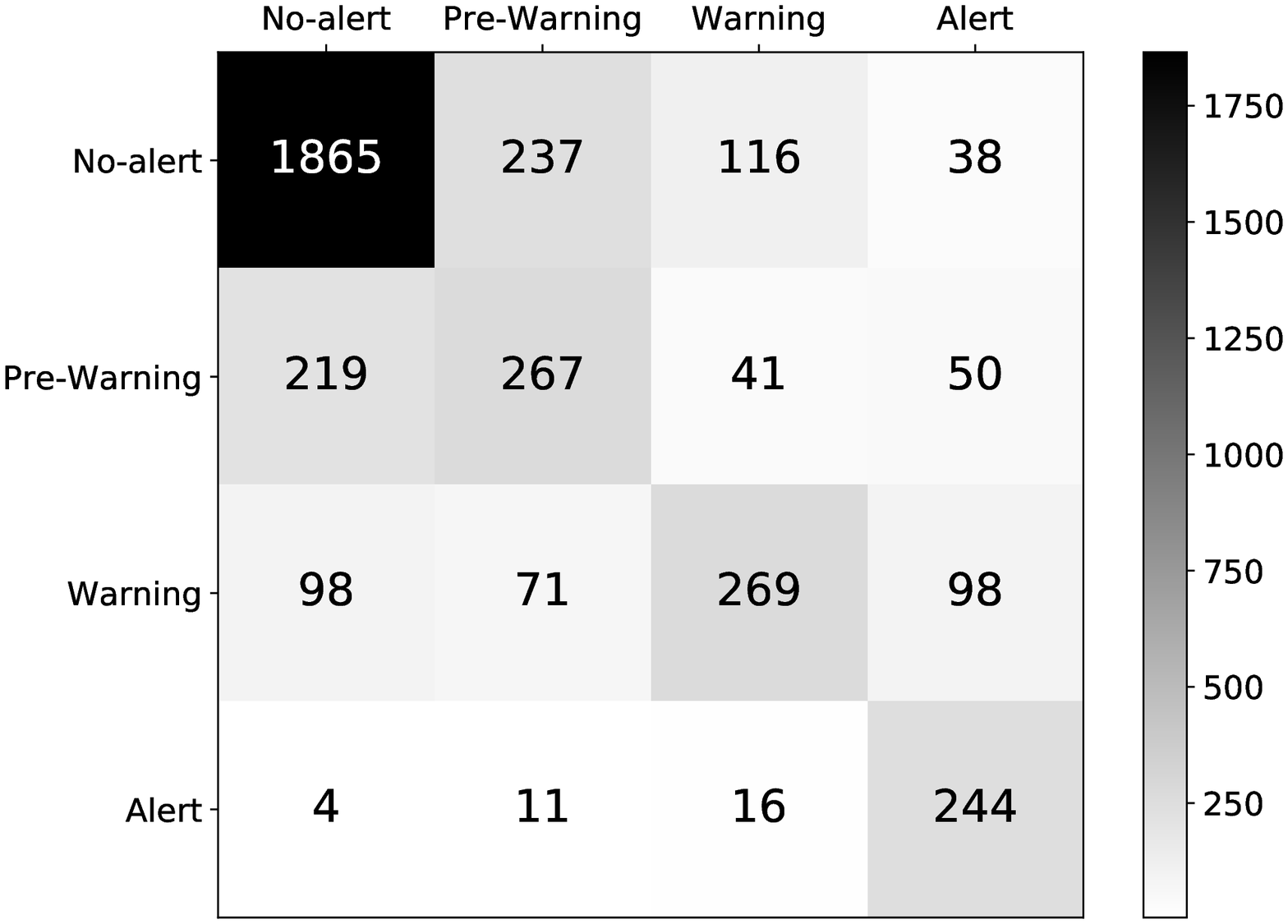}
    \label{fig:UTime:Bloque3UTime957550}
  }\\
  \subfloat[Second block. 168 hours. Normalized.]{
    \includegraphics[width=0.450\textwidth]{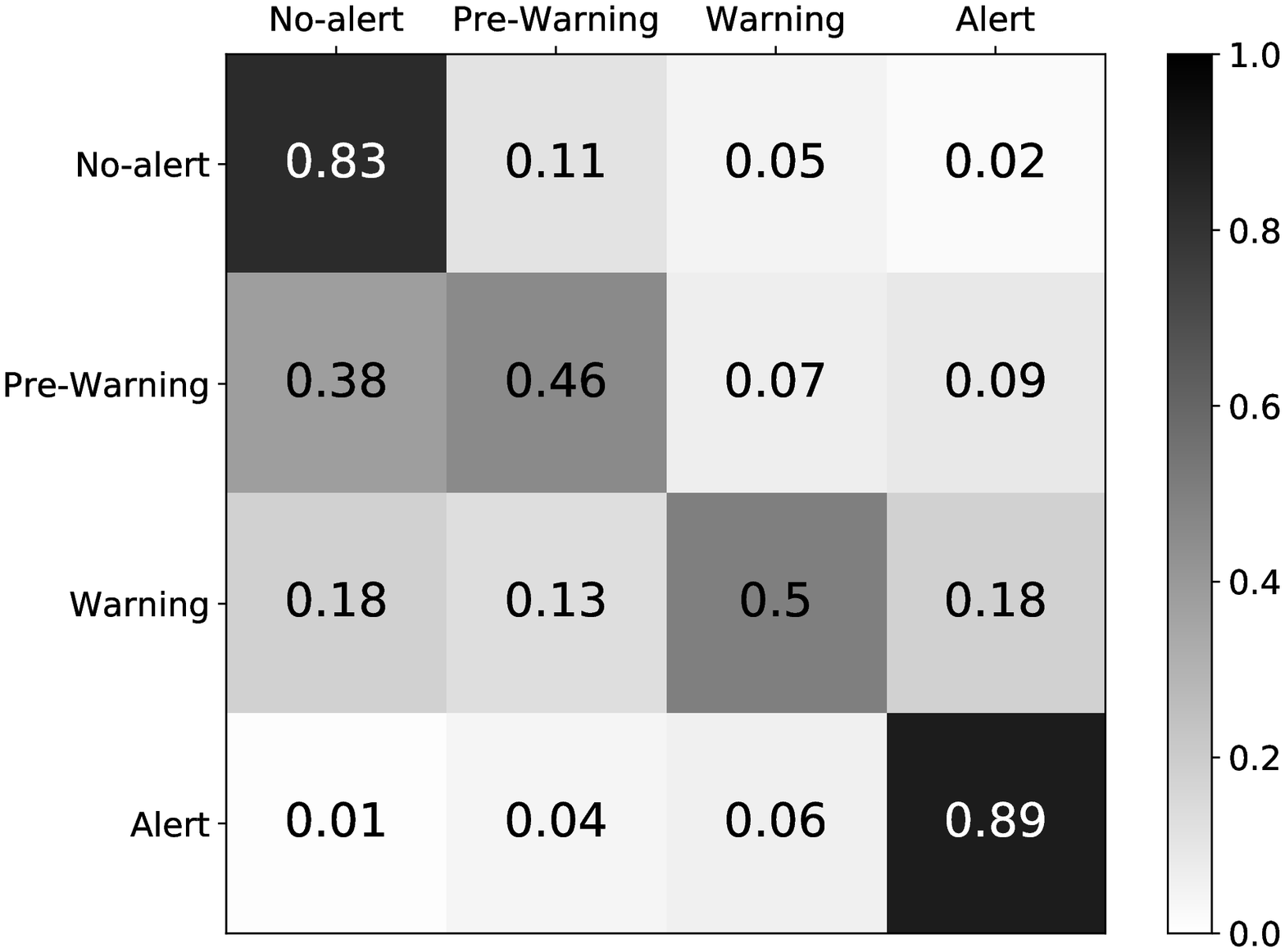}
    \label{fig:UTime:Bloque2UTime957550_norm}
  }
  \subfloat[Third block. 168 hours. Normalized.]{
    \includegraphics[width=0.450\textwidth]{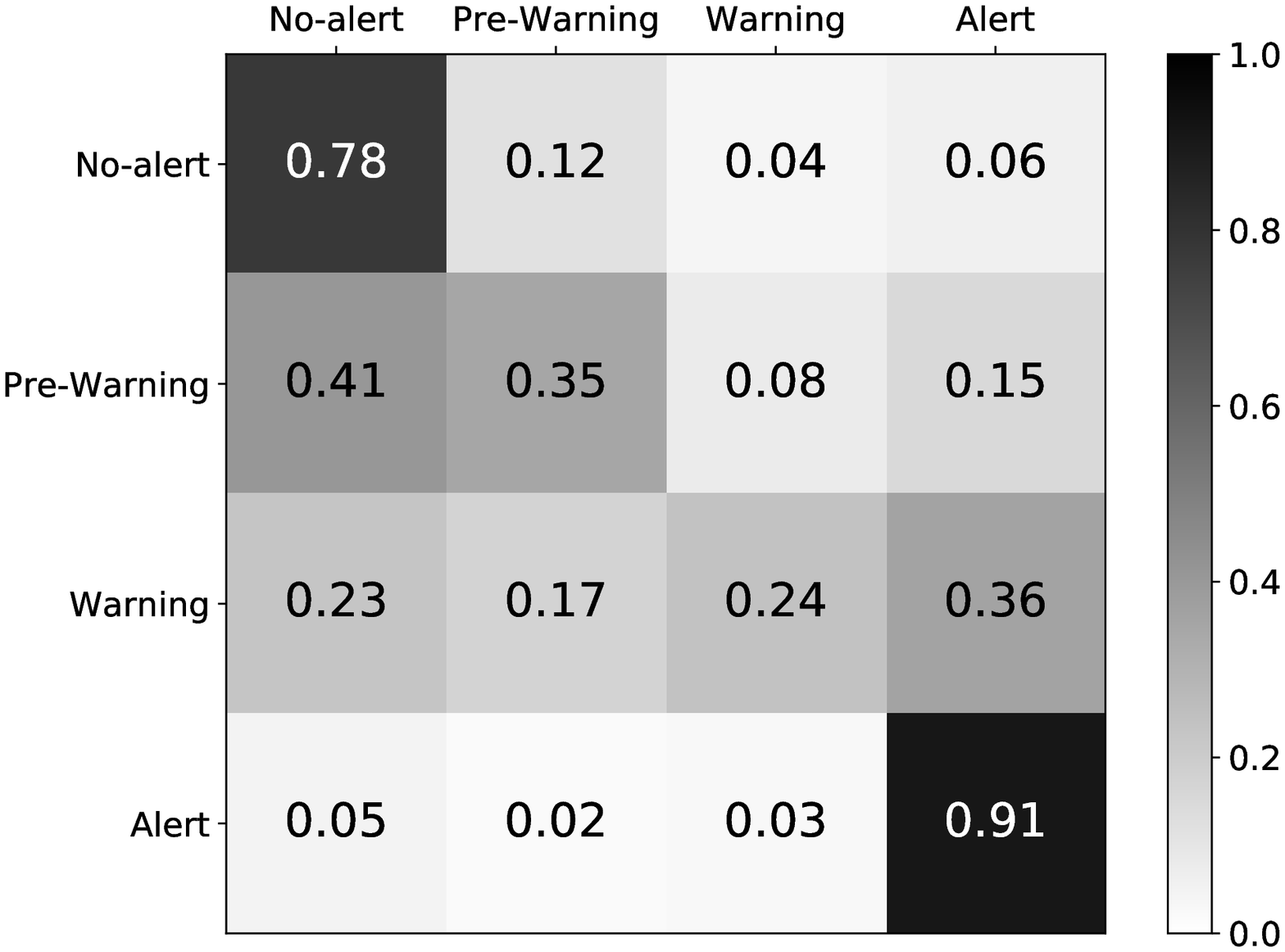}
    \label{fig:UTime:Bloque3UTime957550_norm}
  }
\caption{Confusion matrices obtained from the U-Time model for the Rule II. Only one input size is considered: 168 hours of data per station. Two blocks are considered: the second block (6:00-12:00) and the third block (12:00-18:00). Each row represents the actual class and each column represents the predicted class.}
\label{fig:UTime:RuleII}
\end{figure*}

Due to the structural needs of U-Time, we decide to only use the configuration of 168 hours of input data for all the experiments performed with this model (Figs. \ref{fig:UTime:RuleI} and \ref{fig:UTime:RuleII}). U-Time is a deep autoencoder that reduces 1.920 values of the input data into one single value of the latent vector. If we want to benefit from the properties of U-Time, the input size must be as large as possible. Shorter configurations are unable to pass through the U-Time model.

By analysing the results obtained from the U-Time model for the second and the third blocks, and for the Rule I and the Rule II, it can be observed that they are worse than those obtained from the other models. The accuracy of the U-Time model ranges 
from 43\% to 74\% for the second block and the Rule I (Fig. \ref{fig:UTime:Bloque2UTime959075_norm}), 
from 34\% to 85\% for the third block and the Rule I (Fig. \ref{fig:UTime:Bloque3UTime959075_norm}), 
from 46\% to 89\% for the second block and the Rule II (Fig. \ref{fig:UTime:Bloque2UTime957550_norm}), 
and from 24\% to 91\% for the third block and the Rule II (Fig. \ref{fig:UTime:Bloque3UTime957550_norm}). 
The predicted values are also scattered and not concentrated around the main diagonal. 
Adjacent errors grow up to 41\% for the Pre-Warning labels predicted as No-alert (Fig. \ref{fig:UTime:Bloque3UTime957550_norm}). And for the non-adjacent errors, they rise up to 36\% for the pre-warning labels predicted as Alert (Fig. \ref{fig:UTime:Bloque3UTime959075_norm}). The CNN model clearly outperforms the U-Time model for both blocks and rules.




\subsection{Economic Analyses}
Depending on the error direction, the parking regulatory system could overcharge over the customers. This scenario corresponds when the predicted labels have a higher criticality than the final true labels. Oppositely, when the predicted labels have a lower criticality that the true labels two pernicious effects appear. On the one hand, the charge for the use of the parking is undercharged; and on the other hand, the deterrent effect over the use of private motor vehicles vanish.   

An additional consideration is the fairness of the prediction. A well-balanced deep learning architecture with null economic impact in a temporal period, but with large errors in both directions is not acceptable. This kind of error leads to distrust of the citizens on IA applications. For this reason, a rate for the hours of each criticality level is proposed, the application of this rate is evaluated. The evaluation is undertaken by adding the incorrect charge in each direction of the errors and then their absolute values added.

The prices per hour established are for the criticality levels area: 0.40, 0.60, 1.20 and 2.40 euros. Currently the price per one hour of public parking at Madrid is 0.40 euros. In our proposal, this price is increased by 50\% when the criticality level goes up from No-alert to Pre-Warning, and then doubled for the two next levels: Warning and Alert. The aim of doubling the price for the two last criticality levels is to simulate the deterrent effect in the use of private transport. 

Then the economic impact ---per 6 hours block--- in each direction is evaluated and their absolute values added (Tables \ref{table:moneydistance:blockII} and \ref{table:moneydistance:blockIII}). The lower the values, the fairness of the architecture. Except for the case of the prediction for 1 day of the criticality level in Rule II and Block II, the CNN-based architecture produces the lowest errors and the highest fairness among the architectures tested. 

\begin{table}[h]
\centering
\caption{Evaluation of the deep architectures for block II based on economic criterion.}
\label{table:moneydistance:blockII}
\begin{tabular}{ccr rr} \hline 
    & &  CNN & LSTM & UTime  \\ \hline
\multirow{3}{*}{Rule I} & 1d  & 872.2 & 1044.2 & NaN \\
  & 3d  & 354.8 & 808.6 & NaN\\
  & 7d  & 337.6 & 1006.6 & 816.0\\ \hline
  \multirow{3}{*}{Rule II} & 1d  & 837.8 & 708.2 & NaN \\
  & 3d  & 566.4 & 662.6 & NaN \\
  & 7d  & 493.2 & 833.4 & 1583.2 \\ 
\hline 
\end{tabular}
\end{table}

\begin{table}[h]
\centering
\caption{Evaluation of the deep architectures for block III based on economic criterion.} \label{table:moneydistance:blockIII}
\begin{tabular}{ccr rr} \hline 
    & &  CNN & LSTM & UTime  \\ \hline
\multirow{3}{*}{Rule I} & 1d  & 246.8 & 522.8 & NaN \\
  & 3d  & 174.0 & 643.8 & NaN\\
  & 7d  & 114.0 & 593.2 & 1213.2 \\ \hline
  \multirow{3}{*}{Rule II} & 1d  & 306.4 & 393.0 & NaN \\
  & 3d  & 187.0 & 501.8 & NaN \\
  & 7d  & 202.6 & 362.4 & 660.2 \\ 
\hline 
\end{tabular}
\end{table}

\section{Conclusions}\label{section:conclusions}

In summary, air pollution is a critical problem in densely populated areas. Some measures have been implemented, like traffic limitations during periods of low-quality air or forecasting systems. In our research, we conclude that deep learning models can be useful to implement another kind of measure: some dynamic regulated parking services that would discourage motor vehicles parking when low-quality episodes are predicted. To achieve this objective, three different proposals are considered: one based on LSTM, one based on CNN and one based on the U-Time model.

While U-Time was a promising architecture, the results obtained were less than ideal. Due to its size, the computation time needed is between three and four times the computation time that the LSTM or the CNN models need, and the CNN model outperforms both of them. Analysing the results obtained by these models, we can draw some conclusions. First, it is better to use the percentile levels 95th, 75th and 50th than the 95th, 90th and 75th. This must be because of the fact that the last percentiles are closer together than the first ones. Second, there is a significant difference between the results obtained with the selected input sizes: usually it is better to use greater input sizes. Third, it is a good sign that the amount of false negatives is lesser than the amount of false positives. This is important for us due to the nature of our proposal: it is more desirable to have less motor vehicles in a regular episode, than more motor vehicles in a low-quality episode.

\section{Future Work}

In this work, the input data is taken from 12 of the 24 different air quality stations. There is a simple reason for this: only these 12 stations have been extracting data since 2010/01. Once there is enough data, it would be interesting to train again the models with data taken from the 24 air quality stations. In the case of the U-Time model, it is possible that, once there is more data, the model could be deeper and consequently get better results.

We have decided to use three different sizes of input data: 24 hours, 72 hours and 168 hours. It would be interesting to use greater sizes of input data, such as 336 hours (two weeks) or even 672 hours (a month). With greater input sizes it would be coherent to also train models with more parameters, and because of this it would be reasonable to train CNN or LSTM models with more layers; more filters or bigger kernels in the case of the CNN models; or more neurons per layer in the case of the LSTM model.

Considering our goal, we have proposed a dynamic parking regulatory system considering both the fairness and the deterrent effect in the use of private transport, and we have defined four different alert levels based on percentiles: No-alert, Pre-Warning, Warning and Alert. It would be interesting to work with expert environmental scientists that could establish more useful and realistic thresholds and pollution alert levels. 

\section*{Acknowledgements}

MCM is co-funded by the Spanish Ministry of Science and Innovation for funding support through the grant PID2020-113807RA-I00 "SERVICIOS INNOVADORES DE ANALISIS DE DATOS PARA EL EXPERIMENTO CMS". 
MAGN is partially supported by the Spanish Ministry of Science and Innovation through the project PID2019-107339GB-I00 "Advances in Computational Topology and Applications".






\begin{thebibliography}{20}

\bibitem{Linares2006}
Linares, C., D{\'i}az, J., Tob{\'i}as, A., De~Miguel, J.M., Otero, A., 2006.  Impact of urban air pollutants and noise levels over daily hospital admissions in children in {M}adrid: a time series analysis. Int. Arch. Occup. Environ. Health 79(2), 143--152. \url{https://doi.org/10.1007/s00420-005-0032-0}

\bibitem{Diaz1999}
D{\'i}az, J., Garc{\'i}a, R., Ribera, P., Alberdi, J.C., Hern{\'a}ndez, E., Pajares, M.S., Otero, A., 1999. Modeling of air pollution and its relationship with mortality and morbidity in {M}adrid, {S}pain. Int. Arch. Occup. Environ. Health 72(6), 366--376. \url{https://doi.org/10.1007/s004200050388} 

\bibitem{AlberdiOdriozola1998}
Alberdi Odriozola, J.C., D{\'i}az Jim{\'e}nez, J., Montero Rubio, J.C.,  Mir{\'o}n P{\'e}rez, I.J., Pajares Ort{\'i}z, M.S., Ribera Rodrigues, P., 1998. Air pollution and mortality in {M}adrid, {S}pain: a time-series analysis. Int. Arch. Occup. Environ. Health 71(8) 543--549. \url{https://doi.org/10.1007/s004200050321}

\bibitem{Nel804}
Nel, A., 2005. Air Pollution-Related Illness: Effects of Particles, Science 308(5723) 804--806. \url{https://doi.org/10.1126/science.1108752}

\bibitem{doi:10.1080/10473289.2006.10464485}
Pope, C.A., Dockery, D.W., 2006. Health effects of fine particulate air pollution: lines that connect, J Air Waste Manag Assoc. 56(6) 709--742 
\url{https://doi.org/doi: 10.1080/10473289.2006.10464485}

\bibitem{doi:10.3109/08958378.2011.593587}
R\"{u}ckerl, R., Schneider, A., Breitner, S., Cyrys, J., Peters, A., 2011.  Health effects of particulate air pollution: A review of epidemiological evidence, Inhal Toxicol. 23(10) 555--592
\url{https://doi.org/10.3109/08958378.2011.593587}

\bibitem{BoogaardErp}
Boogaard, H., Erp, A.M., 2019. Assessing health effects of air quality actions: what's next?, Lancet Planet Health 4(1) e4--e5
\url{https://doi.org/10.1016/S2468-2667(18)30235-4}

\bibitem{India}
Balakrishnan, K. {\it et al.}, 2019. The impact of air pollution on deaths, disease burden, and life expectancy across the states of India: the Global Burden of Disease Study 2017, Lancet Planet Health 3(1) e26--e39. 
\url{https://doi.org/10.1016/S2542-5196(18)30261-4}

\bibitem{ERivas} 
Rivas, E., Santiago, J.L., Lech\'on, Y., Mart\'in, F., Ari\~no, A., Pons, J.J., Santamar\'ia, J.M., 2019. CFD modelling of air quality in Pamplona City (Spain): Assessment, stations spatial representativeness and health impacts valuation, Sci Total Environ 649.
\url{https://doi.org/10.1016/j.scitotenv.2018.08.315} 

\bibitem{JMSantamaria}
Santamar\'ia, J.M. {\it et al.} 2017. Reduction of exposure of cyclists to urban air pollution, 2018. Pamplona: Programa LIFE13 ENV/ES/000417. ISBN: 978-84-947947-7-3. 
\url{http://liferespira.com/libro_life_respira_ING.pdf}

\bibitem{BORGE20181561}
Borge, R., {\it et al.} 2018 Application of a short term air quality action plan in {M}adrid ({S}pain) under a high-pollution episode - Part {I}: Diagnostic and analysis from observations, Sci. Total Environ. 1561-1573
\url{https://doi.org/10.1016/j.scitotenv.2018.03.149}

\bibitem{EBannon}
Bannon, E., 2019. Low-Emission Zones are a success - but they must now move to zero-emission mobility, Transport and Environment September 2019.

\bibitem{MadridCentral}
C\'ardenas-Montes, M., Evaluation of the Impact of Low-Emission Zone: Madrid Central as a Case Study, 2012.13782, 
2020.

\bibitem{SDG}
UN General Assembly (UNGA). A/RES/70/1Transforming our world: the 2030 Agenda for Sustainable Development. Resolut 25, 1–35,  2015.

\bibitem{opendatamadrid}
Open data {M}adrid, 2020, https://datos.madrid.es/portal/site/egob

\bibitem{Goodfellow-et-al-2016}
Goodfellow, I., Bengio, Y. Courville, A. 2016 Deep Learning, MIT Press, http://www.deeplearningbook.org

\bibitem{DBLP:conf/hais/Montes19}
C{\'a}rdenas-Montes, M. 2019 Forecast Daily Air-Pollution Time Series with Deep Learning. In: P\'erez Garc\'ia H., S\'anchez Gonz\'alez L., Castej\'on Limas M., Quinti\'an Pardo H., Corchado Rodr\'{\i}guez E. (eds) Hybrid Artificial Intelligent Systems. HAIS 2019. Lecture Notes in Computer Science, vol 11734. Springer, Cham. https://doi.org/10.1007/978-3-030-29859-3\_37
2019

\bibitem{muratov2020convolutional}
Muratov, M., and Kaur, S., and Szpakowicz, M. 2020 Convolutional Neural Networks Towards Arduino Navigation of Indoor Environments, arxiv preprint arXiv:2011.13893.

\bibitem{HochreiterS1997}
Hochreiter, S., Schmidhuber J. 1997 Long short-term memory. Neural Comput. 9(8), 1735-1780
https://doi.org/10.1162/neco.1997.9.8.1735

\bibitem{Choetal2014} 
 Cho, K. {\it et al.} 2014 Learning Phrase Representations using RNN Encoder-Decoder for Statistical Machine Translation. arXiv:1406.1078. 

\bibitem{Ronnenberg2015} 
Ronneberger, O., Fischer, P., Brox, T. 2015 U-Net: Convolutional Networks for Biomedical Image Segmentation. arXiv:1505.04597

\bibitem{perslev2019utime}
Perslev, M.. {\it et al} 2019 U-Time: A Fully Convolutional Network for Time Series Segmentation Applied to Sleep Staging, arxiv 1910.11162

\bibitem{JMLR:v15:srivastava14a}
Srivastava, N. {\it et al.} 2014 Dropout: A Simple Way to Prevent Neural Networks from Overfitting, J Mach Learn Res, 15(56), pp.: 1929-1958



\end{thebibliography}

\end{document}